\documentclass{article}

 \usepackage[preprint]{neurips_2026}

\usepackage[utf8]{inputenc} 
\usepackage[T1]{fontenc}    
\usepackage{hyperref}       
\usepackage{url}            
\usepackage{booktabs}       
\usepackage{amsfonts}       
\usepackage{nicefrac}       
\usepackage{microtype}      
\usepackage{xcolor}         
\usepackage{graphicx}
\usepackage[table]{xcolor}
\usepackage{multirow}
\usepackage{array}
\usepackage{makecell}
\usepackage{subcaption}
\usepackage{booktabs}
\usepackage{amsmath}
\usepackage{bbding}
\usepackage{tabularx}
\usepackage{xparse}
\usepackage[linesnumbered,ruled,vlined]{algorithm2e}
\newcolumntype{Y}{>{\centering\arraybackslash}X}
\newcolumntype{L}[1]{>{\raggedright\arraybackslash}p{#1}}

\newcommand{\cmark}{\Checkmark}
\newcommand{\xmark}{\XSolidBrush}
\newcommand{\pmark}{\(\triangle\)}

\definecolor{cellgray}{HTML}{EDEDED}
\definecolor{cellgreen}{HTML}{E2F0D9}
\definecolor{sectionblue}{HTML}{E8EFFA}
\definecolor{deltagreen}{HTML}{548235}
\definecolor{deltared}{HTML}{C00000}
\definecolor{deltagray}{HTML}{A9A9A9}
\definecolor{ablationblue}{HTML}{E8EFFA}

\title{DriveHierarchy: A Benchmark for Diagnosing VLM Driving Capabilities from Open-Loop Understanding to Closed-Loop Execution}

\author{%
  Chengkai~Xu\textsuperscript{1}\thanks{Equal contributions.}\quad
  Jiaqi~Liu\textsuperscript{2}\footnotemark[1]\quad
  Yicheng~Guo\textsuperscript{1}\footnotemark[1]\quad
  Peng~Hang\textsuperscript{1}\quad
  Jian~Sun\textsuperscript{1}\thanks{Corresponding author: \texttt{sunjian@tongji.edu.cn}.} \\
  \normalfont\textsuperscript{1}College of Transportation, Tongji University, Shanghai, China \\
  \normalfont\textsuperscript{2}Department of Computer Science, UNC Chapel Hill, NC, USA \\
  \normalfont\texttt{\{xuchengkai,guo\_yicheng,hangpeng,sunjian\}@tongji.edu.cn} \\
  \normalfont\texttt{jqliu@cs.unc.edu}
}

\begin{document}

\maketitle

\begin{abstract}
\label{abstract}
Evaluating VLM-based autonomous driving remains difficult because driving competence is composite, where a capable system must ground traffic participants and hazards, integrate context across views and time, reason about future evolution, and act appropriately under closed-loop interaction. Existing benchmarks usually assess either open-loop understanding or closed-loop driving but provide limited structure for explaining how these abilities are organized, how they relate, and how they may inform model diagnosis and improvement. We present \textsc{DriveHierarchy}, a hierarchical benchmark that organizes VLM-based autonomous driving into four ranks, spanning perceptual grounding, contextual memory, mental reasoning, and closed-loop execution. To instantiate this hierarchy, we integrate multiple open-source autonomous-driving datasets into a unified open-loop benchmark with 76,798 question-answer pairs over 84,279 frames and develop a closed-loop simulation platform with interactive scenario construction on a real-world road network, from which 100 driving scenarios are curated for embodied evaluation. 
Experiments on 15 VLMs show that \textsc{DriveHierarchy} captures structured but non-redundant capability variation, relates open-loop understanding to closed-loop driving, and provides a practical basis for diagnosis and benchmark-guided optimization. \textsc{DriveHierarchy} therefore serves as a unified framework for evaluating and improving VLM-based autonomous driving systems. An anonymized project has been released on \url{https://github.com/PerfectXu88/DriveHierarchy}. 
\end{abstract}

\section{Introduction}
\label{introduction}
End-to-end autonomous driving has emerged as a central paradigm for learning-based driving systems by directly mapping sensory observations to driving actions within a unified architecture\cite{chen2024end, zheng2024genad, xu2025survey}.
Relative to traditional modular pipelines, this formulation offers a more integrated framework for coupling perception and decision-making, while avoiding hand-designed intermediate interfaces\cite{sun2025sparsedrive, jia2023think}. 
Recent progress in foundation models has further accelerated this direction, with vision-language models (VLMs) attracting increasing attention for autonomous driving due to their strong multimodal representation and reasoning capabilities\cite{zheng2024genad,jiang2025survey,xu2025tell}.
By unifying visual understanding with semantic inference, VLM-based driving systems provide a promising foundation for integrated scene interpretation and decision-making\cite{zhou2025autovla, huang2026automot}.
However, their growing adoption also exposes a fundamental challenge: \textit{how should the performance of such systems be evaluated?}

Evaluating such systems is nontrivial because driving intelligence is not a monolithic ability\cite{yang2025drivemoe, xu2025knowledge,nie2024reason2drive}.
An excellent driving system must ground traffic participants and hazards in the current scene, integrate information coherently across viewpoints and time, reason about how situations may evolve, and ultimately select appropriate actions under closed-loop interaction\cite{jiang2025survey, zhao2025survey}. 
Existing evaluation protocols can usually capture only part of this capability chain. Open-loop driving-language and question-answering benchmarks can probe perceptual and reasoning-related abilities, but they often entangle heterogeneous capabilities within task-level scores that are difficult to interpret diagnostically\cite{xie2025vlms,marcu2024lingoqa,meng2025your}.
Closed-loop evaluations, in contrast, provide direct evidence of embodied driving performance, yet route-level outcomes such as success, collision, or completion rarely reveal which internal capability is responsible for the observed behavior. Consequently, current benchmarks are valuable for system comparison, but considerably less informative for explaining model behavior, isolating failure modes, or guiding targeted improvement.

\begin{figure}
    \centering
    \includegraphics[width=\linewidth]{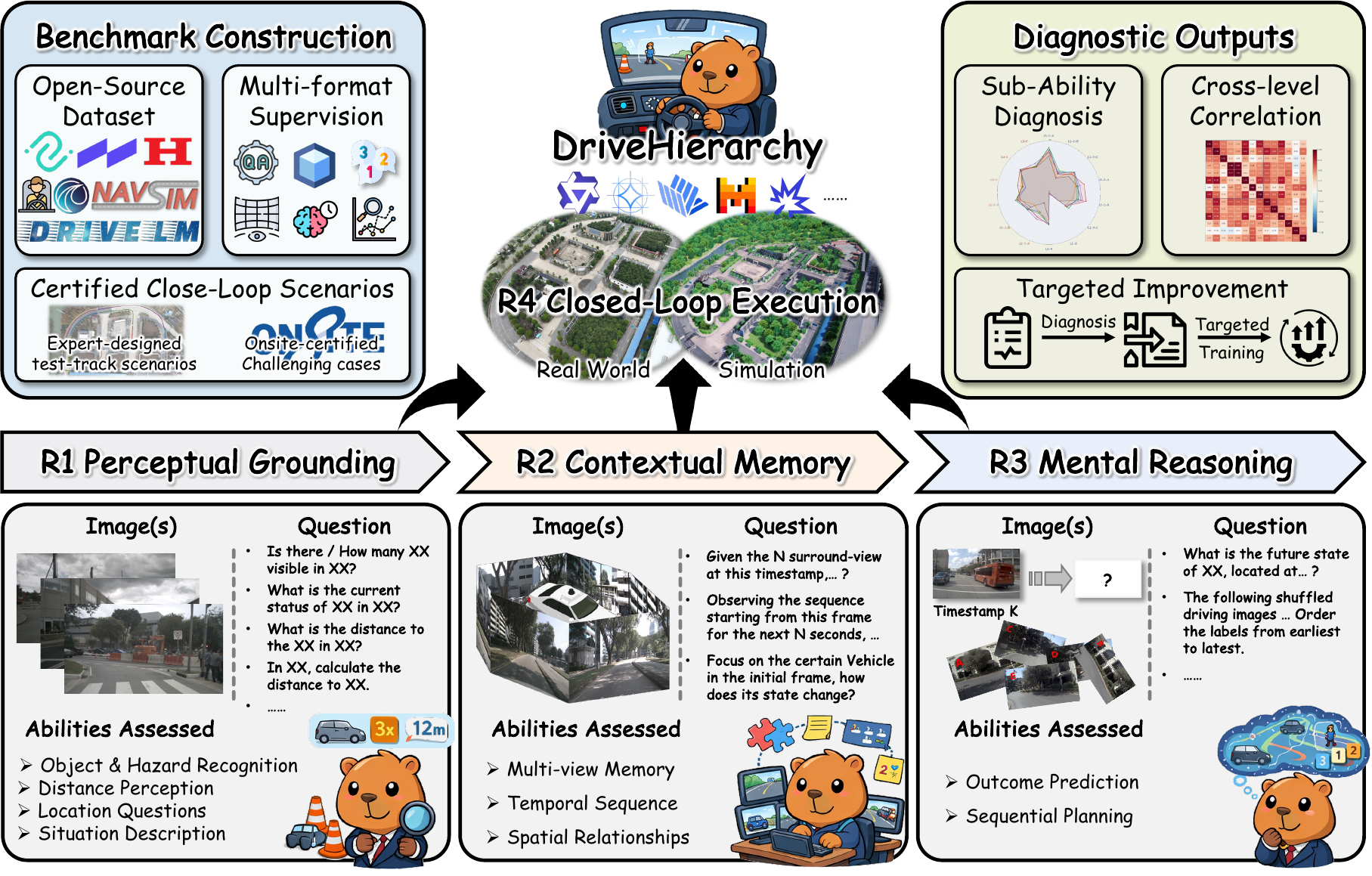}
    \caption{\textbf{Overview of \textsc{DriveHierarchy},} which organizes VLM-based autonomous driving into four progressively structured ranks, spanning perceptual grounding, contextual memory, mental reasoning, and closed-loop execution.}
    \label{fig:Fig_framework_overview}
\end{figure}

To better address this limitation, we draw inspiration from cognitive science, which views complex embodied intelligence as arising from progressively organized levels of capability rather than from a single undifferentiated skill\cite{carroll1993theory, xiao2026spatialtree,eggers2013cognition}.
Based on this view, we propose \textsc{\textbf{DriveHierarchy}}, a hierarchical benchmark for VLM-based autonomous driving, as illustrated in Figure~\ref{fig:Fig_framework_overview}. 
The benchmark structures driving capability into four ranks, spanning \textbf{R1} Perceptual Grounding, \textbf{R2} Contextual Memory, \textbf{R3} Mental Reasoning, and \textbf{R4} Closed-loop Execution. 
By organizing evaluation along this hierarchy, the benchmark supports both fine-grained capability analysis and system-level assessment, enabling a more explicit examination of model strengths, weaknesses, capability relationships, and their connection to downstream driving behavior.

To instantiate this hierarchy, we integrate multiple public autonomous driving datasets to construct a capability-oriented corpus in both full-scale and streamlined forms, comprising 84,279 frames and 76,798 question-answer pairs in total.
To support embodied evaluation beyond static analysis, we further develop a joint simulation platform for interactive scenario construction and create 100 functional driving scenarios for closed-loop testing.
On this basis, we conduct a comprehensive study of 15 open-source VLMs and a targeted supervised fine-tuning (SFT) case study to examine the interactions among different capabilities and their influence on downstream driving performance.
The results show that \textsc{DriveHierarchy} enables fine-grained diagnosis of model strengths and weaknesses, reveals meaningful relationships across capability levels, and provides a useful basis for targeted analysis and intervention design.

Our contributions are as follows:
\begin{itemize}
    \item A hierarchical capability benchmark for VLM-based autonomous driving is proposed, integrating multiple public autonomous driving datasets and decomposing driving intelligence into four progressively organized ranks from perceptual grounding to closed-loop execution.

    \item A joint simulation platform with interactive scenario construction is developed to support closed-loop evaluation under diverse functional driving scenarios.

    \item A comprehensive benchmarking of 15 open-source VLMs and targeted fine-tuning studies are conducted to support capability-level analysis and model improvement.
\end{itemize}

\section{Related Work}
\label{Related_Works}

\paragraph{VLM-Based End-to-End Autonomous Driving.}
End-to-end autonomous driving has increasingly evolved from purely sensor-to-control formulations toward unified architectures that incorporate semantic reasoning and language~\cite{chen2024end, liao2025diffusiondrive, xing2025openemma}.
Recent advances in VLMs have further accelerated this trend by enabling driving systems to couple visual perception with instruction understanding, scene interpretation, and high-level decision support within a common multimodal framework~\cite{xing2025openemma, ma2024dolphins, sima2024drivelm}. 
This line of work has expanded from language-assisted driving agents to VLM-based end-to-end systems and further to Vision-Language-Action models that directly predict driving actions or control signals~\cite{cao2026fastdrivevla, zhou2026opendrivevla, renz2025simlingo, fu2025orion, zhou2025autovla}. 
As the backbone of these approaches, VLMs offer stronger representational unification and greater interpretability, but they also introduce new challenges for evaluation, since perception, memory, reasoning, and decision-making are increasingly entangled within a single model~\cite{xie2025vlms,li2025fine}.

\paragraph{Autonomous Driving Benchmarks.}
Recent autonomous driving benchmarks have substantially advanced the evaluation of learning-based driving systems, yet remaining fragmented in both objective and granularity. 
Open-loop benchmarks, including driving-language, scene-understanding, and question-answering datasets, provide useful testbeds for perception, interaction understanding, and reasoning, but they typically organize evaluation around task-specific scores and therefore offer limited visibility into the internal capability structure of a model~\cite{cao2024maplm,fruhwirth2025stsbench,xie2025vlms,meng2025your,nie2024reason2drive,hao2026styledrive}. 
Closed-loop benchmarks, in contrast, assess embodied performance through interactive simulation or scenario execution, making them more faithful to downstream driving behavior. However, they often reduce evaluation to route completion, collision statistics, or aggregate success rates, which are insufficient for diagnosing which underlying abilities contribute to failure~\cite{yu2025drivee2e, jia2024bench2drive, xu2022safebench,dauner2024navsim}.
As a result, existing benchmarks are effective for model comparison, but less effective for explaining competence, isolating weaknesses, and relating open-loop ability to closed-loop driving outcomes.

\paragraph{Hierarchical Capability Decomposition for Embodied Intelligence.}
Cognitive science has long treated intelligence as hierarchically organized rather than flat\cite{eggers2013cognition}, with Carroll's three-stratum theory as a canonical example~\cite{carroll1993theory}.
Reviews in neuroscience and psychometrics likewise argue that intelligence is a hierarchy in which partially separable abilities contribute to coherent higher-level performance~\cite{deary2022genetic}.
This conceptual tradition has increasingly influenced benchmark design, where recent work in multimodal evaluation has moved beyond flattened aggregate scores toward fine-grained evaluations that separately probe abilities, providing a more informative alternative to flat aggregate metrics~\cite{li2024embodied, yang2025embodiedbench, agrawal2025uinavbench, song2025robospatial}. Such work suggests that capability decomposition is essential for understanding not only whether a model succeeds, but also how its internal competencies are structured and where its failure modes originate\cite{xiao2026spatialtree, wang2025enact,zhang2026theory}. 
However, in autonomous driving, this view remains insufficiently developed. Existing evaluations still tend to organize results around task families or outcome metrics, while providing limited structure for understanding how lower-level capabilities support higher-level reasoning and embodied driving behavior~\cite{jia2024bench2drive,xie2025vlms}.

\section{DriveHierarchy}
\label{drivehierarchy}

\subsection{Overview and Design Principle}
As shown in Fig.~\ref{fig:Fig_framework_overview}, \textsc{DriveHierarchy} organizes VLM-based autonomous driving into four progressively structured ranks: \textbf{R1} Perceptual Grounding, \textbf{R2} Contextual Memory, \textbf{R3} Mental Reasoning, and \textbf{R4} Closed-loop Execution. The first three ranks are instantiated as open-loop tasks that probe distinct aspects of scene understanding and cognitive processing, while the fourth rank evaluates whether these capabilities are associated with actual driving behavior under continuous interaction.
This design provides fine-grained diagnosis at the local level and a structured basis for analyzing how lower-level capabilities relate to higher-level reasoning and embodied driving at the system level. 
To instantiate this hierarchy, we integrate multiple public driving datasets into a capability-oriented open-loop corpus and complement it with a controllable closed-loop benchmark built on an interactive co-simulation platform. 
The refined version used for all reported $R1$-$R3$ results contains 14,000 question-answer pairs over 31,945 frames, together with 100 closed-loop scenarios for $R4$. Dataset provenance and benchmark composition details are provided in Appendix~\ref{appendix:data_sources}.

\begin{figure}[t]
    \centering
    \begin{subfigure}[b]{0.33\linewidth}
        \centering
        \includegraphics[width=\linewidth]{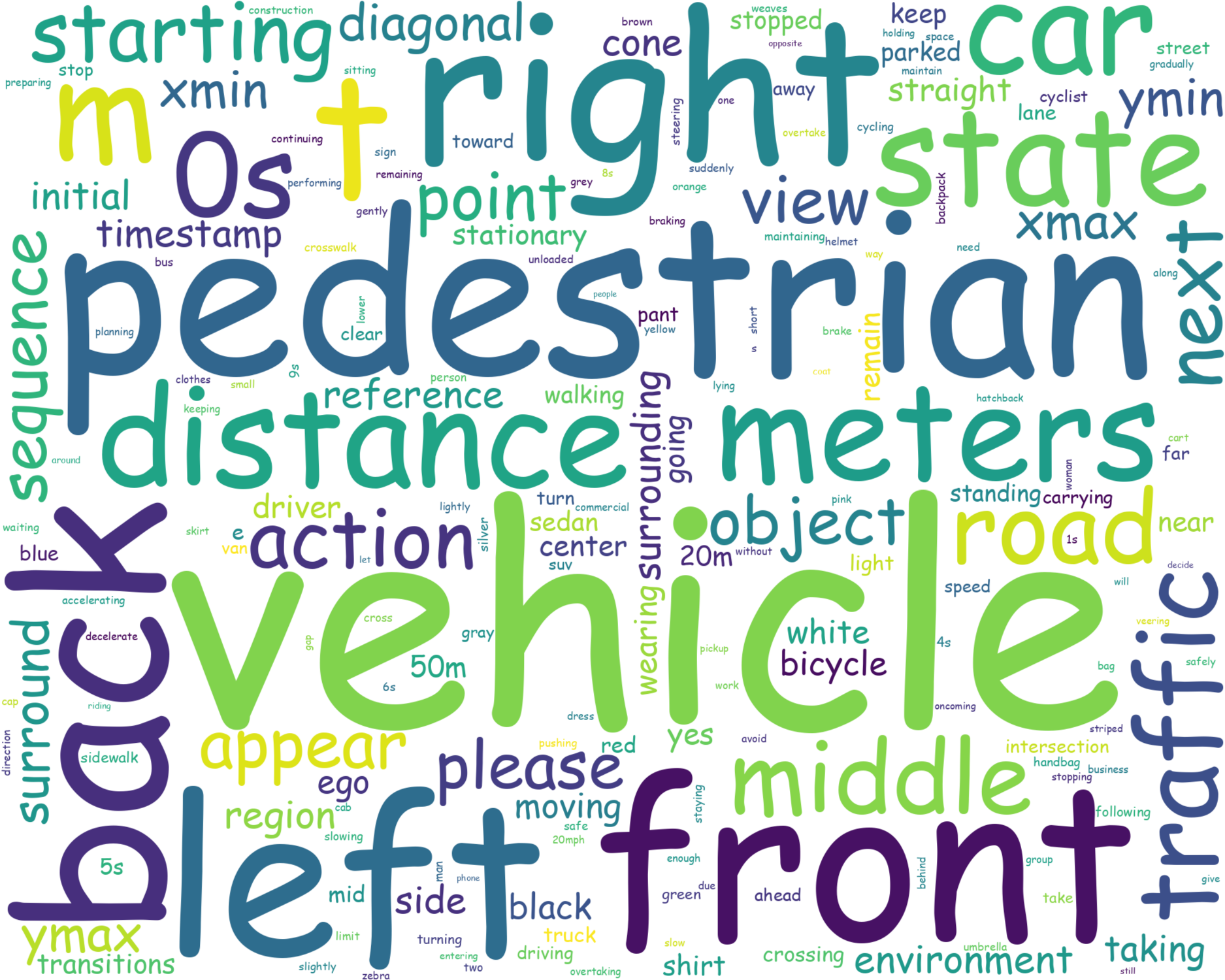}
        \caption{Word Distribution for $R1–R3$}
        \label{fig:Fig_wordcloud}
    \end{subfigure}
    \hfill
    \begin{subfigure}[b]{0.26\linewidth}
        \centering
        \includegraphics[width=\linewidth]{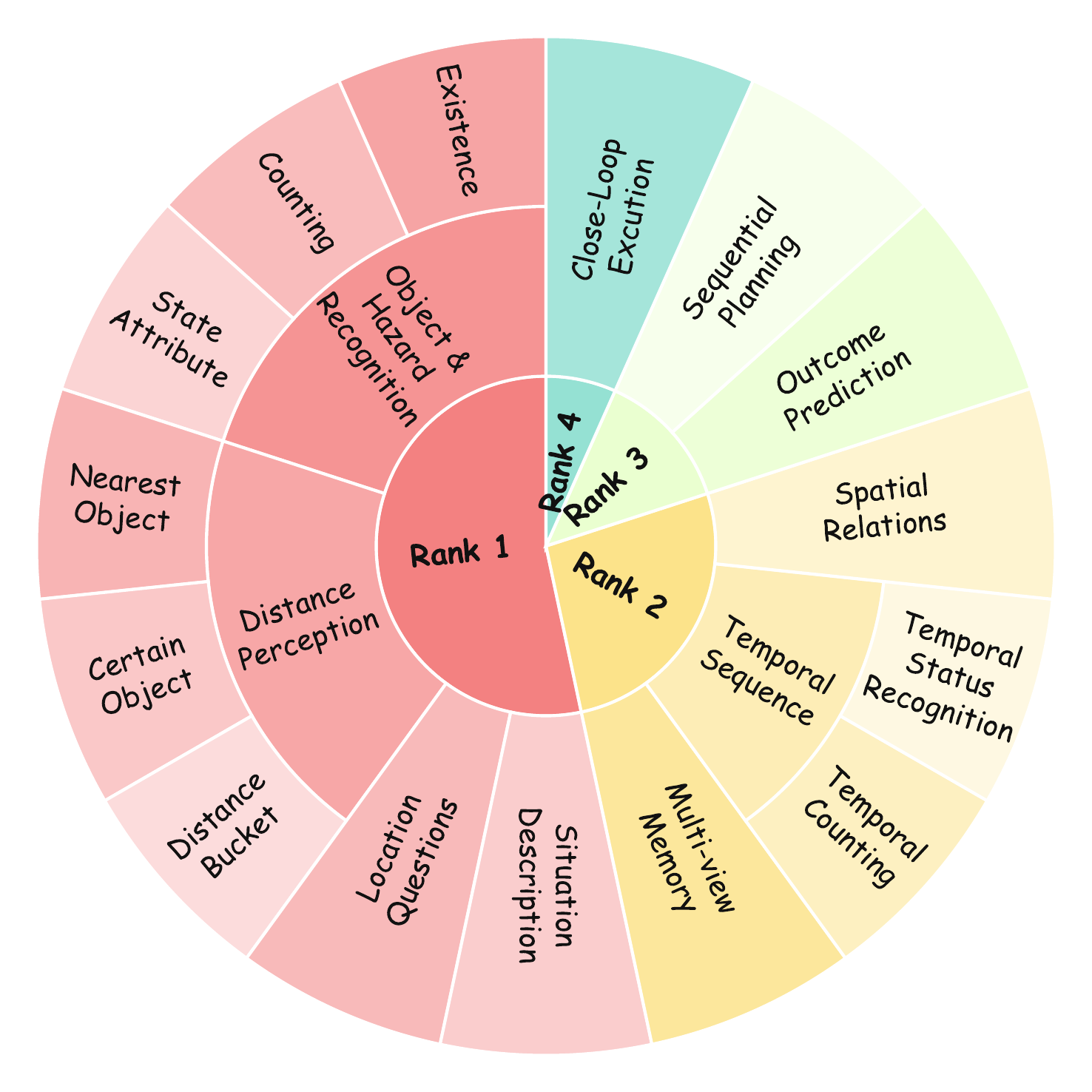}
        \caption{Overall Distribution.}
        \label{fig:Fig_sunburst_chart}
    \end{subfigure}
    \hfill
    \begin{subfigure}[b]{0.38\linewidth}
        \centering
        \includegraphics[width=\linewidth]{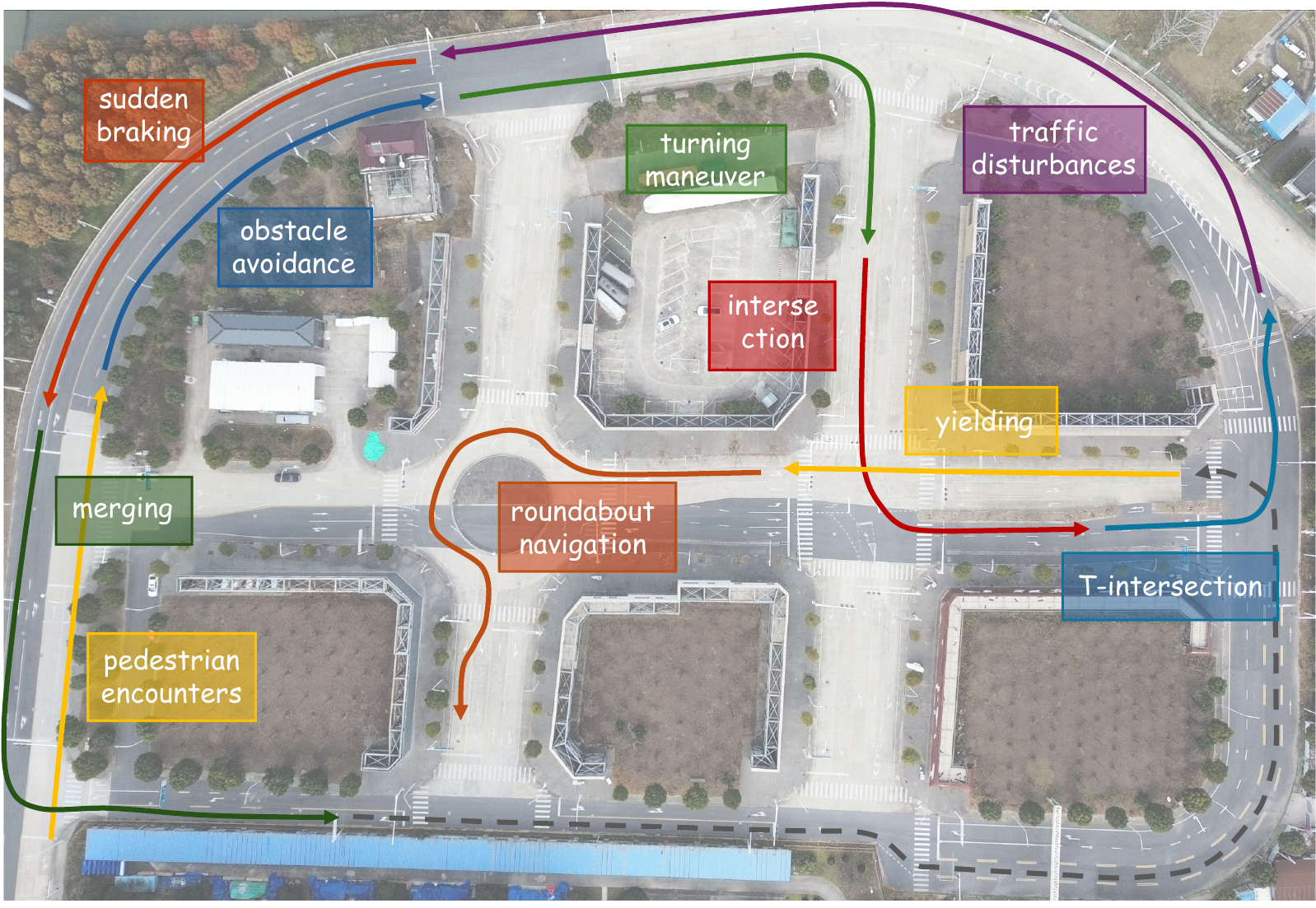}
        \caption{Scenario Distribution for $R4$.}
        \label{fig:Fig_Rank_4_distribution}
    \end{subfigure}
    \caption{\textbf{Distribution of the \textsc{DriveHierarchy} benchmark,} which summarizes the data composition of the open-loop ranks and the scenario allocation of the closed-loop benchmark.}
    \label{fig:Distribution}
\end{figure}

\subsection{Open-Loop Benchmark Construction}

\paragraph{\textit{R1}: Perceptual Grounding.}
$R1$ contains 8 task types and evaluates whether a model can build a reliable single-frame representation of the current scene. We group these tasks into four capability families. The first family, \emph{\textbf{object and hazard recognition}}, includes existence judgment, target counting, and state identification. These tasks are derived from annotated driving scenes by converting object categories and state labels into binary, numeric, or multiple-choice question-answer pairs. The second family, \emph{\textbf{distance perception}}, contains nearest-object distance estimation, referred-object distance estimation, and distance-bucket counting. These tasks are constructed from ego-object geometry and calibrated image coordinates, allowing the benchmark to test both direct metric estimation and language-guided spatial reference. The third family, \emph{\textbf{visual grounding}}, asks the model to localize an object described in natural language and is supervised by annotated bounding boxes. The fourth family, \emph{\textbf{situation description}}, evaluates whether the model can summarize the current traffic state in natural language from image evidence.

\paragraph{\textit{R2}: Contextual Memory.}
$R2$ extends the benchmark from single-frame understanding to structured scene representation across views, time, and object relations, containing 4 task types and evaluates whether a model can preserve memory when information is distributed across views, time, and entities. We instantiate this rank with three capability families. \emph{\textbf{Multi-view memory}} is constructed from synchronized six-camera observations at the same timestamp and asks the model to infer the number of unique surrounding targets after resolving cross-view overlap. To avoid reducing the task to repeated single-view counting, we retain only samples that require genuine cross-view integration. \emph{\textbf{Temporal memory}} is instantiated in two forms. The first uses fixed short clips to count unique targets appearing over time, thereby testing memory-based aggregation across frames. The second tracks a target object across a temporal window and asks the model to identify whether its state changes, which directly probes dynamic state-transition understanding. \emph{\textbf{Spatial relation reasoning}} constructs pairwise queries between visible entities and requires the model to infer relative positional relations such as front, rear, left, or right.

\paragraph{\textit{R3}: Mental Reasoning.}
$R3$ contains 2 task types, marking the transition from contextual understanding to future-oriented reasoning. The first task family, \emph{\textbf{outcome prediction}}, asks the model to infer the likely future behavior of a target traffic participant from the current scene context. These questions are derived from future-aware driving annotations and are posed as multiple-choice predictions over plausible downstream actions or states. The second family, \emph{\textbf{sequence planning}}, presents a set of candidate observations and requires the model to recover their correct temporal order. At this rank, the benchmark tests whether the model can go beyond describing the current scene and instead reason about how the scene may evolve.

\paragraph{Task Formulation and Evaluation.}
Across $R1$-$R3$, each open-loop sample is formulated as a tuple $(x_i, q_i, y_i)$, where $x_i$ denotes the visual input, $q_i$ the task-specific natural-language query, and $y_i$ the ground-truth target. Depending on the task, $x_i$ may be a single image, a synchronized multi-view observation set, or a short temporal clip. Given a model prediction $\hat{y}_i$, we assign each sample a bounded score $s_i \in [0,100]$. This formulation provides a unified evaluation interface while preserving the distinct supervision structure of different capability types. 
Distinct evaluation methods are designed for different question types, with further details regarding scoring and implementation discussed in Appendix~\ref{Evaluation_Metrics_Designs}.

\subsection{Closed-loop Benchmark Construction}

\begin{figure}
    \centering
    \includegraphics[width=\linewidth]{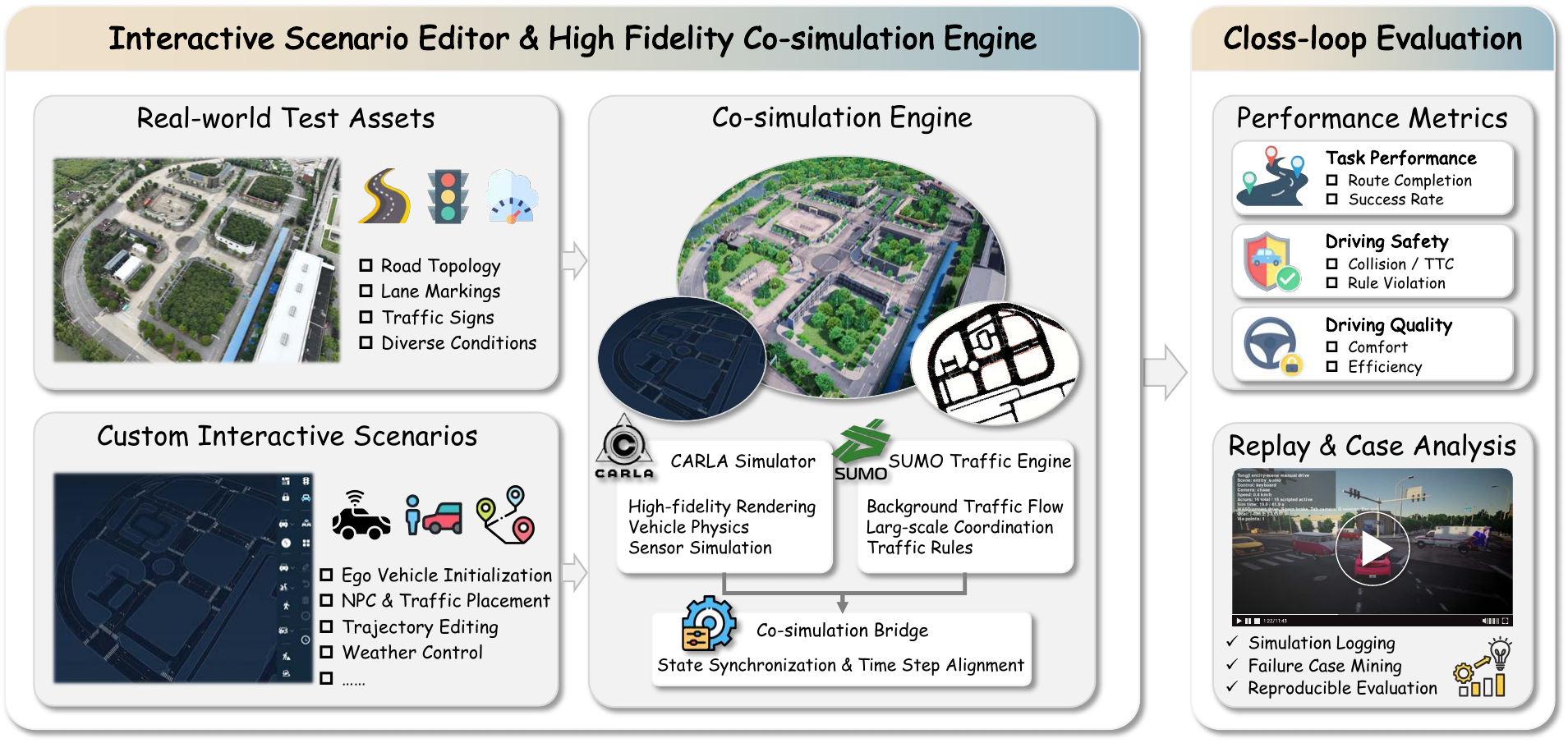}
    \caption{\textbf{Closed-loop evaluation platform for $R4$,} which combines simulation with interactive scenario construction on a real-road layout, enabling controllable evaluation under diverse traffic interactions.}
    \label{fig:Fig_R4_platform}
\end{figure}

\paragraph{Platform and Scenario Design.}
$R4$ evaluates whether the capabilities measured in open loop transfer to actual driving behavior under continuous interaction. As shown in Fig.~\ref{fig:Fig_R4_platform}, we construct a CARLA-SUMO co-simulation platform on a real-world road network that preserves road topology, lane structure, and traffic semantics. CARLA~\cite{dosovitskiy2017carla} provides the ego-centric simulation environment and multimodal observations, while SUMO\cite{krajzewicz2002sumo} generates and controls the background traffic flow, allowing the platform to remain both physically expressive and behaviorally scalable. On top of this backbone, we develop an interactive scenario editing framework that supports configurable placement and control of the ego vehicle, surrounding vehicles, pedestrians, and other dynamic elements together with their trajectories, interactions, and traffic-control conditions. Using this framework, we curate 100 closed-loop scenarios covering representative urban interaction patterns, including \textit{pedestrian encounters, obstacle avoidance, turning maneuvers, intersections, traffic disturbances, sudden braking, merging, yielding, and roundabout navigation}.

\paragraph{Scoring Design.}
For each scenario, the evaluated model receives multimodal driving inputs, including front-view imagery, bird's-eye-view context, and navigation cues, and must output low-level control signals for vehicle actuation. The closed-loop score jointly measures route completion, safety, and efficiency. Let $S_{\mathrm{route}}$, $S_{\mathrm{safety}}$, and $S_{\mathrm{eff}}$ denote the corresponding terms. The final scene score is defined as
\begin{equation}
S_{\mathrm{R4}} = S_{\mathrm{route}} \cdot S_{\mathrm{safety}} \cdot S_{\mathrm{eff}}.
\end{equation}
This multiplicative design prevents route progress alone from masking unsafe or inefficient behavior. The route term measures normalized progress toward the destination, the safety term aggregates infraction penalties and lane-keeping quality, and the efficiency term evaluates whether the realized trajectory is unnecessarily long or slow relative to the route reference. The detailed formulation, event penalties, and implementation-specific thresholds are provided in Appendix~\ref{appendix:r4_scoring}.

\section{Experiments}

\subsection{Experimental Setup}
We evaluate \textsc{DriveHierarchy} on 15 VLMs, covering both general-purpose VLMs and driving-specialized models. The open-loop benchmark spans the three capability ranks $R1$-$R3$, while $R4$ evaluates the same models in closed-loop driving scenarios. To test whether the benchmark can guide targeted model improvement, we further perform SFT on Qwen3-VL-8B-Instruct by selecting the task groups on which its gap to the corresponding state-of-the-art result is the largest.

\subsection{Open-Loop Benchmark Results}

\begin{table*}
\centering
\caption{\textbf{Open-loop benchmark results on DriveHierarchy for 15 VLMs.} The overall score is computed using the frozen calibration weights shown in the header. Gray and green cells mark the second-best and best results, respectively.}
\scriptsize
\setlength{\tabcolsep}{2pt}
\renewcommand{\arraystretch}{1.15}
\resizebox{\textwidth}{!}{%
\begin{tabular}{l*{15}{c}}
\toprule
\multirow{3}{*}{\textbf{Models}} & \multirow{3}{*}{\textbf{Avg.Score}} & \multicolumn{8}{c}{\textbf{R1 Perceptual Grounding}} & \multicolumn{4}{c}{\textbf{R2 Contextual Memory}} & \multicolumn{2}{c}{\textbf{R3 Mental Reasoning}} \\
\cline{3-16}
& & R1\_1\_A & R1\_1\_B & R1\_1\_C & R1\_2\_A & R1\_2\_B & R1\_2\_C & R1\_3 & R1\_4 & R2\_1 & R2\_2\_A & R2\_2\_B & R2\_3 & R3\_1 & R3\_2 \\
\cline{3-16}
& & 0.030 & 0.027 & 0.056 & 0.035 & 0.036 & 0.029 & 0.056 & 0.031 & 0.064 & 0.079 & 0.067 & 0.090 & 0.233 & 0.167 \\
\hline
\rowcolor{sectionblue}
\multicolumn{16}{l}{Generalist} \\
Qwen3-VL(2B)\cite{qwen3technicalreport} & 43.91 & 68.80 & 29.54 & 39.10 & 25.79 & 38.97 & 35.46 & 0.99 & \cellcolor{cellgray}50.53 & 33.78 & 41.36 & 49.90 & 39.70 & 59.20 & 46.58 \\
InternVL3\_5(2B)\cite{wang2025internvl3_5} & 39.28 & 62.00 & 30.97 & 45.40 & 7.62 & 9.92 & 33.41 & 0.07 & 36.35 & 30.71 & 35.94 & 44.90 & 35.20 & 54.70 & 45.45 \\
Qwen3-VL(8B)\cite{qwen3technicalreport} & 45.09 & \cellcolor{cellgray}70.00 & 33.32 & 51.20 & 46.70 & 46.76 & 35.74 & 1.05 & \cellcolor{cellgray}50.23 & 32.87 & 41.25 & \cellcolor{cellgray}53.60 & 37.70 & 56.90 & 45.84 \\
InternVL3\_5(8B)\cite{wang2025internvl3_5} & 43.49 & 62.70 & 27.06 & 49.40 & 42.47 & 38.83 & 32.62 & 0.90 & 44.23 & 33.91 & 29.58 & 49.70 & 44.50 & 57.50 & 45.70 \\
ZwZ(8B)\cite{wei2026zooming} & 45.09 & \cellcolor{cellgray}70.30 & \cellcolor{cellgray}37.41 & 50.60 & 43.92 & 44.60 & 39.29 & 1.00 & 48.91 & 35.63 & 41.98 & 53.30 & 41.10 & 54.60 & 46.11 \\
MiniCPM-V-4\_5(9B)\cite{yu2025minicpmv45cookingefficient} & 45.50 & 66.10 & 32.24 & 45.70 & \cellcolor{cellgray}56.53 & \cellcolor{cellgray}59.60 & 33.67 & 1.06 & 49.87 & 41.95 & 39.46 & 50.90 & 42.40 & 55.00 & 45.16 \\
Gemma-3-it(12B)\cite{gemma_2025} & 46.27 & 63.00 & 34.65 & 50.00 & 49.55 & 53.97 & 31.49 & 2.22 & 47.11 & \cellcolor{cellgray}45.40 & \cellcolor{cellgray}42.94 & 49.10 & 37.20 & 56.40 & \cellcolor{cellgreen}\textbf{50.28} \\
Pixtral(12b)\cite{agrawal2024pixtral} & 42.31 & 56.90 & 25.02 & 39.80 & 45.27 & 51.55 & 21.46 & 1.12 & 41.06 & 17.43 & 36.28 & 46.10 & \cellcolor{cellgray}49.70 & 57.40 & 44.28 \\
Gemma-3-it(27B)\cite{gemma_2025} & \cellcolor{cellgray}49.53 & 68.00 & \cellcolor{cellgray}37.16 & 51.80 & \cellcolor{cellgreen}\textbf{63.29} & \cellcolor{cellgray}65.81 & 36.17 & \cellcolor{cellgray}5.66 & 49.54 & \cellcolor{cellgreen}\textbf{49.31} & \cellcolor{cellgray}44.49 & 50.00 & \cellcolor{cellgray}51.70 & 57.00 & \cellcolor{cellgray}48.77 \\
Qwen3-VL(32B)\cite{qwen3technicalreport} & \cellcolor{cellgray}49.64 & \cellcolor{cellgreen}\textbf{73.80} & 35.61 & \cellcolor{cellgray}52.40 & 45.76 & 49.12 & \cellcolor{cellgray}36.34 & 1.08 & \cellcolor{cellgreen}\textbf{56.83} & \cellcolor{cellgray}46.56 & \cellcolor{cellgreen}\textbf{48.47} & \cellcolor{cellgray}54.10 & 46.50 & \cellcolor{cellgray}62.60 & 48.42 \\
InternVL3\_5(38B)\cite{wang2025internvl3_5} & 46.61 & 64.70 & 31.18 & 50.80 & 43.05 & 42.11 & 34.12 & 1.01 & 48.08 & 29.45 & 41.58 & 52.90 & 49.30 & 60.90 & 48.43 \\
Qwen2.5-VL(72B)\cite{qwen2_5} & \cellcolor{cellgreen}\textbf{54.47} & 68.90 & 35.55 & 49.90 & \cellcolor{cellgray}59.41 & \cellcolor{cellgreen}\textbf{70.67} & \cellcolor{cellgray}36.38 & \cellcolor{cellgreen}\textbf{67.26} & 50.09 & 43.49 & 35.47 & \cellcolor{cellgreen}\textbf{56.80} & \cellcolor{cellgreen}\textbf{62.00} & \cellcolor{cellgray}62.40 & \cellcolor{cellgray}48.97 \\
InternVL3(78B)\cite{chen2024internvl} & 46.93 & 68.60 & \cellcolor{cellgreen}\textbf{38.85} & \cellcolor{cellgray}53.70 & 23.91 & 20.33 & \cellcolor{cellgreen}\textbf{41.82} & 3.77 & 49.86 & 43.48 & 41.18 & 51.70 & 48.60 & 61.70 & 48.05 \\
\hline
\rowcolor{sectionblue}
\multicolumn{16}{l}{Driving-Specialized} \\
DA-DriveLM(4B)\cite{gao2024mini} & 38.31 & 66.80 & 26.13 & \textbf{61.00} & 34.99 & 23.18 & 26.80 & 0.03 & 33.23 & 14.88 & 26.11 & 49.00 & 26.50 & 52.70 & 44.06 \\
ReasonDrive(7B)\cite{chahe2025reasondrive} & 43.13 & 50.20 & 21.20 & 45.10 & 4.51 & 16.28 & 29.28 & \cellcolor{cellgray}63.24 & 10.68 & 3.65 & 9.94 & 47.60 & 28.40 & \cellcolor{cellgreen}\textbf{78.90} & 47.62 \\
\bottomrule
\end{tabular}%
}
\label{tab:table1}
\end{table*}

\paragraph{Overall Capability Coverage.}
Table~\ref{tab:table1} shows an overall ordering on the open-loop benchmark.
Qwen2.5-VL(72B) achieves the best weighted score of $54.47$, followed by Qwen3-VL(32B) and Gemma-3-it(27B) with $49.64$ and $49.53$, respectively. This advantage is not explained by scale alone. The leading models are competitive across all three ranks rather than peaking on a single subset of tasks, which aligns with the characteristic of autonomous driving that requires comprehensive capabilities.
By contrast, driving-specialized VLMs remain substantially weaker in overall coverage, indicating that current domain-specific adaptation more often strengthens local task families than it produces robust hierarchical competence, which will be further highlighted in the subsequent closed-loop testing.
Meanwhile, different models peak on different sub-capabilities, and the identity of the strongest model changes across tasks, which is particularly evident in driving-specific VLMs.

\begin{figure}[t]
    \centering
    \begin{subfigure}[b]{0.32\linewidth}
        \centering
        \includegraphics[width=\linewidth]{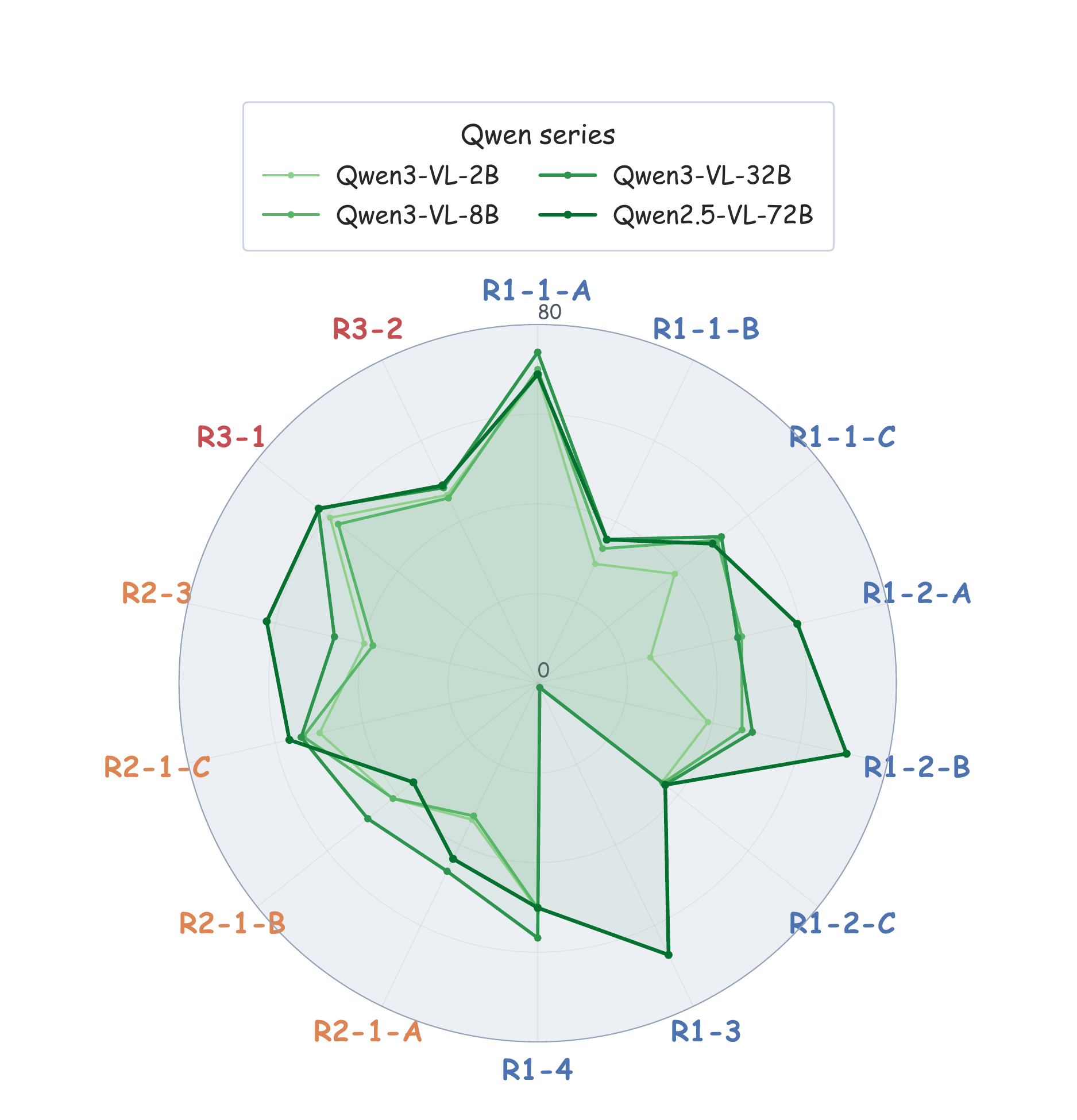}
        \caption{Qwen series.}
        \label{fig:Fig_radar_qwen}
    \end{subfigure}
    \hfill
    \begin{subfigure}[b]{0.32\linewidth}
        \centering
        \includegraphics[width=\linewidth]{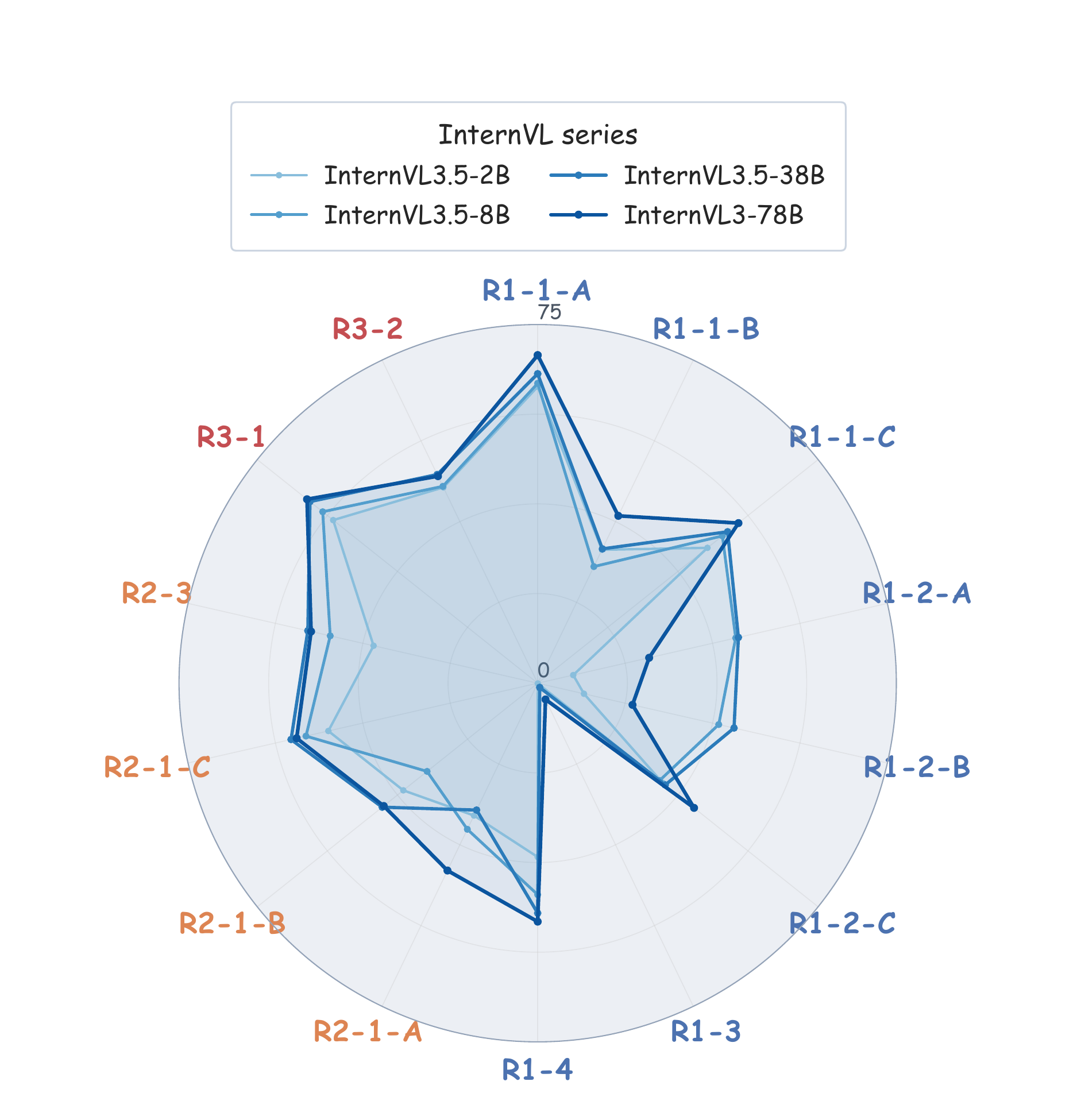}
        \caption{InternVL series.}
        \label{fig:Fig_radar_internvl}
    \end{subfigure}
    \hfill
    \begin{subfigure}[b]{0.32\linewidth}
        \centering
        \includegraphics[width=\linewidth]{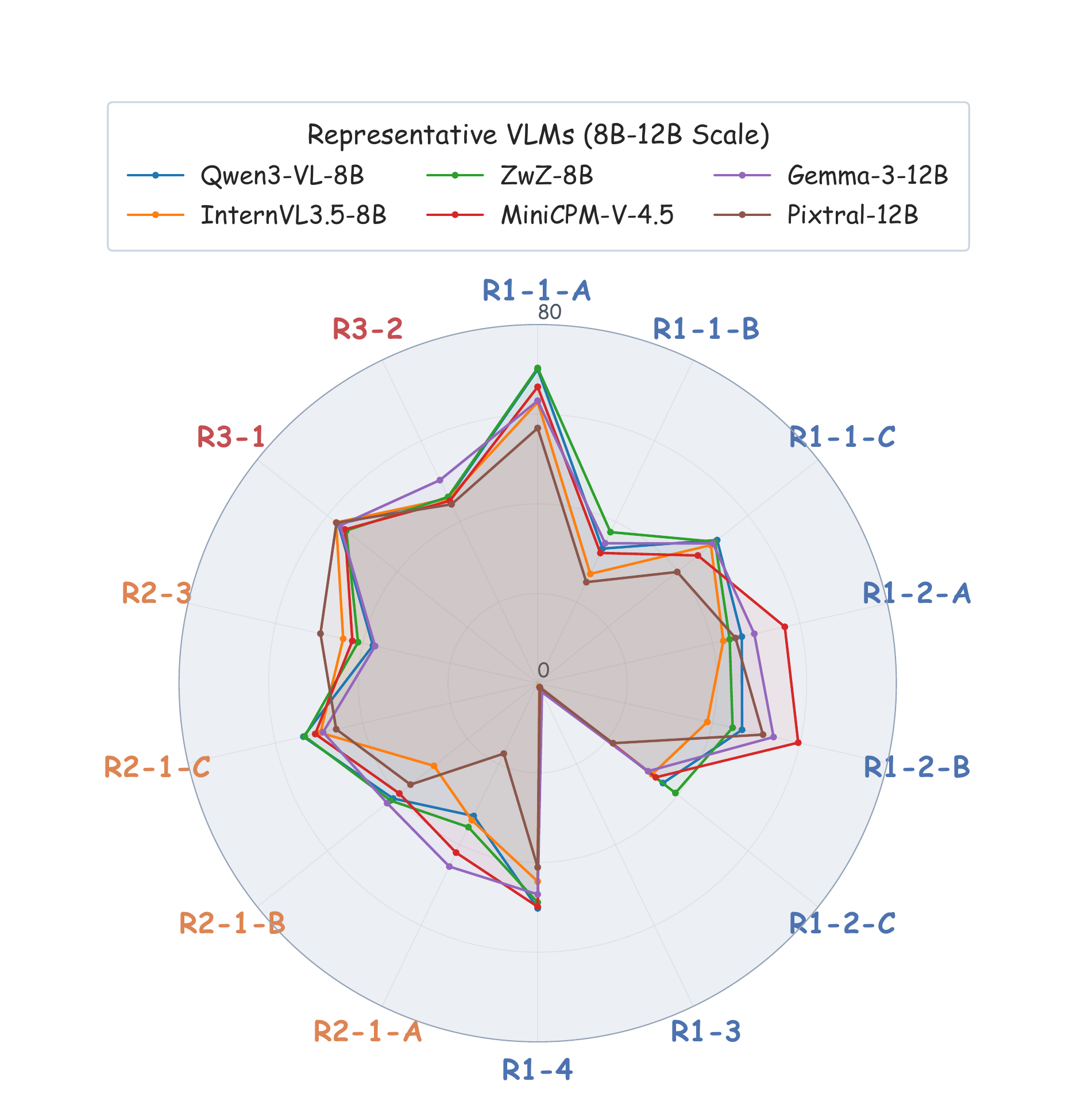}
        \caption{8B-12B model comparison.}
        \label{fig:Fig_radar_midscale_comparison}
    \end{subfigure}
    \caption{\textbf{Representative capability profiles revealed by DriveHierarchy.} The radar plots highlight both within-family scaling trends and architecture-specific trade-offs among similarly sized models.}
    \label{fig:fig_radar_profiles}
\end{figure}

\paragraph{Capability Structure Across Model Families.}
Figure~\ref{fig:fig_radar_profiles} clarifies the family-level trends that are partially hidden in the table. 
Within the Qwen series, scaling improves both average performance and profile balance, yielding progressively more complete coverage from grounding to predictive reasoning. InternVL improves more unevenly, with gains concentrated in recognition and parts of $R2$ but weaker balance on geometric grounding and localization. The 8B-12B comparison is especially informative because it removes large-scale parameter advantages and still exposes markedly different capability profiles, separating balanced models from systems with narrower strengths in distance perception, spatial relations, or temporal reasoning. 
Among similarly sized models, MiniCPM-V-4.5(9B) is strong on distance perception but weaker on temporal reasoning, Gemma-3-it(12B) is the most balanced and leads on $R3\_2$, Pixtral(12B) is relatively strong on spatial relations but weak on multi-view memory, and Qwen3-VL(8B) offers one of the best overall trade-offs at this scale.
The open-loop benchmark therefore measures structured capability variation, rather than merely reproducing a generic VLM ranking.

\subsection{Capability Correlation and Weight Allocation}

\begin{table*}
\centering
\caption{\textbf{Closed-loop benchmark results on $R4$ across 10 scenario families.} Higher scores indicate better driving performance under the composed route, safety, and efficiency metric.}
\scriptsize
\setlength{\tabcolsep}{2pt}
\renewcommand{\arraystretch}{1.05}
\begin{tabularx}{\textwidth}{@{}L{2.4cm}*{12}{Y}@{}}
\toprule
\textbf{Models} & \textbf{Sce 1} & \textbf{Sce 2} & \textbf{Sce 3} & \textbf{Sce 4} & \textbf{Sce 5} & \textbf{Sce 6} & \textbf{Sce 7} & \textbf{Sce 8} & \textbf{Sce 9} & \textbf{Sce 10} & \textbf{Avg.Score} & \textbf{Rank} \\
\hline
\rowcolor{sectionblue}
\multicolumn{13}{l}{Generalist} \\
Qwen3-VL(2B) & 1.176 & 0.062 & 0.000 & 0.686 & 0.114 & 0.299 & 0.518 & 0.620 & 0.207 & 0.084 & 0.377 & 14 \\
InternVL3\_5(2B) & 1.813 & 1.411 & 0.121 & 1.443 & 0.635 & 0.377 & 1.964 & 0.903 & 1.644 & 1.872 & 1.218 & 7 \\
Qwen3-VL(8B) & 1.278 & 1.726 & 0.384 & 1.128 & 0.574 & 0.271 & 1.379 & 0.394 & 0.661 & 1.227 & 0.902 & 9 \\
InternVL3\_5(8B) & 0.001 & 0.246 & \cellcolor{cellgreen}\textbf{0.960} & 0.939 & 0.848 & 0.642 & 1.285 & 0.662 & 0.412 & 1.025 & 0.702 & 13 \\
ZwZ(8B) & 1.147 & 1.034 & 0.415 & 1.312 & 0.471 & 0.164 & 0.773 & 0.508 & 0.948 & 1.322 & 0.809 & 10 \\
MiniCPM-V-4\_5(9B) & \cellcolor{cellgray}10.885 & \cellcolor{cellgray}8.949 & 0.099 & \cellcolor{cellgray}4.931 & 0.439 & 0.000 & \cellcolor{cellgray}17.892 & 2.281 & 3.592 & \cellcolor{cellgreen}\textbf{4.980} & \cellcolor{cellgray}5.405 & 3 \\
Gemma-3-it(12B) & 1.427 & 0.825 & \cellcolor{cellgray}0.553 & 0.926 & 0.625 & 0.431 & 0.794 & 0.364 & 0.697 & 0.548 & 0.719 & 11 \\
Pixtral(12b) & 0.940 & 1.758 & \cellcolor{cellgray}0.642 & 1.103 & 0.107 & 0.158 & 0.323 & 0.206 & 0.874 & 0.932 & 0.704 & 12 \\
Gemma-3-it(27B) & \cellcolor{cellgreen}\textbf{27.950} & \cellcolor{cellgreen}\textbf{10.269} & 0.212 & \cellcolor{cellgreen}\textbf{16.599} & \cellcolor{cellgray}2.276 & \cellcolor{cellgray}8.492 & \cellcolor{cellgreen}\textbf{43.936} & \cellcolor{cellgreen}\textbf{2.992} & \cellcolor{cellgray}9.406 & \cellcolor{cellgray}4.054 & \cellcolor{cellgreen}\textbf{12.619} & 1 \\
Qwen3-VL(32B) & 5.380 & 3.550 & 0.191 & 2.605 & \cellcolor{cellgray}2.290 & \cellcolor{cellgray}9.165 & 7.690 & \cellcolor{cellgray}2.457 & 0.487 & 0.702 & 3.452 & 5 \\
InternVL3\_5(38B) & 0.001 & 1.720 & 0.285 & 0.501 & 0.245 & 0.111 & 6.939 & 0.035 & 0.369 & 0.912 & 1.112 & 8 \\
Qwen2.5-VL(72B) & \cellcolor{cellgray}21.056 & \cellcolor{cellgray}10.123 & 0.052 & \cellcolor{cellgray}14.513 & \cellcolor{cellgreen}\textbf{2.516} & \cellcolor{cellgreen}\textbf{9.325} & 13.508 & \cellcolor{cellgray}2.478 & \cellcolor{cellgreen}\textbf{16.237} & \cellcolor{cellgray}3.849 & \cellcolor{cellgray}9.366 & 2 \\
InternVL3(78B) & 8.145 & 2.468 & 0.063 & 2.317 & 0.524 & 0.202 & 6.132 & 2.077 & 6.500 & 2.988 & 3.142 & 6 \\
\hline
\rowcolor{sectionblue}
\multicolumn{13}{l}{Driving-Specialized} \\
DA-DriveLM(4B) & 0.001 & 0.000 & 0.023 & 0.008 & 0.008 & 0.005 & 0.041 & 0.012 & 0.024 & 0.044 & 0.017 & 15 \\
ReasonDrive(7B) & 1.814 & 1.912 & 0.388 & 2.030 & 0.948 & 0.188 & \cellcolor{cellgray}22.411 & 1.747 & \cellcolor{cellgray}8.590 & 1.983 & 4.201 & 4 \\
\bottomrule
\end{tabularx}
\label{tab:table2}
\end{table*}

\begin{figure}
    \centering
    \begin{subfigure}[b]{0.45\linewidth}
        \centering
        \includegraphics[width=\linewidth]{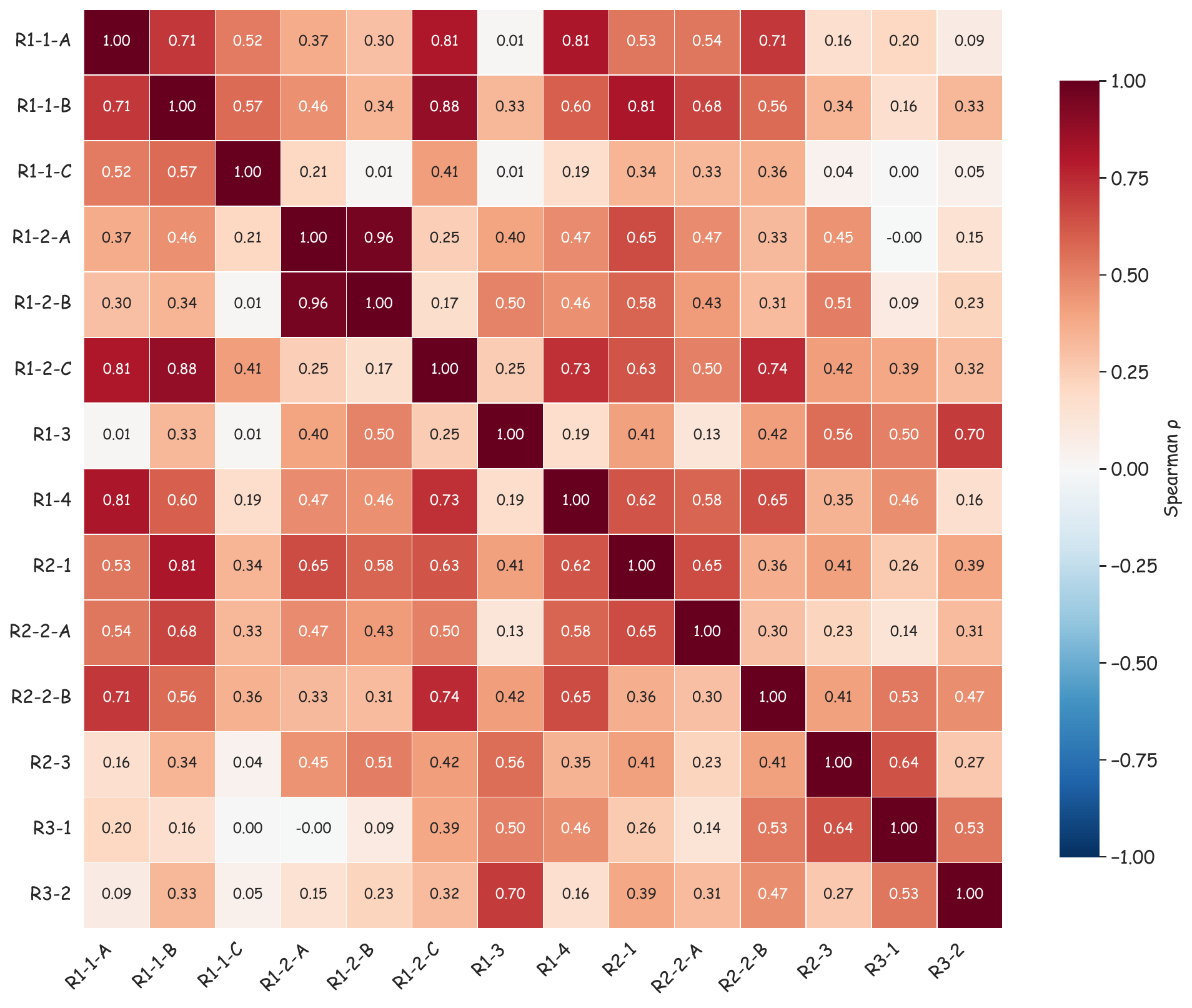}
        \caption{Subtask-level Spearman correlation matrix.}
        \label{fig:heatmap_Correlations_Among_Sub_abilities}
    \end{subfigure}
    \hfill
    \begin{subfigure}[b]{0.45\linewidth}
        \centering
        \includegraphics[width=\linewidth]{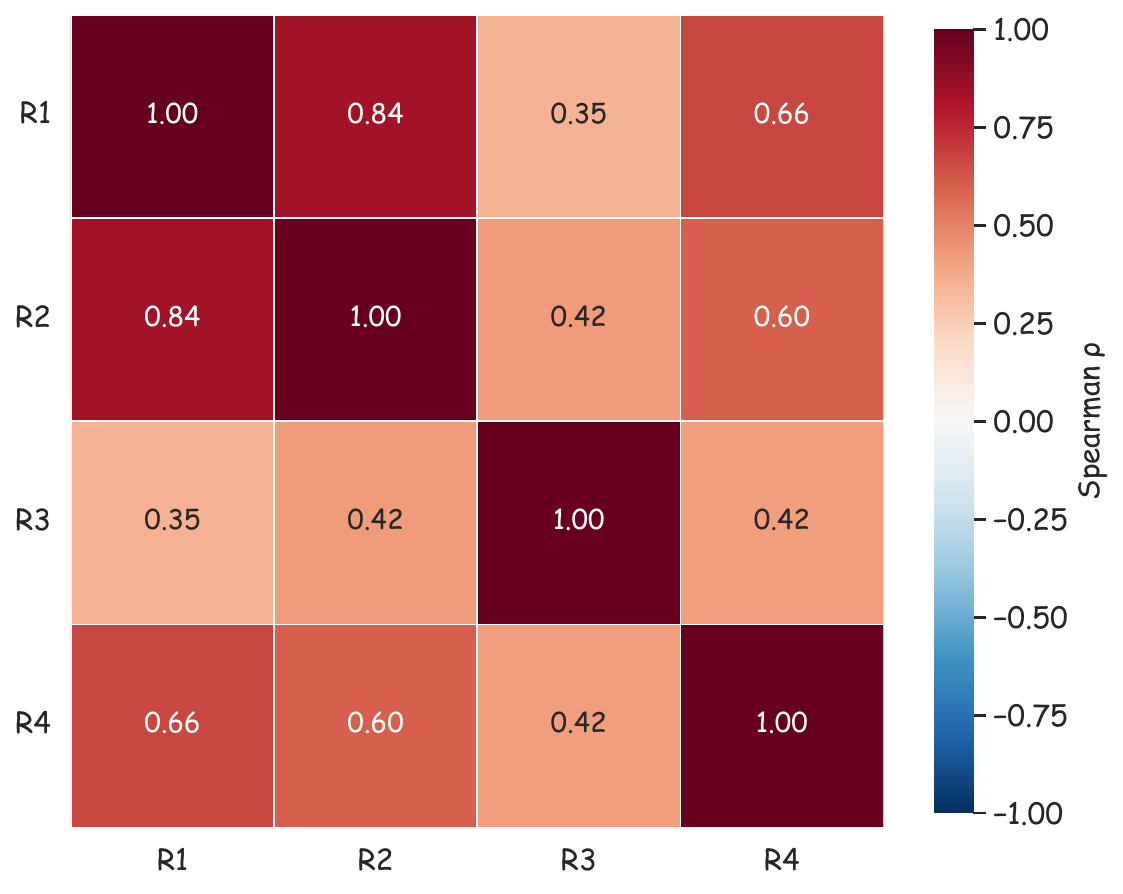}
        \caption{Rank-level Spearman correlation matrix.}
        \label{fig:level_heatmap}
    \end{subfigure}
    \caption{\textbf{Correlation structure of DriveHierarchy across subtasks and ranks,} which shows that the benchmark contains both locally coupled abilities and globally progressive capability levels.}
    \label{fig:two_heatmaps}
\end{figure}

\paragraph{Empirical Correlation Structure.}
Figure~\ref{fig:heatmap_Correlations_Among_Sub_abilities} shows that \textsc{DriveHierarchy} exhibits structured, but non-collapsed, capability dependencies. At the subtask level, semantically adjacent abilities form clear local clusters: nearest-object and referred-object distance estimation are aligned ($\rho=0.961$), and counting-oriented recognition is strongly coupled with distance-bucket estimation ($\rho=0.875$). In contrast, state recognition and outcome prediction ($\rho=0.004$), as well as localization and state recognition ($\rho=0.011$), are nearly independent, which captures both shared structure and genuine capability separation, rather than reducing to either a set of unrelated tasks or a single latent score. The rank-level correlations reinforce this interpretation: $R1$ and $R2$ are strongly associated ($\rho=0.843$), whereas their relations to $R3$ are substantially weaker ($\rho=0.346$ and $\rho=0.425$), indicating that higher-order reasoning builds on lower-level competence without being reducible to it.

\paragraph{Correlation-Informed Weight Allocation.}
For the released benchmark, we adopt a correlation-aware weighting scheme, with the full methodology and implementation details given in Appendix~\ref{appendix:weight_allocation}.
We first fix the rank budgets to $R1{:}R2{:}R3=3{:}3{:}4$, and then allocate each budget according to subtask uniqueness derived from the correlation matrix.
These correlations are estimated once during an initial calibration phase after the first benchmark-testing round, and the resulting subtask weights are then frozen for all subsequent evaluations.
Highly redundant subtasks therefore receive smaller fixed weights, while subtasks carrying more independent signal receive larger fixed shares. This logic is visible in Table~\ref{tab:table1}: within $R1$, the less redundant $R1\_1\_C$ and $R1\_3$ are weighted more heavily than the tightly coupled counting and distance variants, and within $R2$, the relational task $R2\_3$ receives the largest share.
The resulting composite score is therefore designed to preserve distinct capability dimensions rather than repeatedly rewarding the same underlying competence.
Meanwhile, task-level and rank-level profiles remain the primary diagnostic outputs, especially when comparing models whose weighted averages are close.

\subsection{From Open-Loop Capability to Closed-loop Driving}

Table~\ref{tab:table2} reports the Closed-loop results on $R4$, showing that Closed-loop driving is meaningfully related to open-loop capability, but is not determined by it. At the model level, the strongest $R4$ systems, such as Gemma-3-it(27B) and Qwen2.5-VL(72B), also belong to the strongest open-loop models, indicating that embodied driving performance remains grounded in the same capability spectrum measured by $R1$-$R3$. 
This interpretation is consistent with the rank-level correlations shown in Figure~\ref{fig:level_heatmap}, where the Spearman coefficients with $R4$ are $0.664$ for $R1$, $0.596$ for $R2$, and $0.418$ for $R3$. 
These values establish that stronger open-loop capability is generally associated with stronger closed-loop driving, while also showing that the association weakens as one moves from grounding and scene integration to the current form of higher-level reasoning.
This indicates that open-loop evaluation is behaviorally valid but not behaviorally sufficient. For current VLM-based driving systems, lower- and mid-level abilities remain more predictive of closed-loop success than the present $R3$ tasks, suggesting that robust grounding and scene integration are still the dominant bottlenecks for embodied performance.
From the perspective of training optimization, $R1$-$R3$ provide a principled basis for selecting supervision targets to improve closed-loop driving.

\subsection{Benchmark-Guided SFT}
To evaluate whether DriveHierarchy can guide model improvement, we adopt Qwen3-VL(8B) as the base model and derive SFT targets from the task groups with the largest deficits relative to task-wise SOTA performance\cite{zhao2024swiftascalablelightweightinfrastructure}.
Rather than emphasizing the expected gains on directly supervised tasks, we focus on transfer beyond the fine-tuning scope, asking whether capability-targeted supervision leads to measurable improvements on non-target question types and downstream closed-loop driving.
The first row of Table~\ref{tab:table3} reports this deficit profile solely as a reference for target selection.

\begin{table*}
\centering
\caption{\textbf{Open-loop transfer results of benchmark-guided SFT on Qwen3-VL(8B).} The first row reports the baseline deficit relative to the task-wise SOTA result and is used only for target selection.}
\setlength{\tabcolsep}{2pt}
\renewcommand{\arraystretch}{1.15}
\resizebox{\textwidth}{!}{%
\begin{tabular}{l*{15}{c}}
\toprule
\multirow{3}{*}{\textbf{Models}} & \multirow{3}{*}{\textbf{Avg.Score}} & \multicolumn{8}{c}{\textbf{R1 Perceptual Grounding}} & \multicolumn{4}{c}{\textbf{R2 Contextual Memory}} & \multicolumn{2}{c}{\textbf{R3 Mental Reasoning}} \\
\cline{3-16}
& & R1\_1\_A & R1\_1\_B & R1\_1\_C & R1\_2\_A & R1\_2\_B & R1\_2\_C & R1\_3 & R1\_4 & R2\_1 & R2\_2\_A & R2\_2\_B & R2\_3 & R3\_1 & R3\_2 \\
\cline{3-16}
& & 0.03 & 0.027 & 0.056 & 0.035 & 0.036 & 0.029 & 0.056 & 0.031 & 0.064 & 0.079 & 0.067 & 0.09 & 0.233 & 0.167 \\
\hline
Qwen3-VL(8B) & \textcolor{deltagray}{-0.17} & \textcolor{deltagray}{-0.05} & \textcolor{deltagray}{-0.14} & \textcolor{deltagray}{-0.16} & \textcolor{deltagray}{-0.26} & \textcolor{deltagray}{-0.34} & \textcolor{deltagray}{-0.15} & \textcolor{deltagray}{-0.98} & \textcolor{deltagray}{-0.12} & \textcolor{deltagray}{-0.33} & \textcolor{deltagray}{-0.15} & \textcolor{deltagray}{-0.06} & \textcolor{deltagray}{-0.39} & \textcolor{deltagray}{-0.28} & \textcolor{deltagray}{-0.09} \\
\rowcolor{sectionblue}
Baseline* & 45.10 & 70.00 & 33.32 & 51.20 & 46.70 & 46.76 & 35.74 & 1.05 & 50.23 & 32.87 & 41.25 & 53.60 & 37.70 & 56.90 & 45.84 \\
B+R1\_2\_A & \textcolor{deltagreen}{+2.66} & \textcolor{deltagreen}{+7.70} & \textcolor{deltagreen}{+12.78} & \textcolor{deltared}{-1.20} & \cellcolor{ablationblue}\textcolor{deltagreen}{+46.54} & \textcolor{deltagreen}{+14.71} & \textcolor{deltagreen}{+5.86} & \textcolor{deltagray}{0.00} & \textcolor{deltagreen}{+1.06} & \textcolor{deltagray}{+0.17} & \textcolor{deltagreen}{+5.49} & \textcolor{deltared}{-1.00} & \textcolor{deltared}{-1.20} & \textcolor{deltared}{-2.30} & \textcolor{deltagray}{+0.37} \\
B+R1\_2\_B & \textcolor{deltagreen}{+3.57} & \textcolor{deltagreen}{+5.50} & \textcolor{deltagreen}{+10.91} & \textcolor{deltagreen}{+1.80} & \textcolor{deltagreen}{+35.78} & \cellcolor{ablationblue}\textcolor{deltagreen}{+47.82} & \textcolor{deltagreen}{+4.91} & \textcolor{deltagray}{+0.01} & \textcolor{deltared}{-2.09} & \textcolor{deltagreen}{+1.99} & \textcolor{deltagreen}{+3.03} & \textcolor{deltagreen}{+1.10} & \textcolor{deltared}{-1.30} & \textcolor{deltared}{-1.80} & \textcolor{deltagray}{+0.32} \\
B+R1\_3 & \textcolor{deltagreen}{+2.77} & \textcolor{deltared}{-3.70} & \textcolor{deltagreen}{+1.50} & \textcolor{deltared}{-2.40} & \textcolor{deltared}{-27.97} & \textcolor{deltared}{-3.23} & \textcolor{deltagreen}{+3.56} & \cellcolor{ablationblue}\textcolor{deltagreen}{+79.45} & \textcolor{deltagreen}{+1.07} & \textcolor{deltared}{-4.99} & \textcolor{deltared}{-2.58} & \textcolor{deltared}{-1.10} & \textcolor{deltared}{-3.60} & \textcolor{deltagreen}{+1.50} & \textcolor{deltagray}{+0.37} \\
\makecell[l]{B+R1\_2\_A+\\R1\_2\_B+R1\_3} & \textcolor{deltagreen}{+2.14} & \textcolor{deltagreen}{+4.90} & \textcolor{deltagreen}{+7.35} & \textcolor{deltared}{-4.60} & \cellcolor{ablationblue}\textcolor{deltagreen}{+45.49} & \cellcolor{ablationblue}\textcolor{deltagreen}{+46.73} & \textcolor{deltagreen}{+3.91} & \cellcolor{ablationblue}\textcolor{deltagray}{+0.01} & \textcolor{deltared}{-4.40} & \textcolor{deltared}{-7.12} & \textcolor{deltagreen}{+2.13} & \textcolor{deltagray}{+0.80} & \textcolor{deltared}{-9.30} & \textcolor{deltagray}{-0.80} & \textcolor{deltagray}{+0.35} \\
B+R2\_1 & \textcolor{deltagreen}{+4.05} & \textcolor{deltagray}{+0.30} & \textcolor{deltagreen}{+7.02} & \textcolor{deltagray}{+0.90} & \textcolor{deltagreen}{+1.39} & \textcolor{deltagray}{-0.72} & \textcolor{deltagreen}{+6.74} & \textcolor{deltagray}{+0.01} & \textcolor{deltared}{-5.41} & \cellcolor{ablationblue}\textcolor{deltagreen}{+43.71} & \textcolor{deltagreen}{+5.67} & \textcolor{deltagray}{0.00} & \textcolor{deltagreen}{+2.30} & \textcolor{deltagray}{+0.50} & \textcolor{deltagreen}{+1.10} \\
B+R2\_3 & \textcolor{deltagray}{+0.53} & \textcolor{deltagreen}{+4.70} & \textcolor{deltagreen}{+2.21} & \textcolor{deltagray}{-0.60} & \textcolor{deltagreen}{+11.17} & \textcolor{deltagreen}{+3.14} & \textcolor{deltagreen}{+1.30} & \textcolor{deltagray}{+0.01} & \textcolor{deltared}{-8.93} & \textcolor{deltared}{-10.01} & \textcolor{deltared}{-3.32} & \textcolor{deltared}{-7.20} & \cellcolor{ablationblue}\textcolor{deltagreen}{+61.20} & \textcolor{deltared}{-17.70} & \textcolor{deltagray}{+0.59} \\
B+R2\_1+R2\_3 & \textcolor{deltagray}{+0.42} & \textcolor{deltagreen}{+3.00} & \textcolor{deltagreen}{+2.21} & \textcolor{deltagray}{+0.20} & \textcolor{deltagreen}{+8.85} & \textcolor{deltagreen}{+4.31} & \textcolor{deltagray}{-0.51} & \textcolor{deltagray}{0.00} & \textcolor{deltared}{-8.06} & \cellcolor{ablationblue}\textcolor{deltagray}{-0.28} & \textcolor{deltared}{-2.69} & \textcolor{deltared}{-9.50} & \cellcolor{ablationblue}\textcolor{deltagreen}{+61.10} & \textcolor{deltared}{-19.90} & \textcolor{deltagray}{+0.40} \\
B+R3\_1 & \textcolor{deltagreen}{+7.12} & \textcolor{deltared}{-5.60} & \textcolor{deltared}{-1.33} & \textcolor{deltared}{-5.00} & \textcolor{deltared}{-3.26} & \textcolor{deltared}{-6.67} & \textcolor{deltagreen}{+3.06} & \textcolor{deltagray}{-0.01} & \textcolor{deltared}{-5.70} & \textcolor{deltared}{-9.05} & \textcolor{deltared}{-4.13} & \textcolor{deltared}{-1.00} & \textcolor{deltared}{-5.40} & \cellcolor{ablationblue}\textcolor{deltagreen}{+40.50} & \textcolor{deltagray}{+0.45} \\
B+ALL & \textcolor{deltagreen}{+20.88} & \textcolor{deltagreen}{+6.70} & \textcolor{deltagreen}{+12.35} & \textcolor{deltared}{-6.70} & \cellcolor{ablationblue}\textcolor{deltagreen}{+44.33} & \cellcolor{ablationblue}\textcolor{deltagreen}{+46.69} & \textcolor{deltagreen}{+3.14} & \cellcolor{ablationblue}\textcolor{deltagreen}{+79.74} & \textcolor{deltared}{-2.73} & \textcolor{deltagray}{-0.87} & \textcolor{deltagreen}{+1.65} & \textcolor{deltagray}{-0.50} & \cellcolor{ablationblue}\textcolor{deltagreen}{+57.40} & \cellcolor{ablationblue}\textcolor{deltagreen}{+33.20} & \textcolor{deltagray}{+0.46} \\
\bottomrule
\end{tabular}%
}
\label{tab:table3}
\end{table*}

\begin{table*}
\centering
\caption{\textbf{Closed-loop transfer of benchmark-guided supervised fine-tuning on $R4$.} Each intervention is defined by capability groups selected from DriveHierarchy.}
\scriptsize
\setlength{\tabcolsep}{2pt}
\renewcommand{\arraystretch}{1.05}
\rowcolors{2}{white}{gray!10}

\begin{tabularx}{\textwidth}{@{}l*{11}{Y}@{}}
\toprule
\textbf{Models} & \textbf{Sce 1} & \textbf{Sce 2} & \textbf{Sce 3} & \textbf{Sce 4} & \textbf{Sce 5} & \textbf{Sce 6} & \textbf{Sce 7} & \textbf{Sce 8} & \textbf{Sce 9} & \textbf{Sce 10} & \textbf{Avg.Score} \\
\midrule
Baseline & 1.278 & 1.726 & 0.384 & 1.128 & 0.574 & 0.271 & 1.379 & 0.394 & 0.661 & 1.227 & 0.902 \\
\makecell[l]{B+R1\_2\_A+\\R1\_2\_B+R1\_3} & 9.436 & \textbf{4.788} & \textbf{0.816} & 3.199 & \textbf{2.418} & 8.468 & 5.048 & 1.924 & 9.958 & \textbf{3.379} & 4.943 \\
B+R2\_1+R2\_3 & 15.181 & 3.279 & 0.207 & 4.095 & 1.778 & 8.579 & 4.299 & 2.176 & 9.924 & 2.192 & 5.171 \\
B+R3\_1 & 2.453 & 3.512 & 0.122 & 2.657 & 1.762 & \textbf{17.400} & 4.176 & \textbf{2.499} & 9.477 & 1.892 & 4.595 \\
\textbf{B+ALL} & \textbf{22.585} & 3.146 & 0.734 & \textbf{15.888} & 2.144 & 3.694 & \textbf{11.855} & 2.340 & \textbf{16.717} & 1.105 & \textbf{8.021} \\
\bottomrule
\end{tabularx}

\label{tab:table4}
\end{table*}

\paragraph{Single-Capability Adaptation and Cross-Ability Transfer.}
Table~\ref{tab:table3} shows that benchmark-guided SFT produces structured transfer rather than merely improving the directly supervised task. 
While in-scope gains are expected by construction, the more informative signal is the response of non-target question types outside the selected supervision group.
For example, after supervising $R1\_2\_B$, 10 of the 13 non-target abilities increase; the strongest spillovers are $R1\_2\_A$ ($+35.78$), $R1\_1\_B$ ($+10.91$), and $R2\_2\_A$ ($+3.03$), whereas the largest non-target decrease is only $-2.09$. Similar transfer patterns also appear under other interventions.
Even the more localized interventions do not produce a rank-wide collapse. For example, supervising $R3\_1$ still raises $R1\_2\_C$ by $+3.06$ and $R3\_2$ by $+0.45$, while supervising $R1\_3$ raises $R1\_1\_B$ and $R3\_1$ by $+1.50$ each. 
Taken together, these results support the interpretation that the capabilities are sufficiently differentiated for targeted intervention that optimizing one does not induce systematic degradation elsewhere, yet sufficiently coupled that targeted supervision often produces positive cross-ability spillover, which can provide guidance for training optimization, showing that the hierarchy can be used to probe and compare cross-capability transfer rather than merely to report in-scope task repair.

\paragraph{From Open-Loop Improvement to Closed-loop Gain.}
These benchmark-guided interventions transfer to embodied driving.
Importantly, $R4$ is never used as a supervision target in these experiments and all reported closed-loop changes therefore reflect transfer from open-loop capability supervision. 
Table~\ref{tab:table4} shows that every selected intervention yields a substantial gain on $R4$ over the baseline score of $0.902$, and that the magnitude of the gain depends on which capability group is optimized. Strengthening the selected $R2$ capabilities raises the closed-loop score to $5.171$ ($+4.269$), slightly exceeding the gain from the selected $R1$ group, which reaches $4.943$ ($+4.041$), while direct supervision on $R3\_1$ improves $R4$ to $4.595$ ($+3.693$). Joint optimization over the benchmark-identified weak set produces the largest effect, increasing the closed-loop score to $8.021$ ($+7.119$).
These results suggest that the open-loop capability deficits identified by DriveHierarchy are actionable optimization targets for improving closed-loop driving rather than merely descriptive diagnostics.
These results indicate that the hierarchy can help select open-loop supervision targets whose effects propagate to downstream closed-loop behavior, rather than treating $R4$ as an opaque end metric.

\section{Conclusion}
This paper presented \textsc{DriveHierarchy}, a hierarchical benchmark for VLM-based autonomous driving that organizes evaluation from perceptual grounding to closed-loop execution and shifts driving evaluation from flat model comparison toward capability-structured diagnosis.
Its central value lies not only in measuring performance, but also in making the internal organization of driving ability explicit and relating open-loop competence to downstream behavior, with a targeted SFT case study further illustrating how such diagnoses may inform model intervention.
We hope this benchmark can serve as a practical framework for understanding and advancing VLM-based autonomous driving systems and for studying how open-loop capabilities relate to closed-loop performance.

\begin{ack}
This work was supported by National Natural Science Foundation of China (52125208).
\end{ack}

{
\small

\bibliographystyle{unsrt}
\bibliography{related}

}


\newpage
\appendix

\section{Limitations, Broader impacts and Future work}
\label{Limitations}
\paragraph{Limitations.}
\textsc{DriveHierarchy} still has several important limitations. Firstly, although $R4$ is constructed on a real-road-layout co-simulation platform, it remains a simulator-based closed-loop benchmark and therefore cannot be interpreted as a direct proxy for real-world deployment performance.
Secondly, the current hierarchy is broad but not exhaustive. For example, higher-level reasoning is represented by a limited number of task families, leaving some capabilities under-covered.
Thirdly, the composite open-loop score depends on a correlation-aware weighting scheme that is more principled than uniform averaging but still reflects a modeling choice rather than a uniquely determined ground truth.
Fourth, given that repeated evaluations are computationally expensive, the current experiments report only deterministic single-run scores. Consequently, minor differences between models should be interpreted with caution.
Finally, the benchmark-guided optimization results, while informative, are established on a single base model and should therefore be interpreted as evidence of practical utility rather than as a universal characterization of capability transfer.

\paragraph{Broader Impacts.}
The primary positive impact of \textsc{DriveHierarchy} is to make evaluation of VLM-based autonomous driving more diagnostic, interpretable, and actionable by exposing the internal capability structure that underlies downstream driving behavior. This may help researchers move beyond flat success metrics, identify concrete failure modes, and design more targeted optimization strategies. At the same time, the benchmark should be used with appropriate caution. Strong performance on \textsc{DriveHierarchy}, including strong $R4$ results, does not imply readiness for real-world autonomous driving deployment, and benchmark-oriented optimization may encourage overfitting to the task distributions, annotations, or scenario designs represented here. More broadly, because the benchmark is built from public datasets and simulation infrastructure, any omissions, biases, or simplifications in those resources may propagate into model development and evaluation. \textsc{DriveHierarchy} should therefore be treated as a research instrument for capability analysis and controlled comparison, rather than as a certification standard for safety-critical deployment.

\paragraph{Future Work.}
\label{Future_work}
Several works are going to be conducted following this work. 
At the platform level, the current interactive scene editor for $R4$ shown in Fig~\ref{fig:Fig_R4_platform} remains at an early stage, although it already supports integration with hardware components and real test vehicles\cite{cui2025vp}; an important next step is therefore to mature this infrastructure and optimize it specifically for end-to-end autonomous driving. Building on this foundation, we plan to refine the virtual-real fusion testing system that couples controllable simulation with physical testing and larger-scale replay- or vehicle-based validation, thereby strengthening the connection between benchmark performance and real-world driving behavior.
At the methodology level, the weighting and aggregation scheme can be further strengthened through robustness analysis, alternative composite-score constructions, and uncertainty estimates over repeated scenario executions; the generality of benchmark-guided intervention should also be examined across more model families, training recipes, and intervention strategies.
More broadly, \textsc{DriveHierarchy} is positioned as a dynamic diagnostic and analytical framework, rather than merely a static leaderboard.
We plan to maintain and extend it over time so that it supports sustained analysis of capability structure instead of becoming a narrow optimization target.

\section{Benchmark Construction Details}
\label{appendix:data_sources}
The open-loop component of \textsc{DriveHierarchy} is constructed by integrating multiple public autonomous driving datasets, including NuScenes\cite{caesar2020nuscenes}, NuPlan\cite{karnchanachari2024towards}, NAVSIM\cite{Dauner2024NEURIPS}, DriveLM\cite{sima2024drivelm}, DriveBench\cite{xie2025vlms}, LingoQA\cite{marcu2024lingoqa}, and DRAMA\cite{yuan2024drama}, and converting them into structured multimodal evaluation items aligned with the capability hierarchy. The resulting corpus contains 76,798 question-answer pairs over 84,279 frames, spanning recognition, grounding, memory, reasoning, and planning. From this corpus, we further derive a refined evaluation set containing 14,000 question-answer pairs over 31,945 frames to enable more compact and controlled benchmarking. The closed-loop component is built on the interactive CARLA-SUMO co-simulation platform and contains 100 curated driving scenarios.

\begin{table*}[htp]
\centering
\caption{\textbf{Comparison with representative autonomous-driving benchmarks.} ``Hier.'' indicates whether the benchmark explicitly organizes evaluation into a capability hierarchy. ``Diag.'' indicates whether it exposes capability-level diagnostic outputs rather than only task or route scores.}
\scriptsize
\setlength{\tabcolsep}{3pt}
\renewcommand{\arraystretch}{1.08}
\begin{tabularx}{\textwidth}{lcccccY}
\toprule
\textbf{Benchmark} & \textbf{Open QA} & \textbf{Closed loop} & \textbf{VLM iface.} & \textbf{Hier.} & \textbf{Diag.} & \textbf{Primary focus} \\
\midrule
DriveLM~\cite{sima2024drivelm}              & \cmark & \xmark & \cmark & \xmark & \pmark & Language-grounded driving QA \\
LingoQA~\cite{marcu2024lingoqa}             & \cmark & \xmark & \cmark & \xmark & \xmark & Language-grounded driving QA \\
DriveBench~\cite{xie2025vlms}               & \cmark & \xmark & \cmark & \xmark & \pmark & The Visual Robustness of VLMs \\
NAVSIM~\cite{dauner2024navsim}              & \xmark & \pmark & \xmark & \xmark & \xmark & Planning-oriented simulation \\
SafeBench~\cite{xu2022safebench}            & \xmark & \cmark & \xmark & \xmark & \pmark & Safety-critical scenarios \\
Bench2Drive~\cite{jia2024bench2drive}       & \xmark & \cmark & \xmark & \xmark & \pmark & Closed-loop driving agents \\
Bench2Drive-VL~\cite{jia2026bench2drive}  & \cmark & \cmark & \cmark & \xmark & \pmark & Closed-loop VLM driving agents \\
\textsc{DriveHierarchy}                     & \cmark & \cmark & \cmark & \cmark & \cmark & Capability-structured diagnosis \\
\bottomrule
\end{tabularx}
\label{tab:related_benchmark_comparison}
\end{table*}

\begin{table*}[htp]
\centering
\caption{\textbf{Construction-quality controls for the refined open-loop set.} The released manifest stores per-item source identifiers and answer labels so that users can audit source proportions and answer distributions for each task.}
\scriptsize
\setlength{\tabcolsep}{3pt}
\renewcommand{\arraystretch}{1.05}
\begin{tabularx}{\textwidth}{l X X}
\toprule
\textbf{Task group} & \textbf{Filtering and balance controls} & \textbf{Validation signal} \\
\midrule
Recognition/counting & visibility projection checks, category vocabulary normalization, scene-level quotas & nonnegative labels, valid category IDs, duplicate-item checks \\
Distance/grounding & calibrated ego-object geometry, projected-coordinate validity, invalid-box removal & finite distances, in-frame boxes, IoU-compatible box format \\
Situation description & LingoQA-context filtering for current-scene and ego-action semantics & learned-judge compatibility and prompt/answer schema checks \\
Multi-view/temporal memory & overlap trigger, fixed clip windows, instance-ID deduplication, transition balancing & valid synchronized views, valid temporal order, instance continuity checks \\
Relation/reasoning/planning & minimum visible-object constraints, randomized choices, nontrivial motion clips & answer-choice consistency, source-scene sanity checks, order-label validation \\
\bottomrule
\end{tabularx}
\label{tab:data_quality_controls}
\end{table*}

\paragraph{Artifact Availability.}
For anonymous review, the project release at \url{anonymous.4open.science/r/DriveHierarchy-2FD5} separates redistributable benchmark assets from upstream resources that must be obtained under their original licenses.
The release includes benchmark metadata, task annotations permitted for redistribution, prompt templates, scoring code, closed-loop scenario configurations, reproduction scripts, and component-level logs for route, safety, efficiency, and termination status.
For large or license-restricted resources, the release provides inspectable samples and reconstruction scripts rather than redistributing raw upstream data.

\paragraph{Reproduction Contract.}
Re-running the full benchmark requires obtaining the upstream datasets and model checkpoints listed in Table~\ref{tab:public_resources}; the release provides reconstruction scripts, split manifests, prompt templates, parser/scoring code, and scenario configurations. The package pins the exact software environment in environment files, including Python, CUDA, vLLM, SWIFT, CARLA, SUMO, and benchmark-code versions, and provides commands for (i) rebuilding the refined $R1$-$R3$ manifests, (ii) recomputing open-loop scores, (iii) replaying $R4$ scenarios, and (iv) aggregating route, safety, efficiency, termination, and parse-status logs. The released annotations and code use the project license specified in the repository, while raw upstream assets remain governed by their original licenses and terms.

\section{Experiment details and reproducibility}
\label{Experiment_details_and_reproducibility}
\subsection{Model Configurations}
Table~\ref{tab:model_configurations} lists the exact checkpoints used in the open-loop evaluation. Unless otherwise stated, all models are loaded from their official open-source releases, and the short names reported in the main paper are aliases for the full checkpoints shown here. The reported GPU counts correspond to the tensor-parallel configuration used during inference on NVIDIA A800 GPUs.

\begin{table}[htp]
    \centering
    \caption{\textbf{Model configurations used in the open-loop evaluation.}}
    \begin{tabular}{l|c c c}
    \toprule
    \textbf{Models} & \textbf{Provider} & \textbf{Full name} & \textbf{Inference Hardware} \\
    \midrule
        Qwen3-VL(2B) &  Qwen & Qwen3-VL-2B-Instruct & 1 $\times$ A800\\
        InternVL3\_5(2B)& OpenGVLab & InternVL3\_5-2B-Instruct & 1 $\times$ A800 \\
        Qwen3-VL(8B) &  Qwen & Qwen3-VL-8B-Instruct & 1 $\times$ A800\\
        InternVL3\_5(8B)& OpenGVLab & InternVL3\_5-8B-Instruct & 1 $\times$ A800 \\
        ZwZ(8B) & inclusionAI & ZwZ-8B & 1 $\times$ A800 \\
        MiniCPM-V-4\_5(9B) & openbmb & MiniCPM-V-4\_5 & 1 $\times$ A800 \\
        Gemma-3-it(12B) & google & Gemma-3-12b-it & 1 $\times$ A800 \\
        Pixtral(12B) & mistralai & Pixtral-12B-2409 & 1 $\times$ A800 \\
        Gemma3-it(27B) & google & Gemma-3-27b-it & 2 $\times$ A800 \\
        Qwen3-VL(32B) &  Qwen & Qwen3-VL-32B-Instruct & 2 $\times$ A800\\
        InternVL3\_5(38B)& OpenGVLab & InternVL3\_5-38B-Instruct & 2 $\times$ A800 \\
        Qwen2.5-VL(72B) & Qwen & Qwen2.5-VL-72B-Instruct & 4 $\times$ A800 \\
        InternVL3(78B) & OpenGVLab & InternVL3-78B & 4 $\times$ A800 \\
        DA-DriveLM(4B) & OpenGVLab & Mini-InternVL2-4B-DA-DriveLM & 1 $\times$ A800 \\
        ReasonDrive(7B) & ac4462 & Qwen2.5-VL-7B-DriveLM & 1 $\times$ A800 \\
    \bottomrule
    \end{tabular}
    \label{tab:model_configurations}
\end{table}

\subsection{Inference and Evaluation Details}
Each sample is instantiated as a multimodal chat input composed of a fixed system prompt and a task-specific user prompt. Depending on the task, the visual input may consist of a single image, a synchronized multi-view set, or a short temporal sequence. To keep evaluation deterministic and parsing-stable across heterogeneous task types, the output format is constrained at the task level: binary tasks require a yes/no response, numeric tasks require a bare number, multiple-choice tasks require only the selected option label, grounding tasks require a bounding box, and sequence-ordering tasks require an ordered list of labels. Multi-image inputs are presented in a fixed order, and temporal-order tasks are explicitly indexed so that the model receives a consistent visual-textual interface across all $R1$-$R3$ evaluations.

Inference is conducted under a unified deterministic decoding protocol. The principal settings are summarized in Table~\ref{tab:open_loop_inference_settings}.

\begin{table}[htp]
    \centering
    \caption{\textbf{Open-loop inference settings.}}
    \begin{tabular}{l l c}
        \toprule
        \textbf{Symbol} & \textbf{Meaning} & \textbf{Value} \\
        \midrule
        $T$ & Decoding temperature & $0.0$ \\
        $p$ & Nucleus sampling threshold & $1.0$ \\
        $N_{\mathrm{gen}}$ & Maximum generated tokens & $256$ \\
        $\lambda_{\mathrm{rep}}$ & Repetition control coefficient & $1.0$ \\
        $\lambda_{\mathrm{pre}}$ & Presence penalty coefficient & $0.0$ \\
        $B$ & Number of samples per inference batch & $2$ \\
        $s$ & Random seed & $0$ \\
        $\eta_{\mathrm{mem}}$ & Fraction of GPU memory reserved for serving & $0.90$ \\
        $L_{\mathrm{ctx}}$ & Maximum model context length & $16384$ \\
        \bottomrule
    \end{tabular}
    \label{tab:open_loop_inference_settings}
\end{table}

Predictions are stored in per-model result files, and partially completed runs are resumed by skipping previously processed evaluation items.
Samples with missing visual inputs, malformed records, or failed generations are logged, excluded from scoring, and re-run when evaluation resumes from a breakpoint.
This protocol ensures that the open-loop evaluation remains deterministic, restartable, and auditable while preserving a uniform interface across all evaluated models.

\subsection{Closed-Loop Evaluation Setting}
The reported $R4$ results are obtained under a unified closed-loop runtime protocol in which each model is deployed as an online VLM service and evaluated in a fresh CARLA-SUMO co-simulation instance for every scenario.
For fair comparison, the scenario loop, simulator configuration, observation interface, and termination rules are fixed across models, and only the model-specific serving configuration is changed.
Several key parameters are summarized in Table~\ref{tab:closed_loop_runtime_settings}.
At runtime, the controller queries the VLM service through an HTTP endpoint, receives action decisions at a fixed control frequency, and applies them directly to the ego vehicle while SUMO governs the surrounding traffic flow.
Each scenario is evaluated independently to ensure that simulator state, traffic evolution, and control history do not carry over across runs.

\begin{table}[htp]
    \centering
    \caption{\textbf{Closed-loop runtime settings used for the \textit{R4} benchmark.}}
    \setlength{\tabcolsep}{5pt}
    \renewcommand{\arraystretch}{1.1}
    \begin{tabular}{l l l}
        \toprule
        \textbf{Setting} & \textbf{Meaning} & \textbf{Value} \\
        \midrule
        Hardware & Evaluation node configuration for CARLA & 1$\times$A800 \\
        Control rate & Frequency of VLM action execution & $4$ Hz \\
        Simulation step & Fixed simulator step length & $0.05$ s \\
        Conversation window & Number of retained dialogue turns & $1$ \\
        image size & Front-view image resolution sent to the model & $800 \times 450$ \\
        Off-road grace time & Allowed off-road duration before termination & $1.5$ s \\
        \bottomrule
    \end{tabular}
    \label{tab:closed_loop_runtime_settings}
\end{table}

To ensure that no mutual interference occurs during scene execution, the CARLA server is restarted for every scenario, the simulator and VLM service will also be checked whether they are healthy before rollout, and results are written to a scenario-specific output directory.
The controller runs with an optional chase-view ego camera, a BEV spectator view, and explicit route visualization, while the scripted traffic actors are managed by the synchronized CARLA-SUMO backend.
A run terminates when the ego vehicle reaches the destination or encounters a benchmark-defined stopping condition, including collision, persistent off-road behavior, blocked driving, or scenario timeout.
Completed scenarios are skipped automatically during resumed evaluation, whereas scenarios that produce control traces without valid VLM inference cause the batch to terminate immediately, which keeps the $R4$ evaluation restartable, while preventing partially failed runs or stale simulator state from contaminating the reported results.

\subsection{Model-to-Control Interface for R4}
\label{appendix:r4_interface}
The $R4$ interface converts a general VLM into an executable closed-loop driving policy without introducing model-specific controllers. 
At each decision step, the runner captures the ego front-view image and a bird's-eye-view route image from CARLA. 
It also extracts structured state from the simulator, including ego speed, the remaining route, the distance to destination, a coarse navigation command, and short-horizon future route points expressed in ego coordinates. 
The default online chain uses the action-decision question, while an optional waypoint-prediction question can be enabled for analyses that require waypoint-style intermediate outputs.

The textual query is instantiated by prepending a live driving condition to the fixed prompt below, where variables are filled from the current CARLA state and route sampler.

\begin{quote}\small
\textbf{Dynamic condition prefix.}
The ego vehicle is driving at the speed of $v_t$ m/s, and it wants to [navigation command].
Remaining route distance to destination is approximately $d_t$ meters.
The local future route in ego coordinates is given as (\texttt{forward\_m}, \texttt{right\_m}): [future route points].

\textbf{Action-decision prompt.}
Your primary objective is to reach the destination in the minimum time while remaining safe and legal.
Hard constraints: do not collide with vehicles, pedestrians, or static obstacles; do not leave the drivable route; obey traffic lights, stop signs, and right-of-way rules.
Driving policy: do not crawl or wait without a concrete safety or rule-based reason; if the route ahead is clear, actively increase speed toward the a safe speed; use steering proactively to follow the future path and begin steering before turns instead of defaulting to zero steering; slow down when required by route geometry, traffic rules, obstacles, or collision risk.
For the steer value, a negative value indicates left, while a positive value indicates right.
When making the decision, consider the ego speed, the future route geometry, drivable space, nearby vehicles, pedestrians, traffic lights, traffic signs, and any immediate collision or off-road risk.
Output only the final action decision without explanation in the exact format:
\textit{steer: <float in [-1.0, 1.0]>};
\textit{target\_speed\_mps: <float in [0.0, $v_{max}$]>}.
\end{quote}

The model response is parsed as a compact action decision consisting of steering intent $\hat{s}_t$ and target speed $\hat{v}_t$.
The bridge clips $\hat{s}_t$ to $[-1,1]$, clips $\hat{v}_t$ to $[0,v_{max}]$\,m/s, rate-limits steering by $0.12$ per control update, and converts the target-speed error $e_t=\hat{v}_t-v_t$ into a CARLA command $u_t=(\delta_t,\tau_t,b_t)$:
\begin{equation}
\delta_t=\mathrm{clip}(\hat{s}_t,\delta_{t-1}-0.12,\delta_{t-1}+0.12)
\end{equation}
\begin{equation}
\tau_t=\mathbf{1}[e_t\ge0]\min(0.85,0.25+0.22e_t)
\end{equation}
\begin{equation}
b_t=\mathbf{1}[e_t<0]\min(0.85,0.30|e_t|)
\end{equation}
If the response cannot be parsed, the previous valid control is retained and the failure is recorded. The resulting command is applied to the ego vehicle at the fixed $R4$ control rate, with the last command held between model queries.
This interface isolates the capability under evaluation from simulator-specific implementation details. All models receive the same visual inputs, route context, query protocol, parsing rules, and actuation bridge; only the model backend differs. 

\subsection{Benchmark-Guided SFT Details}
All benchmark-guided SFT experiments are conducted on the base model Qwen/Qwen3-VL-8B-Instruct\cite{qwen3technicalreport} using LoRA-based SFT, launched through the swift-sft\cite{zhao2024swiftascalablelightweightinfrastructure} pipeline. A validation split of $5\%$ is held out from each training dataset. Across all runs, the vision tower and aligner are frozen, LoRA adapters are attached to all linear layers, and the same optimization recipe is used to ensure the differences in performance can be attributed to the selected capability supervision rather than to changing training conditions. The analytical focus of this study is transfer beyond the supervised scope: in the open-loop analysis, we pay primary attention to changes on non-target question types, and in the closed-loop analysis, we test whether open-loop supervision transfers to $R4$. No $R4$ scenario is used for training or fine-tuning in any reported experiment. The main parameters used during training are shown in Table \ref{tab:sft_training_config}.

\begin{table}[htp]
    \centering
    \caption{\textbf{Training configuration for benchmark-guided SFT.}}
    \setlength{\tabcolsep}{5pt}
    \renewcommand{\arraystretch}{1.1}
    \begin{tabular}{l l l}
        \toprule
        \textbf{Symbol} & \textbf{Meaning} & \textbf{Value} \\
        \midrule
        $r$ & LoRA rank & $8$ \\
        $\alpha$ & LoRA scaling factor & $32$ \\
        $d_{\mathrm{train}}$ & Training dtype & \texttt{bfloat16} \\
        $B_{\mathrm{train}}$ & Per-device training batch size & $1$ \\
        $B_{\mathrm{eval}}$ & Per-device evaluation batch size & $1$ \\
        $G$ & Gradient accumulation steps & $8$ \\
        $\eta$ & Learning rate & $1\times10^{-4}$ \\
        $\rho_{\mathrm{warm}}$ & Warmup ratio & $0.05$ \\
        $L_{\max}$ & Maximum training sequence length & $16384$ \\
        $N_{\mathrm{img}}$ & Maximum visual token budget per sample & $1024$ \\
        $S_{\mathrm{save}}$ & Checkpoint interval & $50$ \\
        $S_{\mathrm{eval}}$ & Validation interval & $50$ \\
        $\mathcal{H}$ & Training hardware & 4$\times$A800\\
        \bottomrule
    \end{tabular}
    \label{tab:sft_training_config}
\end{table}

All A800 GPUs used in this work are 80GB variants. Compute requirements vary across models due to differences in model size, tensor-parallel configuration, and serving throughput. For the reported benchmark runs, open-loop evaluation requires approximately 24 hours  per model on average, with a total cost of roughly 48 to 144 A800 GPU-hours. Closed-loop evaluation over the 100 scenarios requires approximately 20 hours of wall-clock time per model on average, with a total cost of roughly 36 to 72 A800 GPU-hours.

\section{Open-Loop Evaluation Metrics Designs}
\label{Evaluation_Metrics_Designs}
For binary judgment and multiple-choice tasks, we use exact-match scoring,
\begin{equation}
s_i = 100 \cdot \mathbf{1}[\hat{y}_i = y_i].
\end{equation}
This protocol is used for existence recognition, state identification, spatial relation reasoning, and outcome prediction. In implementation, yes/no answers are normalized to canonical binary labels, while multiple-choice responses are first matched by option letter and then, when needed, by normalized answer text. Unparseable outputs receive score $0$ and are logged as parse failures.

For counting tasks, we first extract the valid numeric token from the response, round it to the nearest non-negative integer, and define the normalized error
\begin{equation}
E_i^{\mathrm{cnt}} = \frac{|\hat{y}_i - y_i|}{\max(y_i, c)},
\end{equation}
where $c>0$ prevents the denominator from becoming unstable for small counts. We fix $c=1$, $\tau_{\mathrm{cnt}}=0.05$, and choose $\sigma_{\mathrm{cnt}}$ from the half-credit target $E_{50}^{\mathrm{cnt}}=0.20$, yielding
\begin{equation}
\sigma_{\mathrm{cnt}} = \frac{0.20-0.05}{\ln 2}.
\end{equation}
For distance estimation, we extract the first valid numeric token, clip it to a non-negative scalar, and adopt a scale-invariant log-ratio error
\begin{equation}
E_i^{\mathrm{dist}} = \left|\log \frac{\hat{y}_i + \epsilon}{y_i + \epsilon}\right|,
\end{equation}
where $\epsilon>0$ avoids numerical singularity near zero distance. We fix $\epsilon=0.1$, $\tau_{\mathrm{dist}}=0$, and set the half-credit target to $E_{50}^{\mathrm{dist}}=\ln 2$, which gives $\sigma_{\mathrm{dist}}=1.0$. Both errors are converted into scores through the same bounded decay function
\begin{equation}
G(E;\tau,\sigma)=
\begin{cases}
    100, & E \le \tau, \\
    100 \exp\left(-\frac{E-\tau}{\sigma}\right), & E > \tau,
\end{cases}
\end{equation}
yielding $s_i = G(E_i^{\mathrm{cnt}};\tau_{\mathrm{cnt}},\sigma_{\mathrm{cnt}})$ for counting and $s_i = G(E_i^{\mathrm{dist}};\tau_{\mathrm{dist}},\sigma_{\mathrm{dist}})$ for distance estimation. This design assigns full credit to exact or near-exact predictions while penalizing progressively larger numeric deviations in a smooth and bounded manner.

For localization, the model predicts a bounding box $\hat{B}_i$ and is scored by intersection-over-union with the ground-truth box $B_i$:
\begin{equation}
s_i = 100 \cdot \mathrm{IoU}(B_i,\hat{B}_i).
\end{equation}



For situation description, we follow the official protocol\cite{marcu2024lingoqa}. Given the prompt, we run the released text-classification model and take the classifier confidence returned in the score field as the semantic-correctness probability $p_i \in [0,1]$. The sample score is then
\begin{equation}
s_i = 100 \, p_i.
\end{equation}
The evaluation script also reports a diagnostic pass flag at threshold $0.5$, but this flag is not used in the benchmark score. 

For sequence planning, let $\pi_i$ and $\hat{\pi}_i$ denote the ground-truth and predicted temporal orders. If $I_i$ is the number of pairwise inversions between them and $\binom{n_i}{2}$ is the total number of ordered pairs, the score is defined as
\begin{equation}
s_i = 100 \left(1 - \frac{I_i}{\binom{n_i}{2}}\right),
\end{equation}
which rewards partial order consistency rather than requiring exact full-sequence recovery.

We aggregate scores hierarchically. For a task type $t$ with $N_t$ evaluation items, the task score is
\begin{equation}
S_t = \frac{1}{N_t}\sum_{i=1}^{N_t} s_i^{(t)}.
\end{equation}

For a rank $r$ containing task set $\mathcal{T}_r$, the rank score is the arithmetic mean over task scores:
\begin{equation}
S_r = \frac{1}{|\mathcal{T}_r|}\sum_{t \in \mathcal{T}_r} S_t.
\end{equation}

For the headline open-loop score used in Table~\ref{tab:table1}, we report a fixed weighted sum over the 14 task scores,
\begin{equation}
S_{\mathrm{open}} = \sum_{t=1}^{14} w_t S_t, \qquad \sum_{t=1}^{14} w_t = 1,
\end{equation}
where the released task weights $w_t$ are the frozen calibration weights shown in the table header and defined in Appendix~\ref{appendix:weight_allocation}. This aggregation preserves interpretability at the task and rank levels while avoiding repeated over-rewarding of highly redundant subtasks.

\section{Closed-Loop Scoring Details}
\label{appendix:r4_scoring}
Inspired by Bench2Drive\cite{jia2024bench2drive}, the closed-loop score for $R4$ is implemented as follows. Let $S_{\mathrm{route}} \in [0,100]$, $S_{\mathrm{safety}} \in [0,1]$, and $S_{\mathrm{eff}} \in [0,1]$ denote the route-completion, safety, and efficiency terms, respectively. The final scene score is defined as
\begin{equation}
S_{\mathrm{R4}} = S_{\mathrm{route}} \cdot S_{\mathrm{safety}} \cdot S_{\mathrm{eff}}.
\end{equation}
This multiplicative design prevents successful route completion from masking unsafe or inefficient behavior.

The route term measures normalized progress toward the designated destination. In implementation, the ego trajectory is projected onto a reference route polyline, and route progress is updated only when the projected lateral deviation does not exceed $6$\,m. If the ego vehicle reaches the goal region, the route score is set to $100$; otherwise it is proportional to the completed fraction of the reference route:
\begin{equation}
S_{\mathrm{route}} =
\begin{cases}
    100, & \text{destination reached}, \\
100 \cdot \dfrac{d_{\mathrm{comp}}}{d_{\mathrm{route}}}, & \text{otherwise},
\end{cases}
\end{equation}
where $d_{\mathrm{comp}}$ is the maximum completed route distance and $d_{\mathrm{route}}$ is the reference route length. Destination arrival is triggered when the ego vehicle enters the scenario-defined goal radius, which is set set to 8\,m by default.

The safety term combines event penalties with lane-keeping quality:
\begin{equation}
S_{\mathrm{safety}} = \left(\prod_{e \in \mathcal{E}} p_e\right)\cdot (1-r_{\mathrm{offroad}}),
\end{equation}
where $\mathcal{E}$ is the set of recorded safety-critical events, $p_e \in [0,1]$ is the penalty multiplier for event $e$, and $r_{\mathrm{offroad}}$ is the proportion of time spent outside the valid route region. In the current implementation, pedestrian collisions incur zero score for the corresponding event term, vehicle collisions and static-object collisions incur multiplicative penalties of $0.2$ and $0.3$, respectively, and scenario timeout and blocked timeout incur penalties of $0.7$. Off-road timeout, manual interruption, and system error are treated as terminal failure events in the result record, with their corresponding penalty multipliers defined directly in the implementation.

The efficiency term evaluates whether the realized trajectory is unnecessarily long or slow relative to the route reference. Distance efficiency and time efficiency are computed as
\begin{equation}
\eta_{\mathrm{dist}} = \min\left(1,\frac{\alpha \, d_{\mathrm{route}}}{d_{\mathrm{travel}}}\right), \qquad
\eta_{\mathrm{time}} = \min\left(1,\frac{\beta \, t_{\mathrm{ref}}}{t_{\mathrm{sim}}}\right),
\end{equation}
where $d_{\mathrm{travel}}$ is the actual traveled distance, $t_{\mathrm{sim}}$ is the scenario duration, $t_{\mathrm{ref}}$ is the reference duration, and $\alpha$, $\beta$ are grace factors. The implementation sets $\alpha = 1.10$ and $\beta = 1.50$. The final efficiency term is the weighted sum
\begin{equation}
S_{\mathrm{eff}} = w_d \eta_{\mathrm{dist}} + w_t \eta_{\mathrm{time}},
\end{equation}
with $w_d = 0.7$ and $w_t = 0.3$.

In addition to the composed score, the evaluator records route, safety, and efficiency terms separately, together with the event list, traveled distance, off-road ratio, lane-keeping penalty, reference duration, and scenario termination status. This design preserves interpretability at both the overall and component levels.

\section{Weight Allocation Details}
\label{appendix:weight_allocation}
The objective of weight allocation is to make the composite open-loop score reflect distinct capability dimensions rather than repeated measurements of highly correlated subtasks. To keep the released metric stable, we estimate the correlation structure once during an initial calibration phase after the first benchmark-testing round, derive the weights from that calibration pool, and then freeze them for all subsequent evaluations. We therefore fix the rank-level budgets to
\begin{equation}
B_{R1}=0.3,\qquad B_{R2}=0.3,\qquad B_{R3}=0.4,
\end{equation}
and distribute each budget according to correlation-adjusted subtask uniqueness.

Let $\rho_{ij}$ denote the Spearman correlation between subtasks $i$ and $j$. For a subtask $i$ in rank $r(i)$, we define its redundancy load as
\begin{equation}
L_i = 1 + \lambda_{\mathrm{same}} \sum_{\substack{j \neq i \\ r(j)=r(i)}} \bigl[\max(\rho_{ij},0)\bigr]^\gamma
+ \lambda_{\mathrm{cross}} \sum_{\substack{j \neq i \\ r(j)\neq r(i)}} \bigl[\max(\rho_{ij},0)\bigr]^\gamma,
\end{equation}
where $\lambda_{\mathrm{same}}$ and $\lambda_{\mathrm{cross}}$ control the penalty strength for within-rank and cross-rank redundancy, respectively, and $\gamma$ controls how strongly the penalty concentrates on highly correlated pairs. In our implementation, $\lambda_{\mathrm{same}}=1.0$, $\lambda_{\mathrm{cross}}=0.35$, and $\gamma=2.0$. Negative correlations are clipped to zero so that anticorrelation is not treated as additional uniqueness.

The raw uniqueness of subtask $i$ is then defined as the inverse redundancy load,
\begin{equation}
u_i=\frac{1}{L_i},
\end{equation}
and normalized within each rank:
\begin{equation}
\pi_i=\frac{u_i}{\sum_{k \in \mathcal{T}_{r(i)}} u_k},
\end{equation}
where $\mathcal{T}_{r(i)}$ is the set of subtasks in the same rank as $i$. The final subtask weight is
\begin{equation}
w_i = B_{r(i)} \cdot \pi_i.
\end{equation}
As a result, two nearly identical subtasks with similar correlation profiles tend to split a shared effective weight, whereas a less redundant subtask within the same rank will receive a larger allocation, which can guarantee that subtask weights sum to the preassigned rank budget while naturally down-weighting subtasks whose information is already extensively covered by correlated neighbors, as reflected in the weights reported in Table~\ref{tab:table1}.

\newcommand{\rowexample}[5]{
\vspace{0.5em}
\noindent
\setlength{\fboxsep}{6pt}
\setlength{\fboxrule}{0.5pt}
\noindent\rule{0.97\linewidth}{0.4pt}
\begin{minipage}[t]{0.97\linewidth}
\vspace{0pt}
\begin{minipage}[t]{0.60\linewidth}
    \vspace{0pt}
    \centering
    #1
\end{minipage}
\hfill
\begin{minipage}[t]{0.37\linewidth}
    \vspace{0pt}
    {\small \textbf{#5}\par}
    \vspace{0.4em}
    {\footnotesize \textbf{Sys. Prompt:} #2\par}
    \vspace{0.4em}
    {\footnotesize \textbf{Prompt:} #3\par}
    \vspace{0.4em}
    {\footnotesize \textbf{Answer:} #4\par}
\end{minipage}
\end{minipage}
\vspace{0.5em}
}

\newcommand{\rowexamplefive}[9]{
\vspace{0.5em}
\noindent\rule{0.97\linewidth}{0.4pt}

\begin{minipage}[t]{0.97\linewidth}
\vspace{0pt}

\begin{minipage}[t]{0.62\linewidth}
    \vspace{0pt}
    \centering

    \begin{minipage}[t]{0.47\linewidth}
        \centering
        \includegraphics[width=\linewidth]{#1}
        {\scriptsize $t_1$}
    \end{minipage}
    \hfill
    \begin{minipage}[t]{0.47\linewidth}
        \centering
        \includegraphics[width=\linewidth]{#2}
        {\scriptsize $t_2$}
    \end{minipage}

    \vspace{0.4em}

    \begin{minipage}[t]{0.47\linewidth}
        \centering
        \includegraphics[width=\linewidth]{#3}
        {\scriptsize $t_3$}
    \end{minipage}
    \hfill
    \begin{minipage}[t]{0.47\linewidth}
        \centering
        \includegraphics[width=\linewidth]{#4}
        {\scriptsize $t_4$}
    \end{minipage}

    \vspace{0.4em}

    \makebox[\linewidth][c]{%
        \begin{minipage}[t]{0.47\linewidth}
            \centering
            \includegraphics[width=\linewidth]{#5}
            {\scriptsize $t_5$}
        \end{minipage}
    }
\end{minipage}
\hfill
\begin{minipage}[t]{0.33\linewidth}
    \vspace{0pt}
    {\small \textbf{#9}\par}
    \vspace{0.4em}
    {\footnotesize \textbf{Sys. Prompt:} #6\par}
    \vspace{0.4em}
    {\footnotesize \textbf{Prompt:} #7\par}
    \vspace{0.4em}
    {\footnotesize \textbf{Answer:} #8\par}
\end{minipage}

\end{minipage}
\vspace{0.5em}
}

\newcommand{\rowexamplesix}[7]{
\vspace{0.5em}
\noindent\rule{0.97\linewidth}{0.4pt}

\begin{minipage}[t]{0.97\linewidth}
\vspace{0pt}

\begin{minipage}[t]{0.66\linewidth}
    \vspace{0pt}
    \centering

    \begin{minipage}[t]{0.47\linewidth}
        \centering
        \includegraphics[width=\linewidth]{#5}

        {\scriptsize FRONT\_RIGHT}
    \end{minipage}
    \hfill
    \begin{minipage}[t]{0.47\linewidth}
        \centering
        \includegraphics[width=\linewidth]{#6}

        {\scriptsize BACK\_RIGHT}
    \end{minipage}

    \vspace{0.4em}

    \begin{minipage}[t]{0.47\linewidth}
        \centering
        \includegraphics[width=\linewidth]{#3}

        {\scriptsize FRONT}
    \end{minipage}
    \hfill
    \begin{minipage}[t]{0.47\linewidth}
        \centering
        \includegraphics[width=\linewidth]{#4}

        {\scriptsize BACK}
    \end{minipage}

    \vspace{0.4em}
    
    \begin{minipage}[t]{0.47\linewidth}
        \centering
        \includegraphics[width=\linewidth]{#1}

        {\scriptsize FRONT\_LEFT}
    \end{minipage}
    \hfill
    \begin{minipage}[t]{0.47\linewidth}
        \centering
        \includegraphics[width=\linewidth]{#2}

        {\scriptsize BACK\_LEFT}
    \end{minipage}

\end{minipage}
\hfill
\begin{minipage}[t]{0.29\linewidth}
    \vspace{0pt}
    #7
\end{minipage}

\end{minipage}
\vspace{0.5em}
}

\newcommand{\rowexampleblock}[2]{
\vspace{0.5em}
\noindent\rule{0.97\linewidth}{0.4pt}

\begin{minipage}[t]{0.97\linewidth}
\vspace{0pt}

\begin{minipage}[t]{0.66\linewidth}
    \vspace{0pt}
    \centering
    #1
\end{minipage}
\hfill
\begin{minipage}[t]{0.29\linewidth}
    \vspace{0pt}
    #2
\end{minipage}

\end{minipage}
\vspace{0.5em}
}

\newcommand{\imgpair}[4]{
\begin{minipage}[t]{0.47\linewidth}
    \centering
    \includegraphics[width=\linewidth]{#1}

    {\scriptsize #2}
\end{minipage}
\hfill
\begin{minipage}[t]{0.47\linewidth}
    \centering
    \includegraphics[width=\linewidth]{#3}

    {\scriptsize #4}
\end{minipage}
}

\newcommand{\qatextblock}[4]{
{\small \textbf{#1}\par}
\vspace{0.4em}
{\footnotesize \textbf{Sys. Prompt:} #2\par}
\vspace{0.4em}
{\footnotesize \textbf{Prompt:} #3\par}
\vspace{0.4em}
{\footnotesize \textbf{Answer:} #4\par}
}

\newcommand{\imgsinglecenter}[2]{
\makebox[\linewidth][c]{%
\begin{minipage}[t]{0.47\linewidth}
    \centering
    \includegraphics[width=\linewidth]{#1}

    {\scriptsize $#2$}
\end{minipage}
}
}

\section{Examples for QAs in DriveHierarchy}

\begin{figure}[!htbp]
    \centering
    
    \rowexample
    {\includegraphics[width=0.95\linewidth]{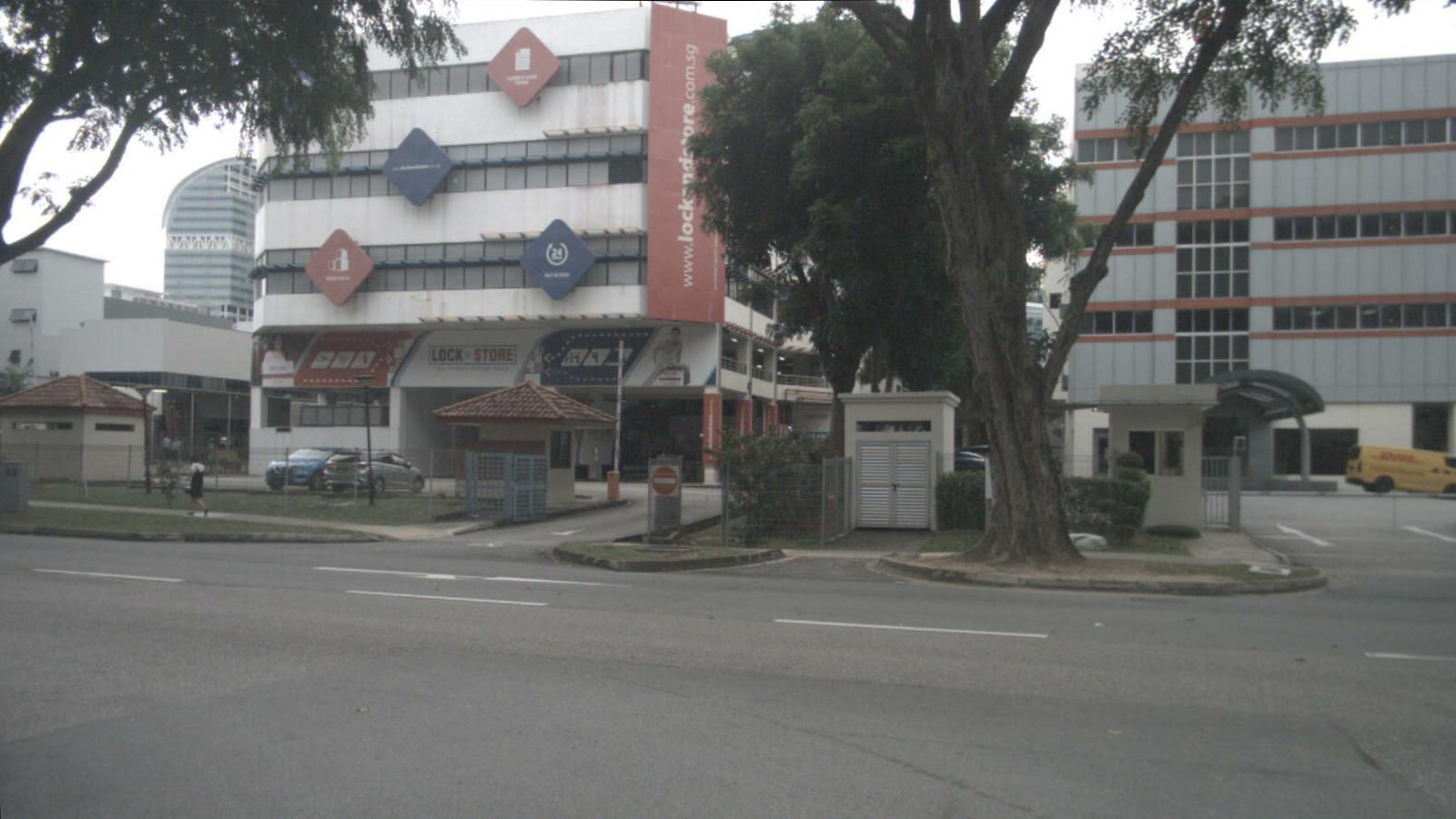}}
    {You are a mature and cautious driver with many years of driving experience. You now need assistance answering questions related to the driving process. \par Answer with only 'yes' or 'no'. }
    {Is there any \textbf{Pedestrian} visible in CAM\_FRONT?}
    {Yes.}
    {R1-1-A Existence}

    \rowexample
    {\includegraphics[width=0.95\linewidth]{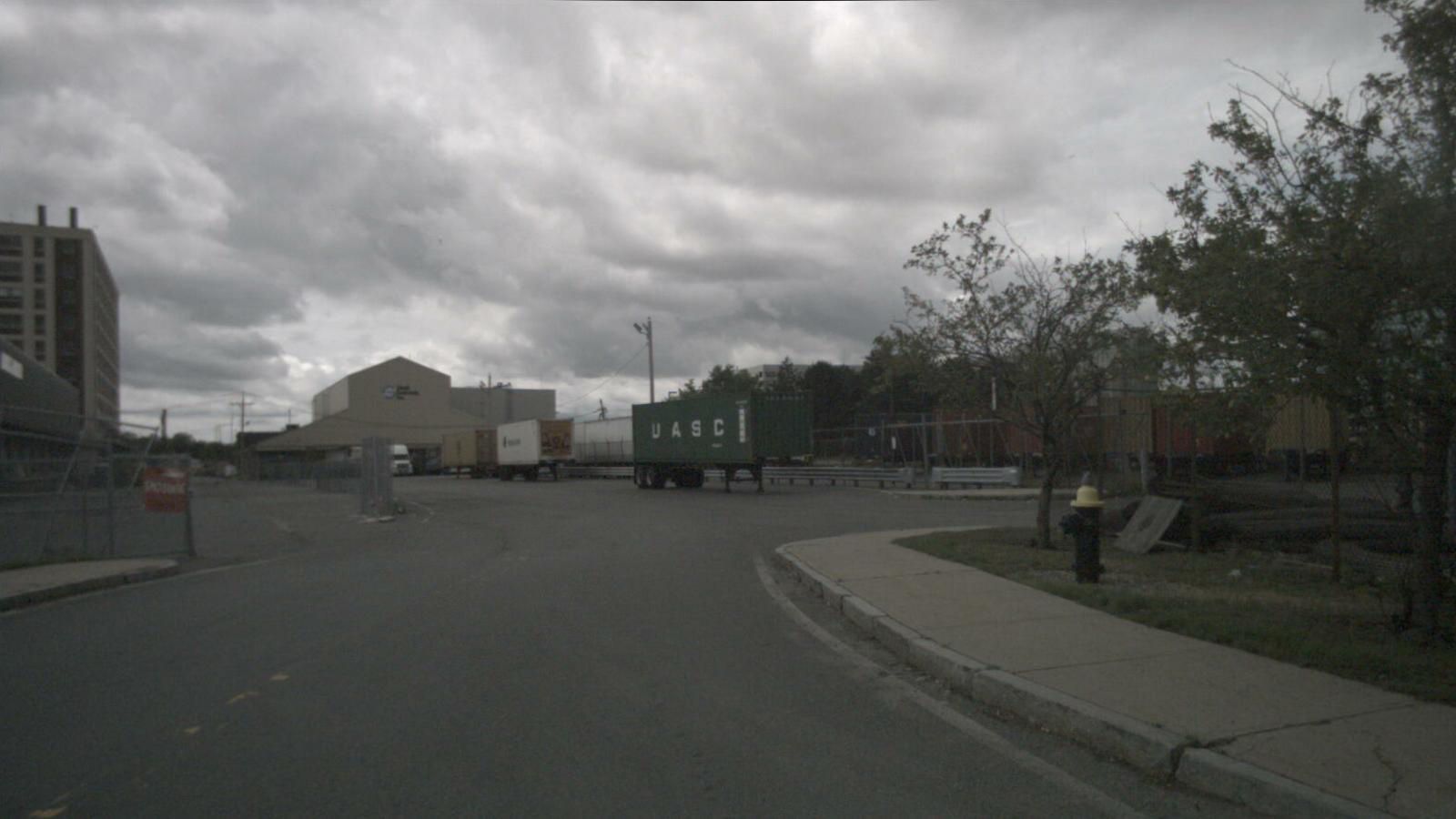}}
    {You are a mature and cautious driver with many years of driving experience. You now need assistance answering questions related to the driving process. \par Answer with only 'yes' or 'no'. }
    {Is there any \textbf{Traffic Cone} visible in CAM\_FRONT?}
    {No.}
    {R1-1-A Existence}

    \rowexample
    {\includegraphics[width=0.95\linewidth]{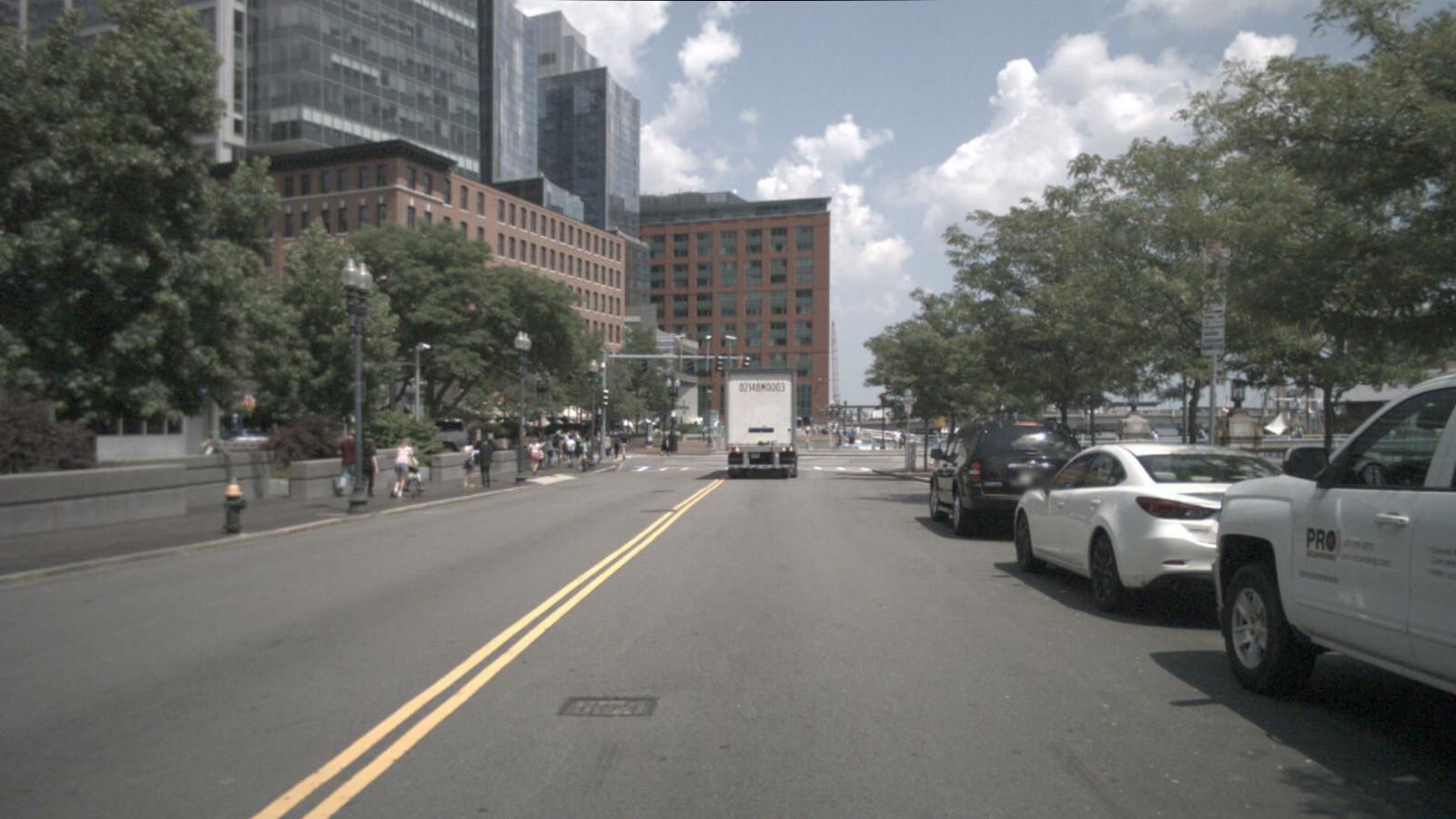}}
    {You are a mature and cautious driver with many years of driving experience. You now need assistance answering questions related to the driving process. \par Answer with only the numeric result (integer). }
    {How many \textbf{Vehicles} are visible in CAM\_FRONT?}
    {4.}
    {R1-1-B Counting}

    \rowexample
    {\includegraphics[width=0.95\linewidth]{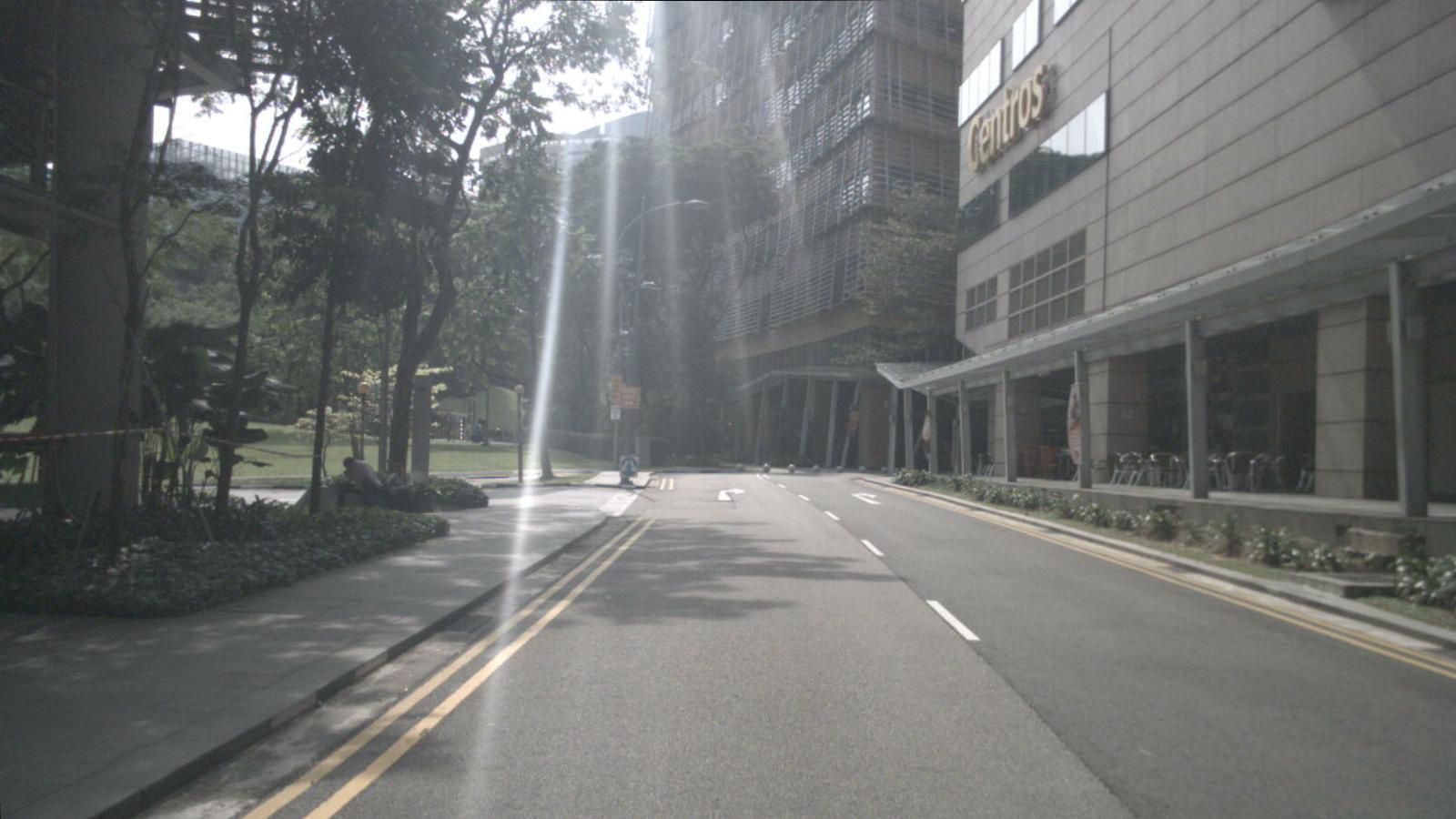}}
    {You are a mature and cautious driver with many years of driving experience. You now need assistance answering questions related to the driving process. \par Choose the correct option and output only the option letter (e.g., A/B/C). }
    {What is the current status of the nearest Pedestrian in CAM\_FRONT?\par
    ["A. moving.", "B. standing.", "C. sitting or lying."]}
    {C.}
    {R1-1-C State Attribute}

    \vspace{0.5em}
    \noindent\rule{0.97\linewidth}{0.4pt}
    \caption{Representative examples from Rank 1-1 Questions.}
    \label{fig:r1_1}
\end{figure}

\begin{figure}[!htbp]
    \centering
    
    \rowexample
    {\includegraphics[width=0.95\linewidth]{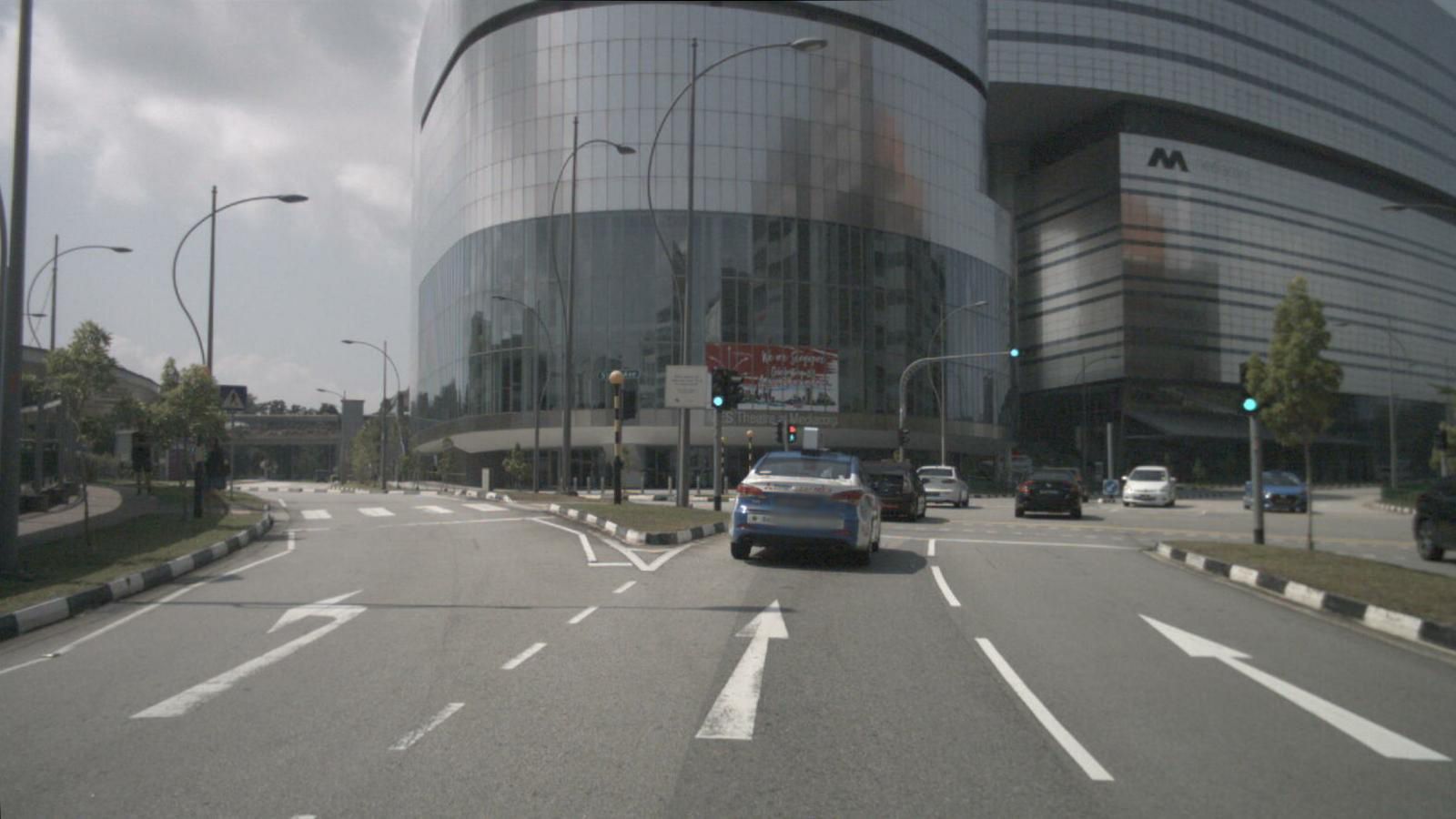}}
    {You are a mature and cautious driver with many years of driving experience. You now need assistance answering questions related to the driving process. \par Answer with only the numeric result (decimal).}
    {What is the distance to the nearest Vehicle in CAM\_FRONT? (Image resolution: 1600x900). Answer in meters.}
    {17.77 m.}
    {R1-2-A Nearest Object}

    \rowexample
    {\includegraphics[width=0.95\linewidth]{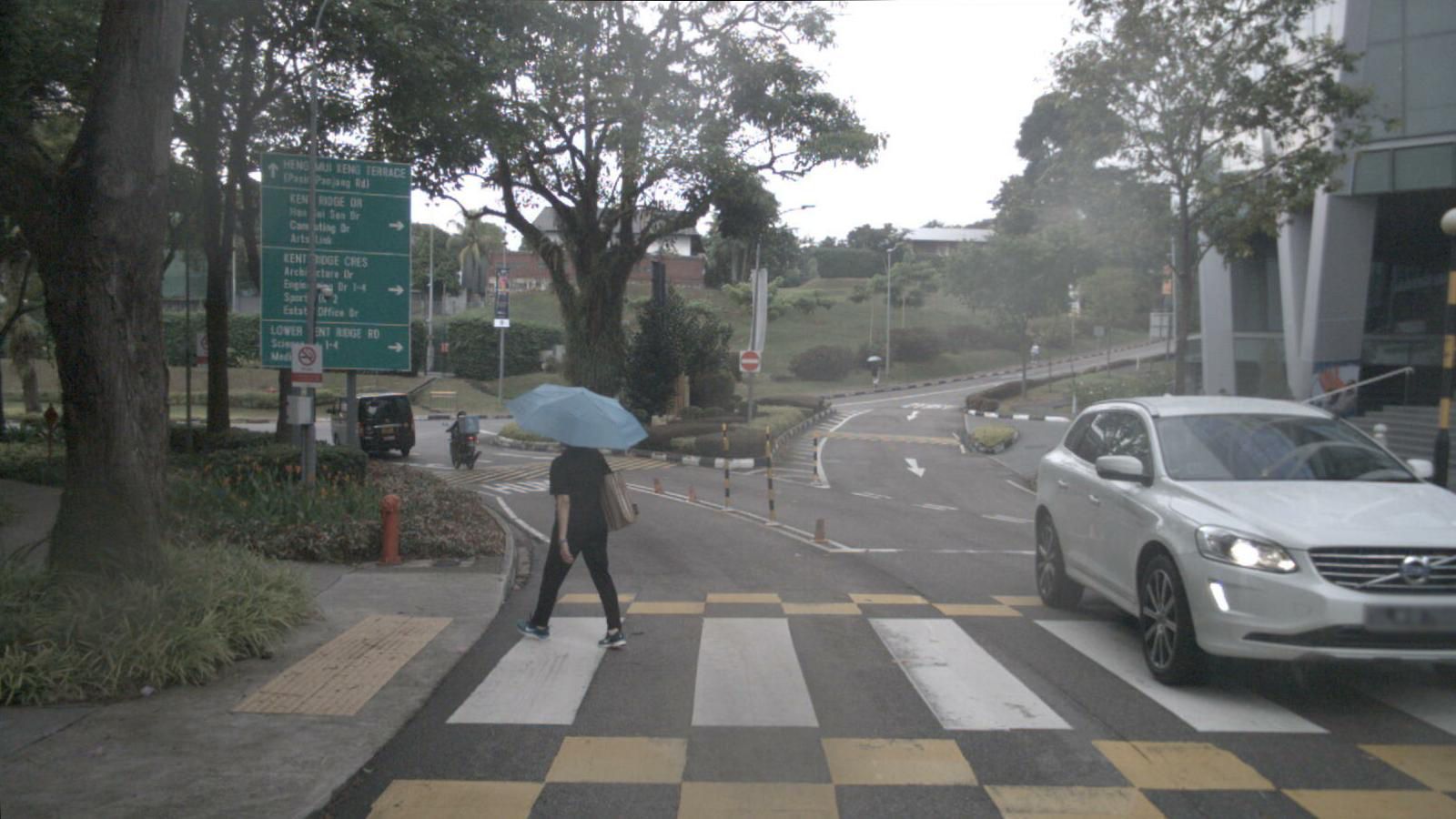}}
    {You are a mature and cautious driver with many years of driving experience. You now need assistance answering questions related to the driving process. \par Answer with only the numeric result (decimal).}
    {What is the distance to the nearest Bicycle in CAM\_FRONT? (Image resolution: 1600x900). Answer in meters.}
    {32.00 m.}
    {R1-2-A Nearest Object}

    \rowexample
    {\includegraphics[width=0.95\linewidth]{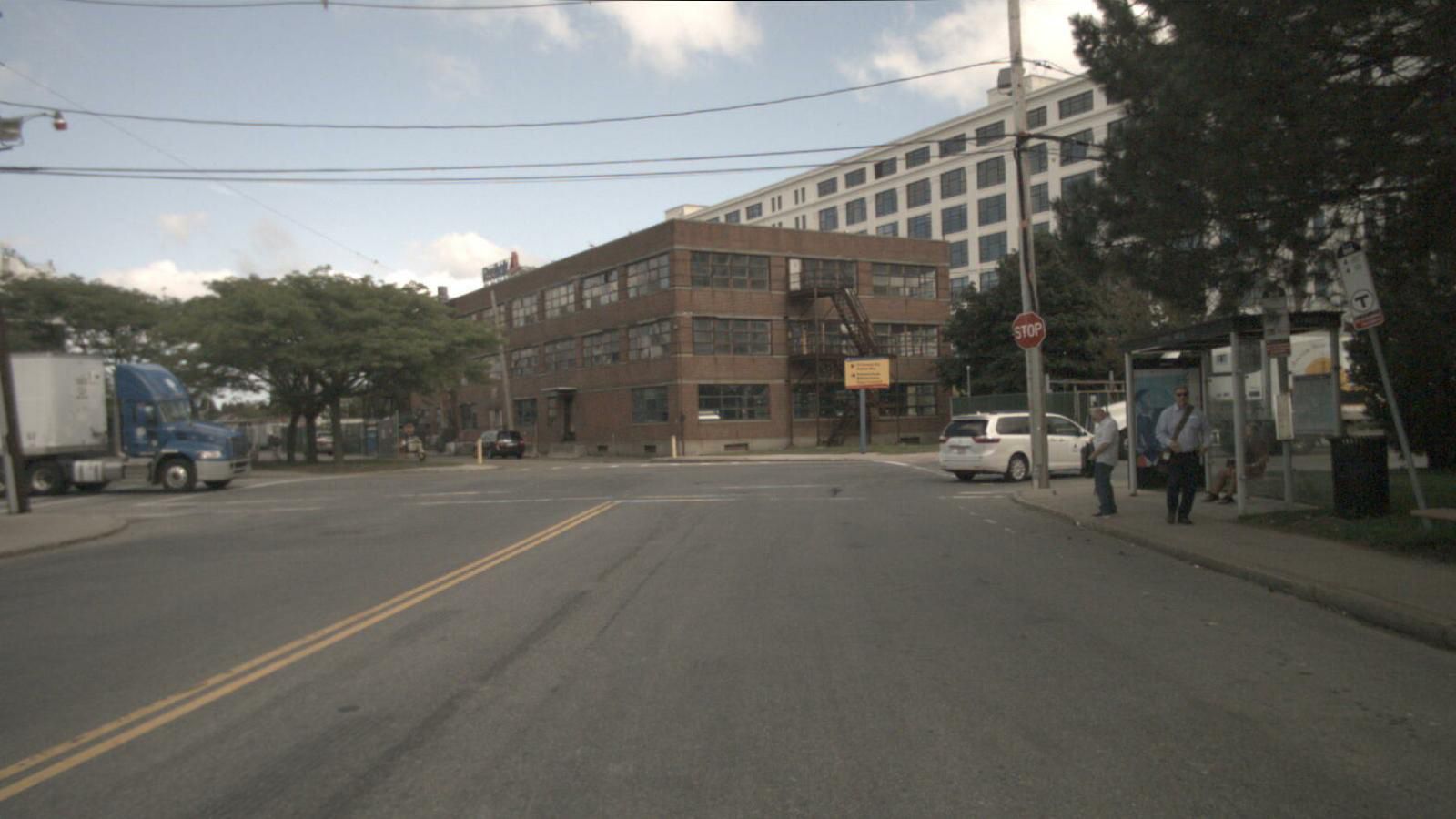}}
    {You are a mature and cautious driver with many years of driving experience. You now need assistance answering questions related to the driving process. \par Answer with only the numeric result (decimal).}
    {Calculate the distance to the Vehicle located at pixel coordinates [105, 475] (Image resolution: 1600x900). Answer in meters.}
    {40.94 m}
    {R1-2-B Certain Object}

    \rowexample
    {\includegraphics[width=0.95\linewidth]{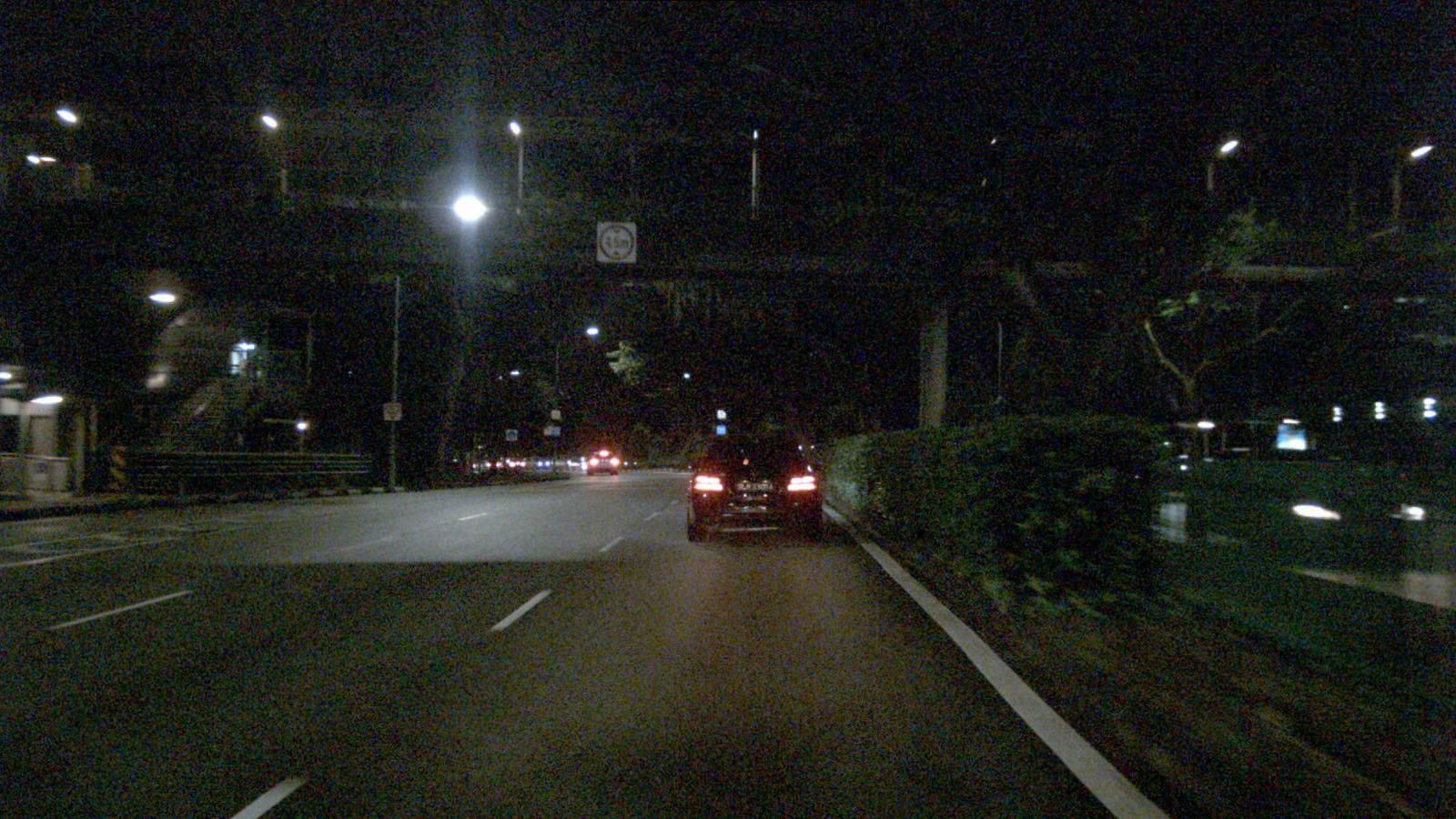}}
    {You are a mature and cautious driver with many years of driving experience. You now need assistance answering questions related to the driving process. \par Answer with only the numeric result (integer).}
    {How many Vehicles are within the Mid range (20-50m) in camera?}
    {2.}
    {R1-2-C Distance Bucket}

    \vspace{0.5em}
    \noindent\rule{0.97\linewidth}{0.4pt}
    \caption{Representative examples from Rank 1-2 Questions.}
    \label{fig:r1_2}
\end{figure}

\begin{figure}[!htbp]
    \centering
    
    \rowexample
    {\includegraphics[width=0.95\linewidth]{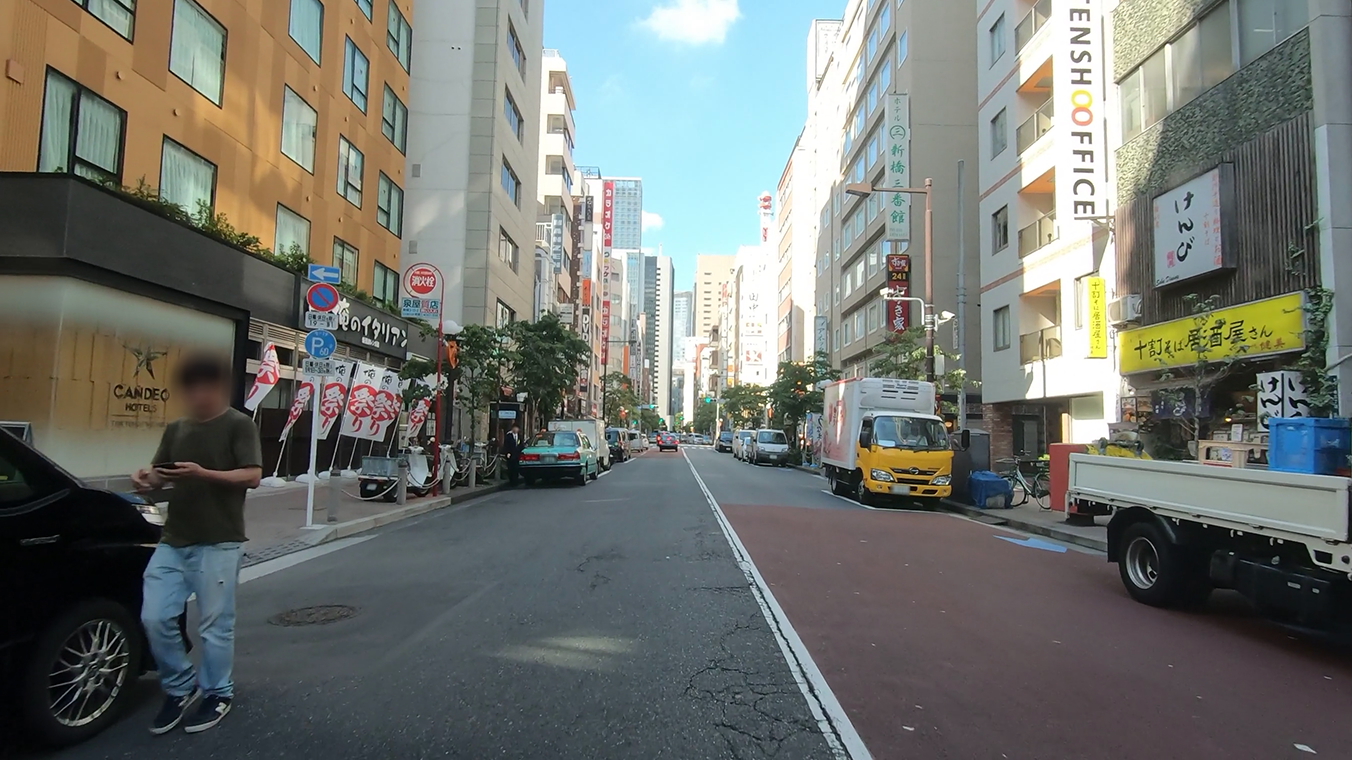}}
    {You are a mature and cautious driver with many years of driving experience. You now need assistance answering questions related to the driving process. \par Output only the bounding box as [x1, y1, x2, y2] in pixels.}
    {Please locate the object described as: "There is a man wearing blue pants and holding a phone, walking on the left side of the road, towards the ego car.". Image resolution: 2704x1520. Find the bounding box and output its diagonal points in the format [xmin, ymin, xmax, ymax].}
    {[264, 709, 525, 1469]}
    {R1-3 Location Questions}

    \rowexample
    {\includegraphics[width=0.95\linewidth]{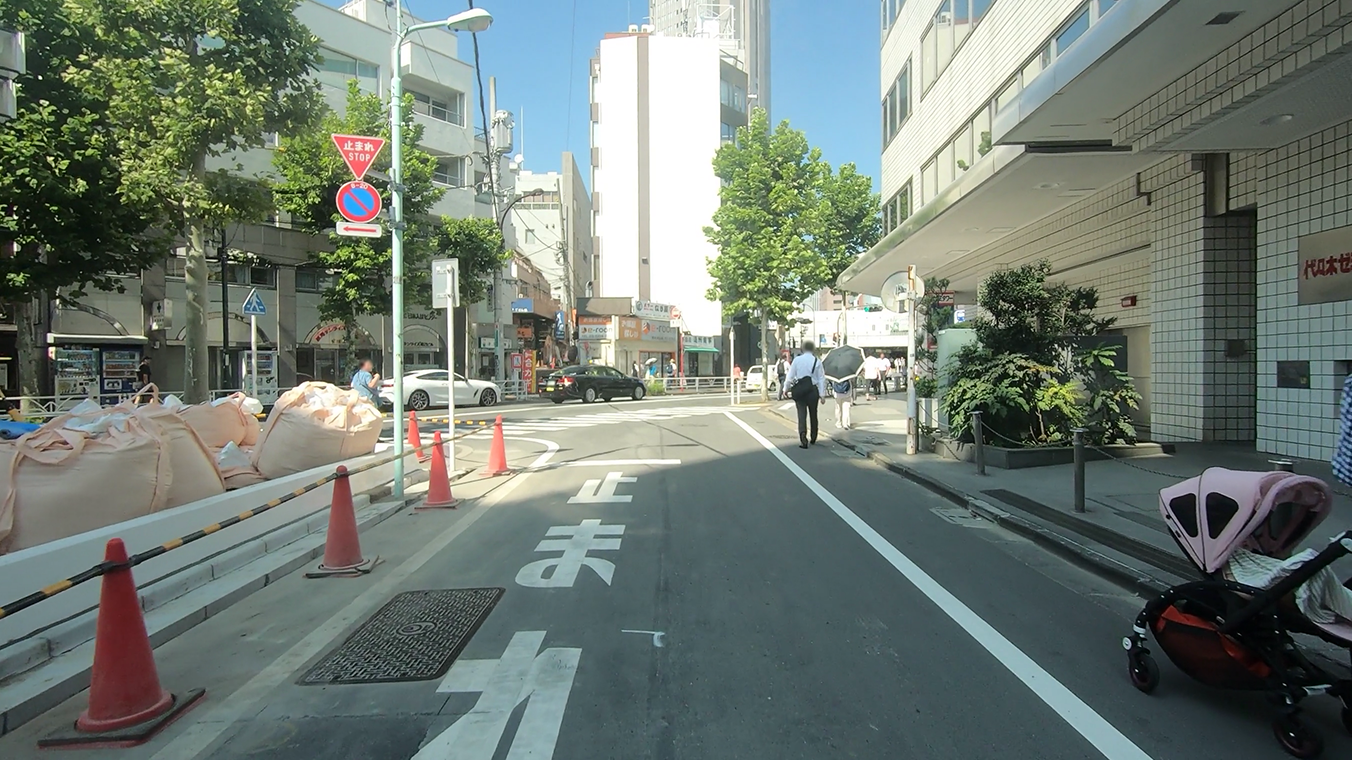}}
    {You are a mature and cautious driver with many years of driving experience. You now need assistance answering questions related to the driving process. \par Output only the bounding box as [x1, y1, x2, y2] in pixels.}
    {Please locate the object described as: "There is a traffic stop sign, in front of the ego car.". Image resolution: 2704x1520. Find the bounding box and output its diagonal points in the format [xmin, ymin, xmax, ymax].}
    {[660, 263, 771, 360]}
    {R1-3 Location Questions}

    \vspace{0.5em}
    \noindent\rule{0.97\linewidth}{0.4pt}
    \caption{Representative examples from Rank 1-3 Questions.}
    \label{fig:r1_3}
\end{figure}

\begin{figure}[!htbp]
    \centering
    
    \rowexamplefive
    {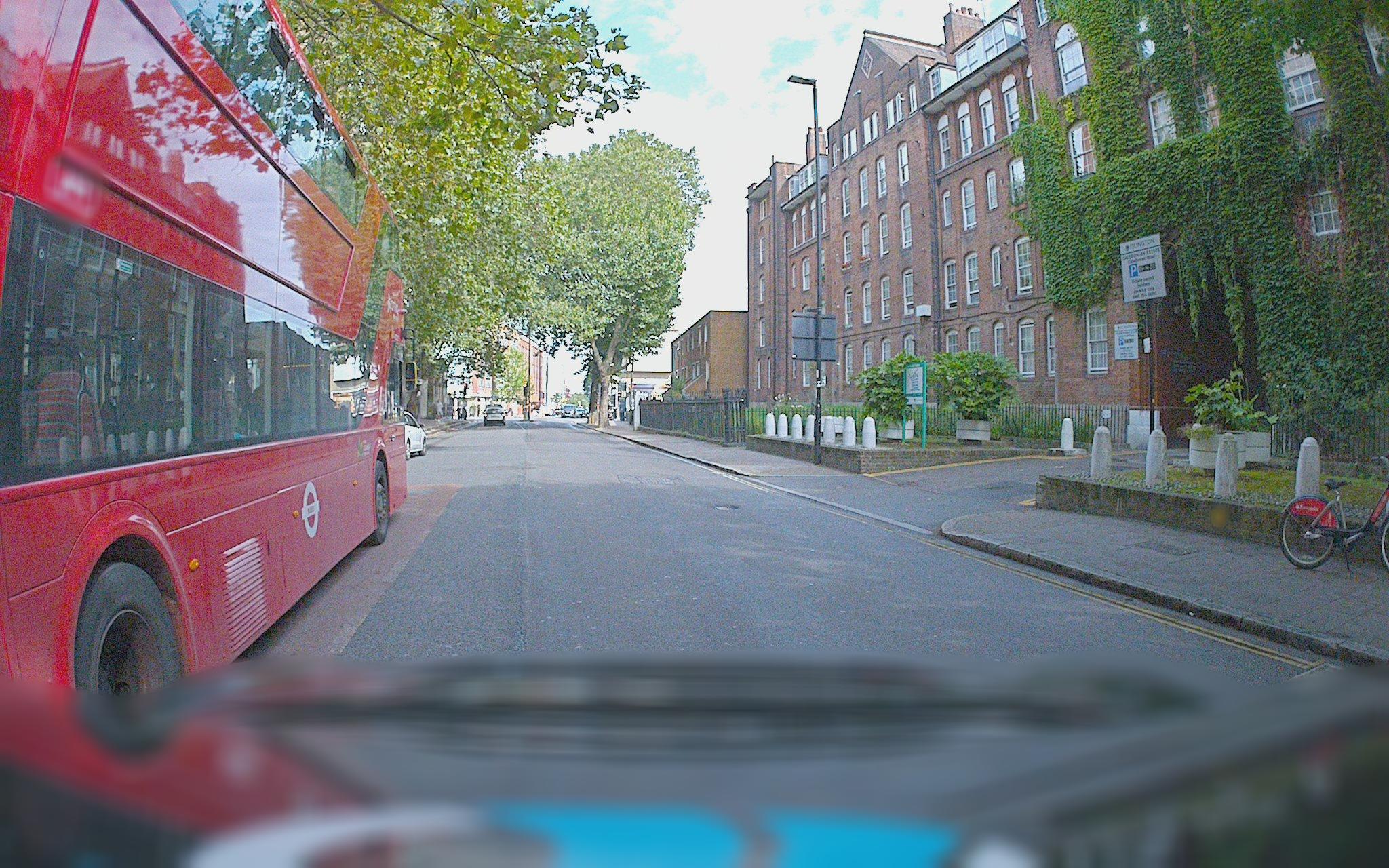}
    {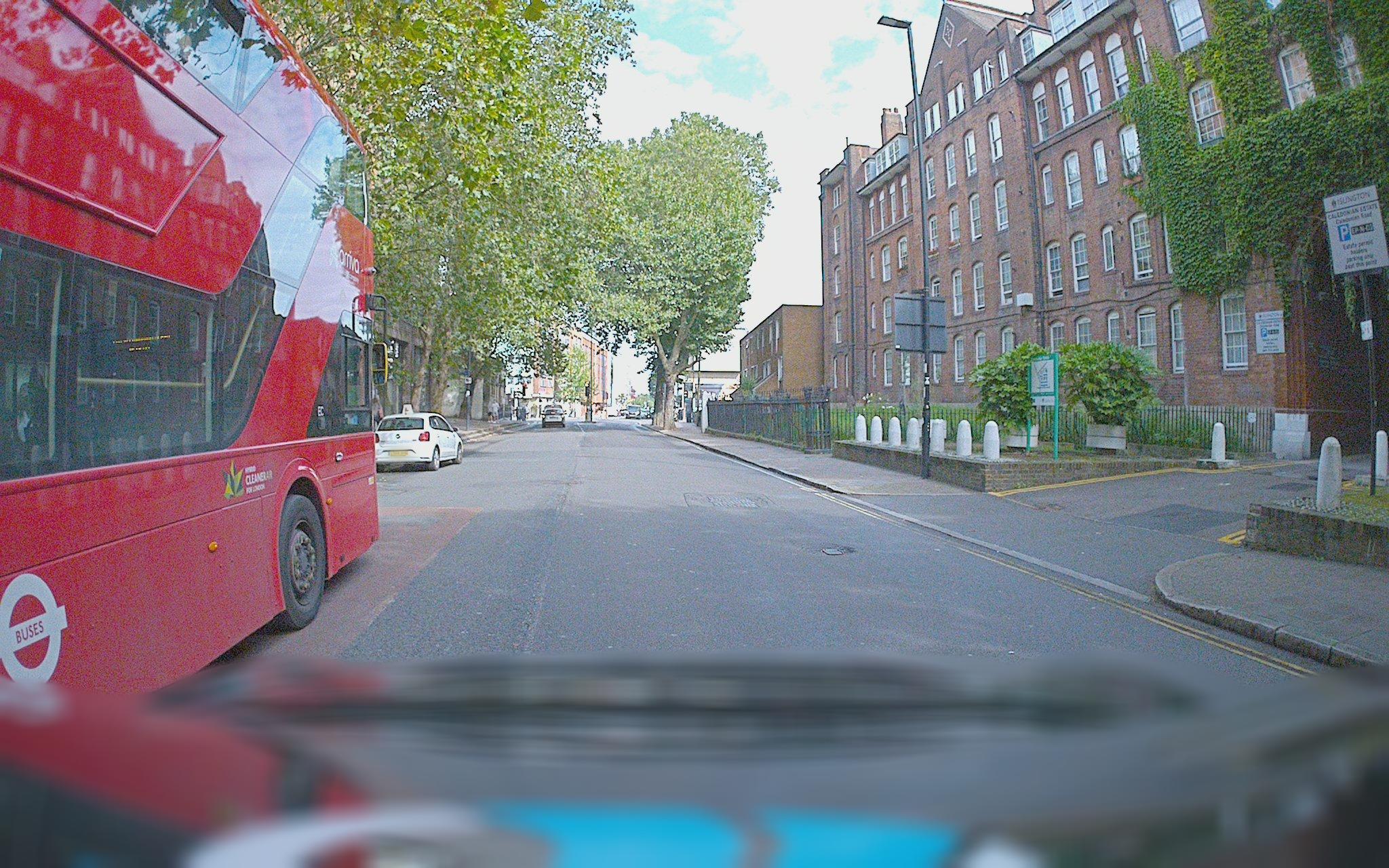}
    {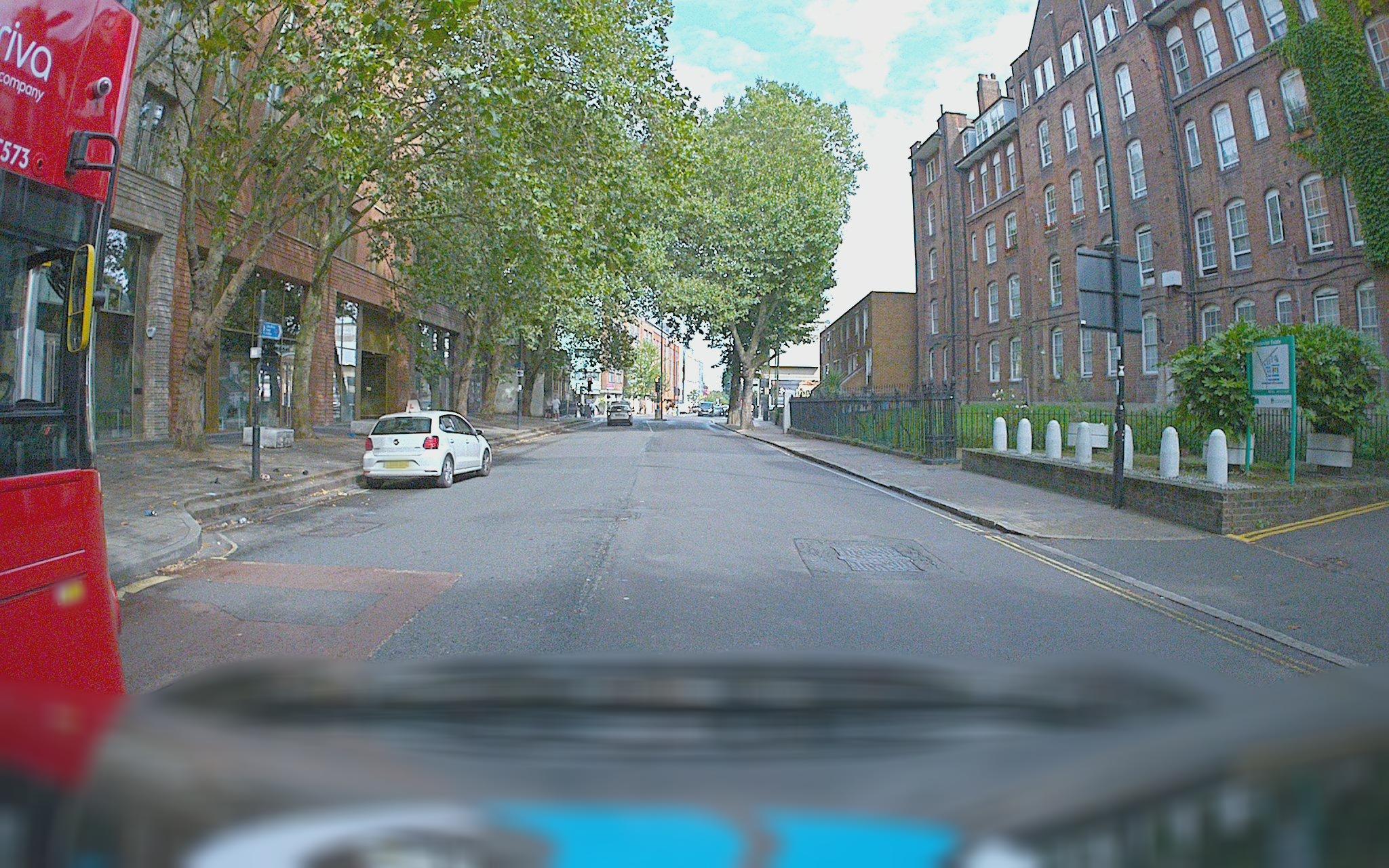}
    {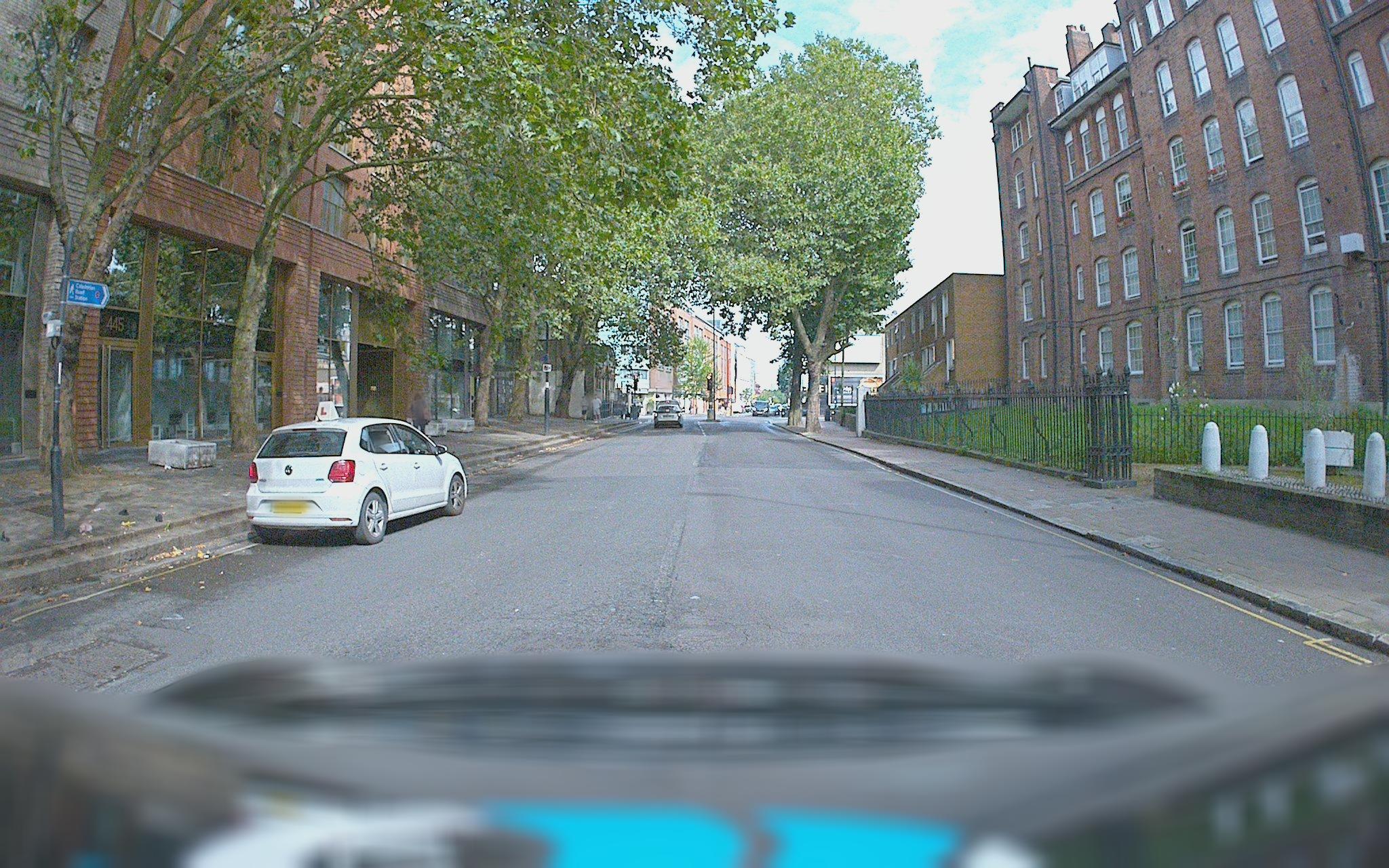}
    {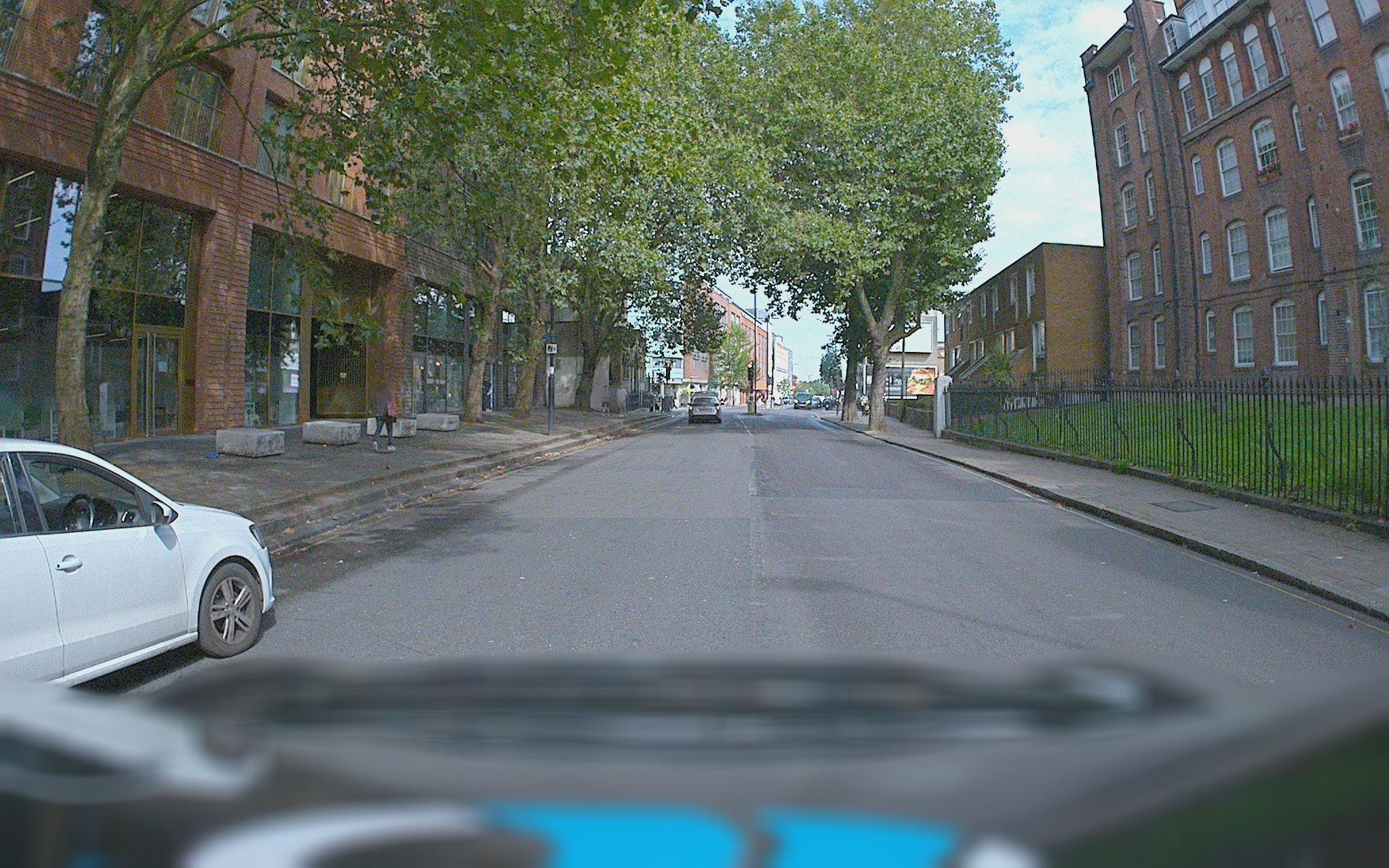}
    {You are a mature and cautious driver with many years of driving experience. You now need assistance answering questions related to the driving process. \par Output only the final answer.}
    {What actions are you taking currently?}
    {I am trying to safely overtake a stopped bus on its right side and moving to the left of the road to be correctly positioned.}
    {R1-4 Situation Description}

    \vspace{0.5em}
    \noindent\rule{0.97\linewidth}{0.4pt}
    \caption{Representative example from Rank 1-4 Questions.}
    \label{fig:r1_4}
\end{figure}

\begin{figure}[!htbp]
    \centering

    \rowexamplesix
    {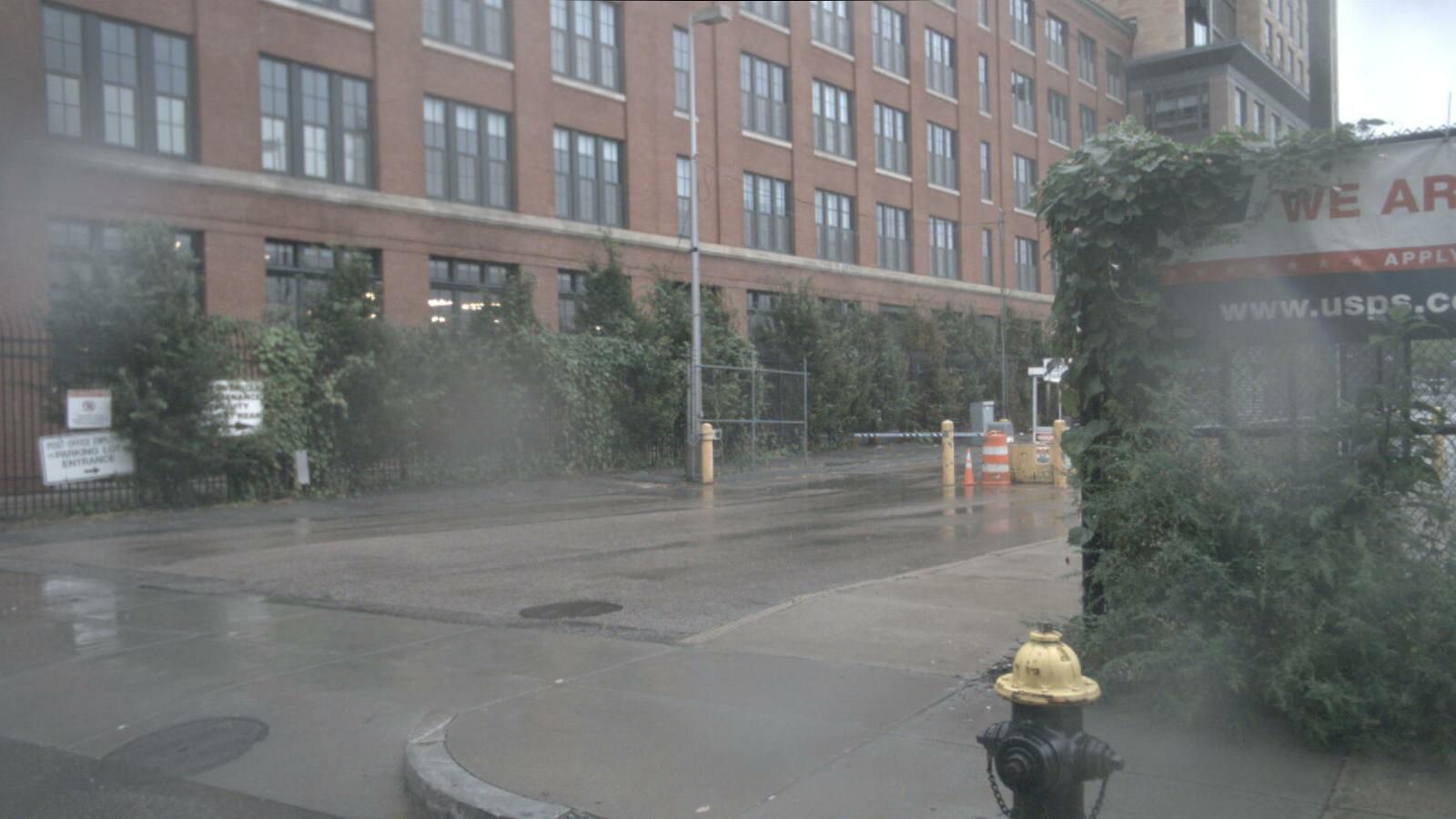}
    {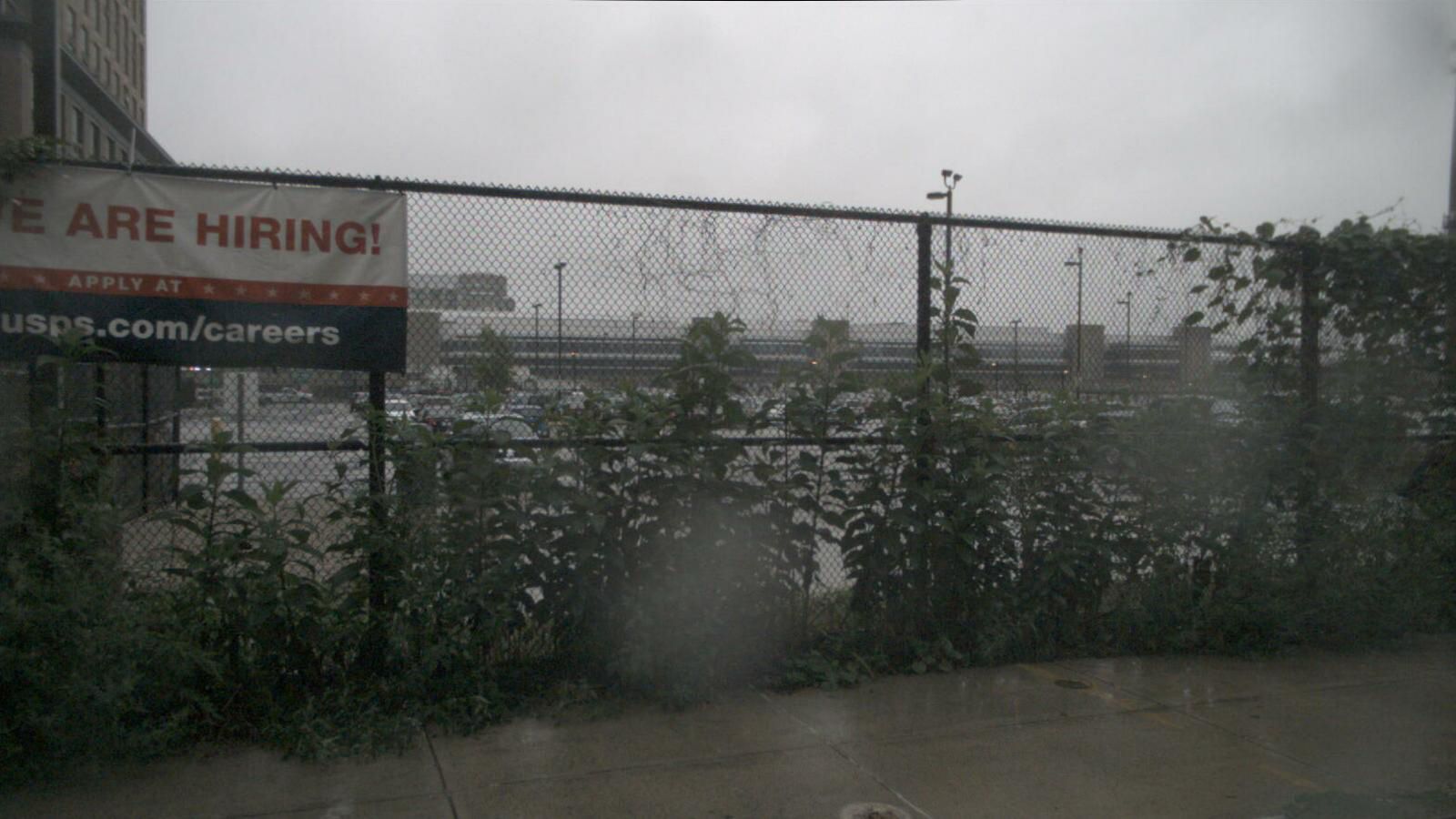}
    {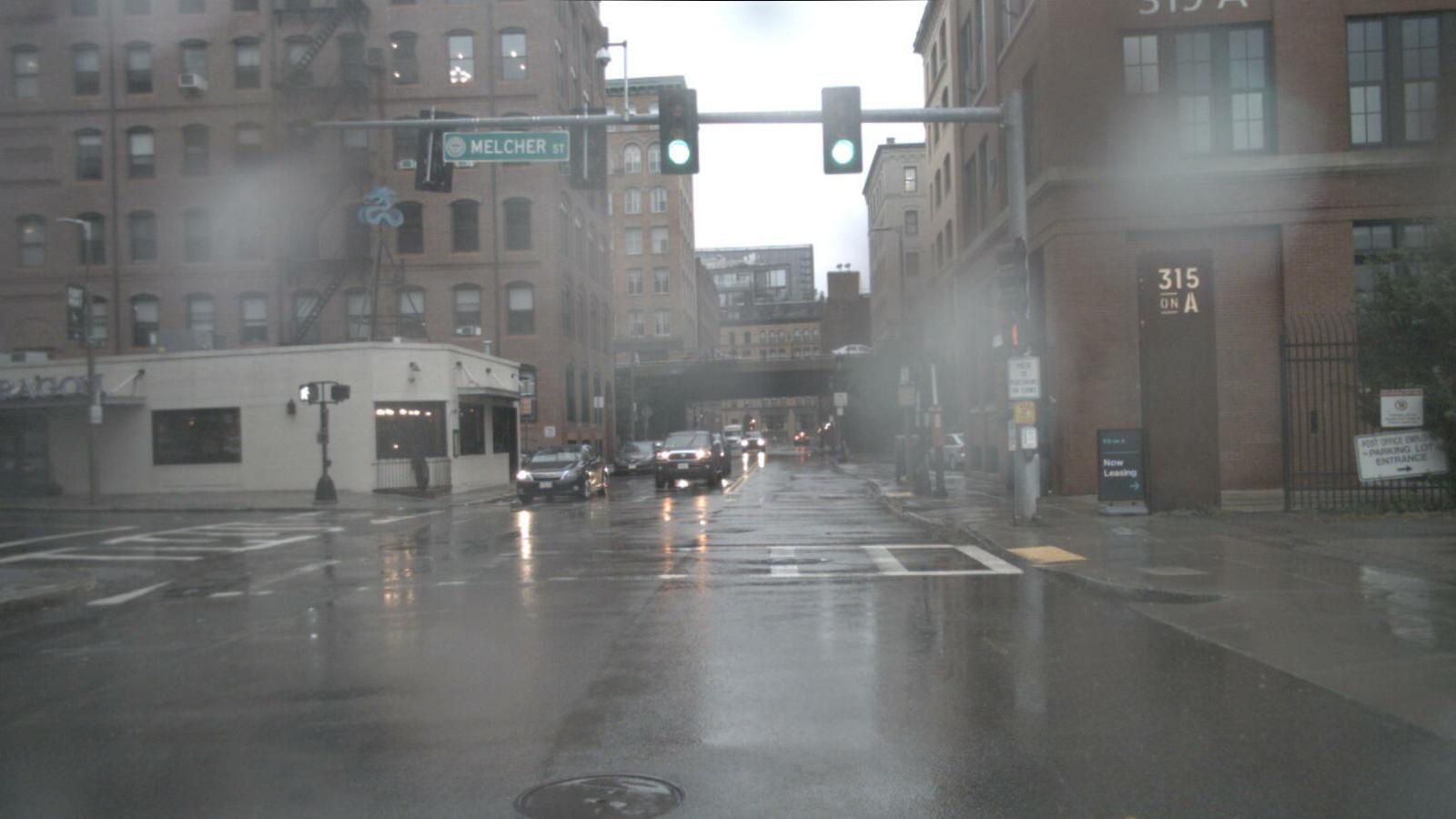}
    {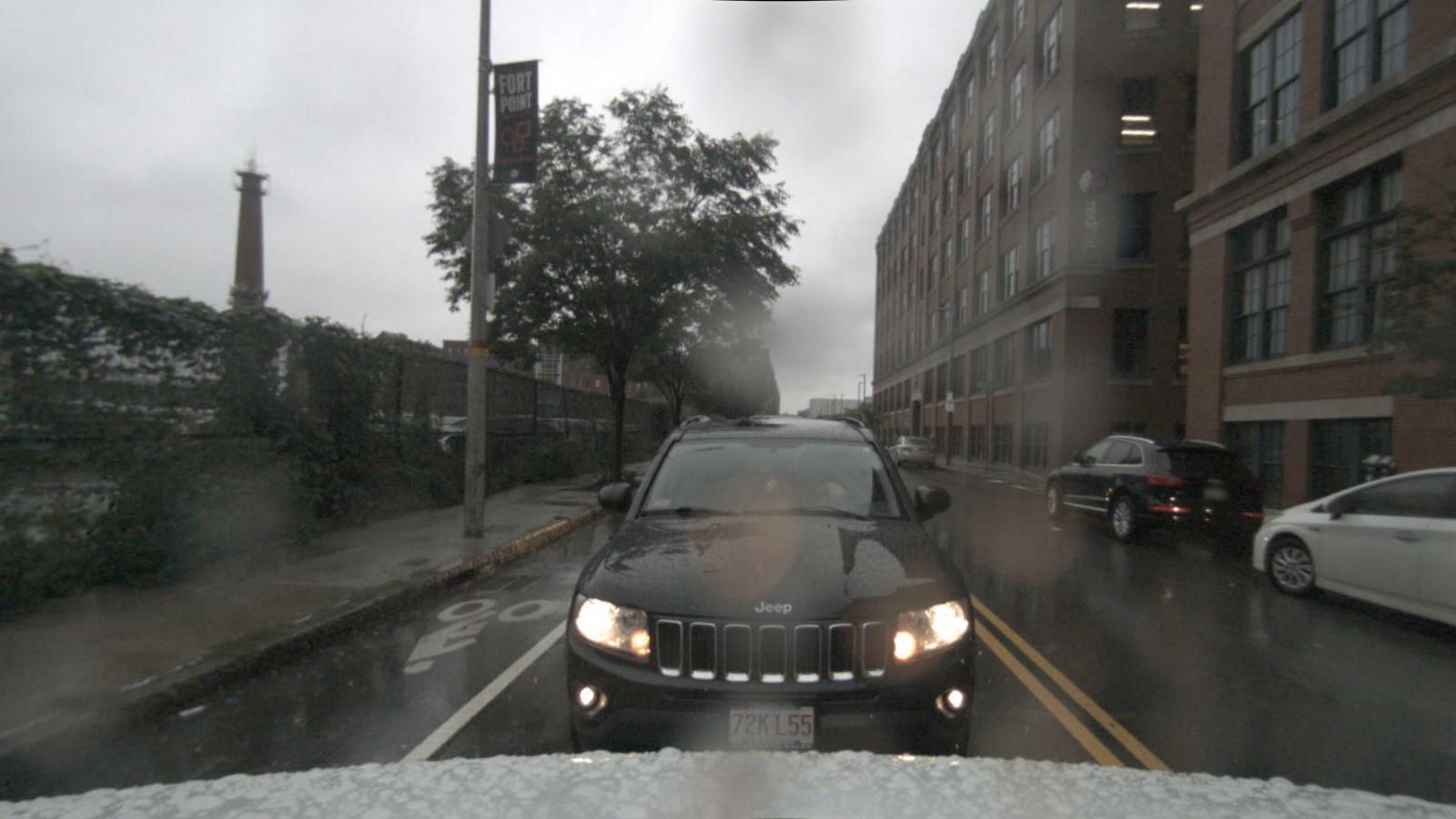}
    {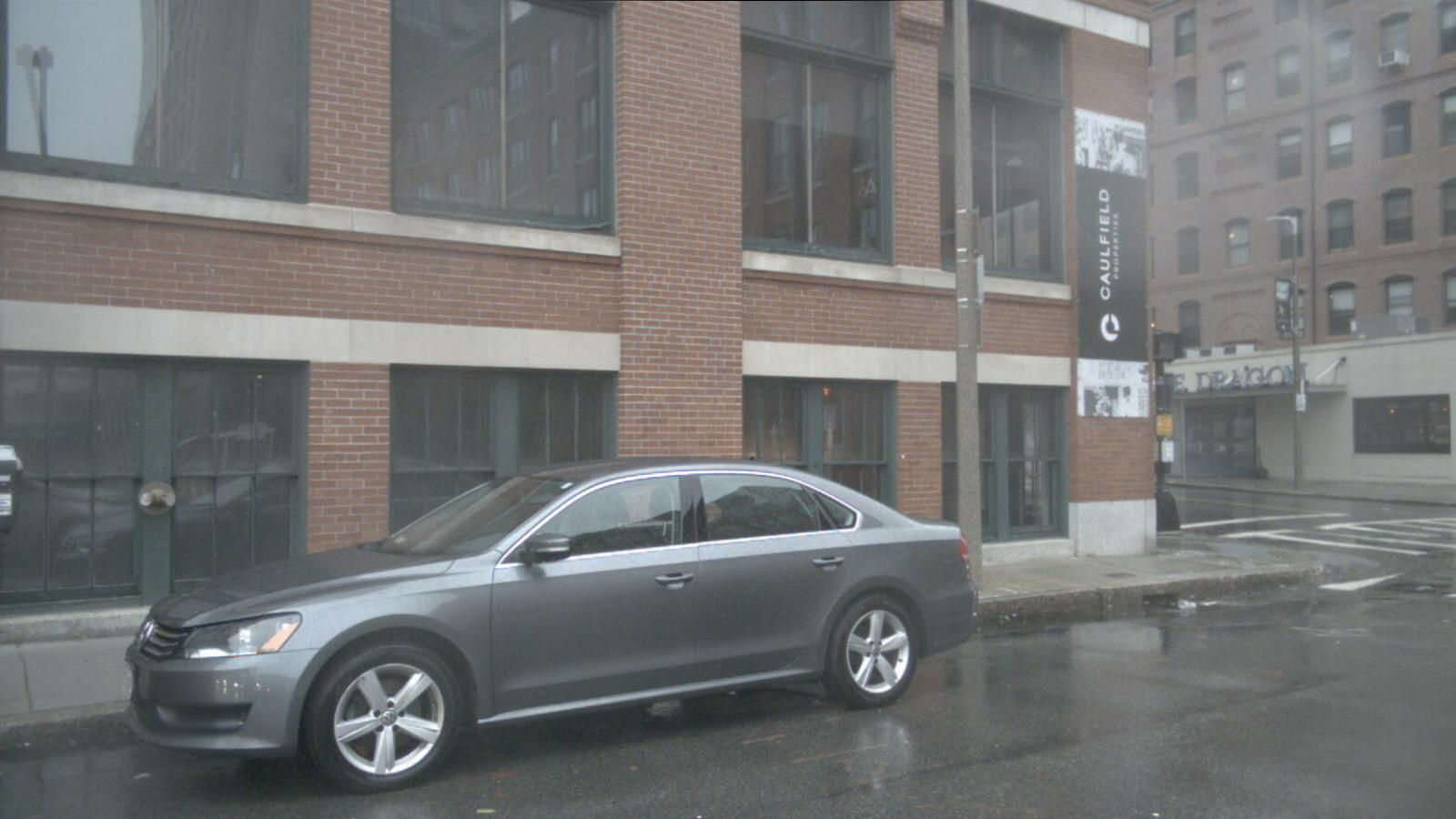}
    {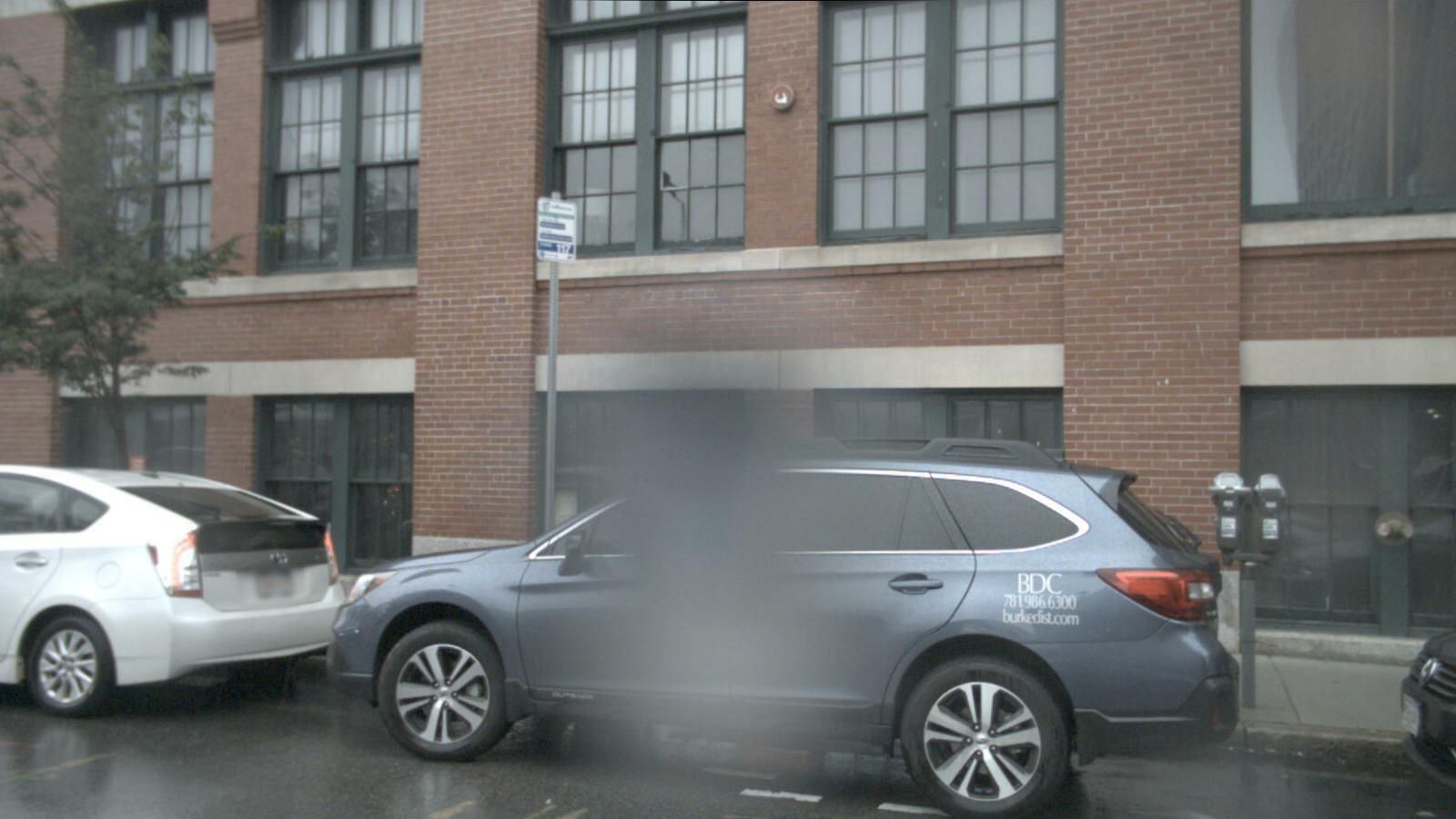}
    {
    {\small \textbf{R2-1 Multi-view Memory}\par}
    \vspace{0.4em}
    {\footnotesize \textbf{Sys. Prompt:} You are a mature and cautious driver with many years of driving experience. You now need assistance answering questions related to the driving process. \par Answer with only the numeric result (integer). \par}
    \vspace{0.4em}
    {\footnotesize \textbf{Prompt:} Given the 6 surround-view images (Front, Front-Left, Front-Right, Back, Back-Left, Back-Right) at this timestamp, how many unique vehicles are visible in the surrounding environment? \par}
    \vspace{0.4em}
    {\footnotesize \textbf{Answer:} 10.\par}
    }

    \vspace{0.5em}
    \noindent\rule{0.97\linewidth}{0.4pt}
    \caption{Representative example from Rank 2-1 Questions.}
    \label{fig:r2_1}
\end{figure}

\begin{figure}[!htbp]
    \centering

    \rowexampleblock
    {
        \imgpair{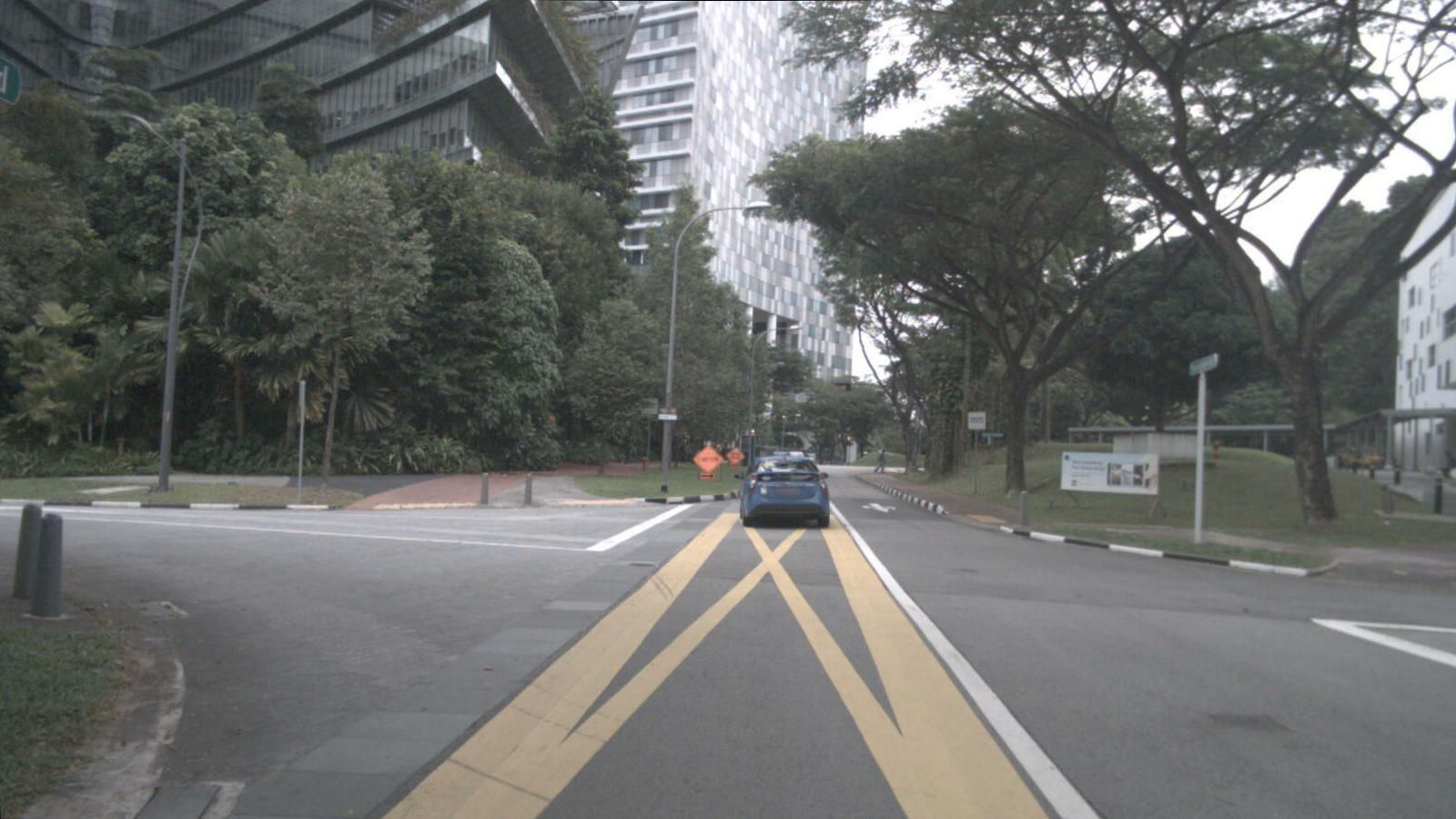}{$t_0$}{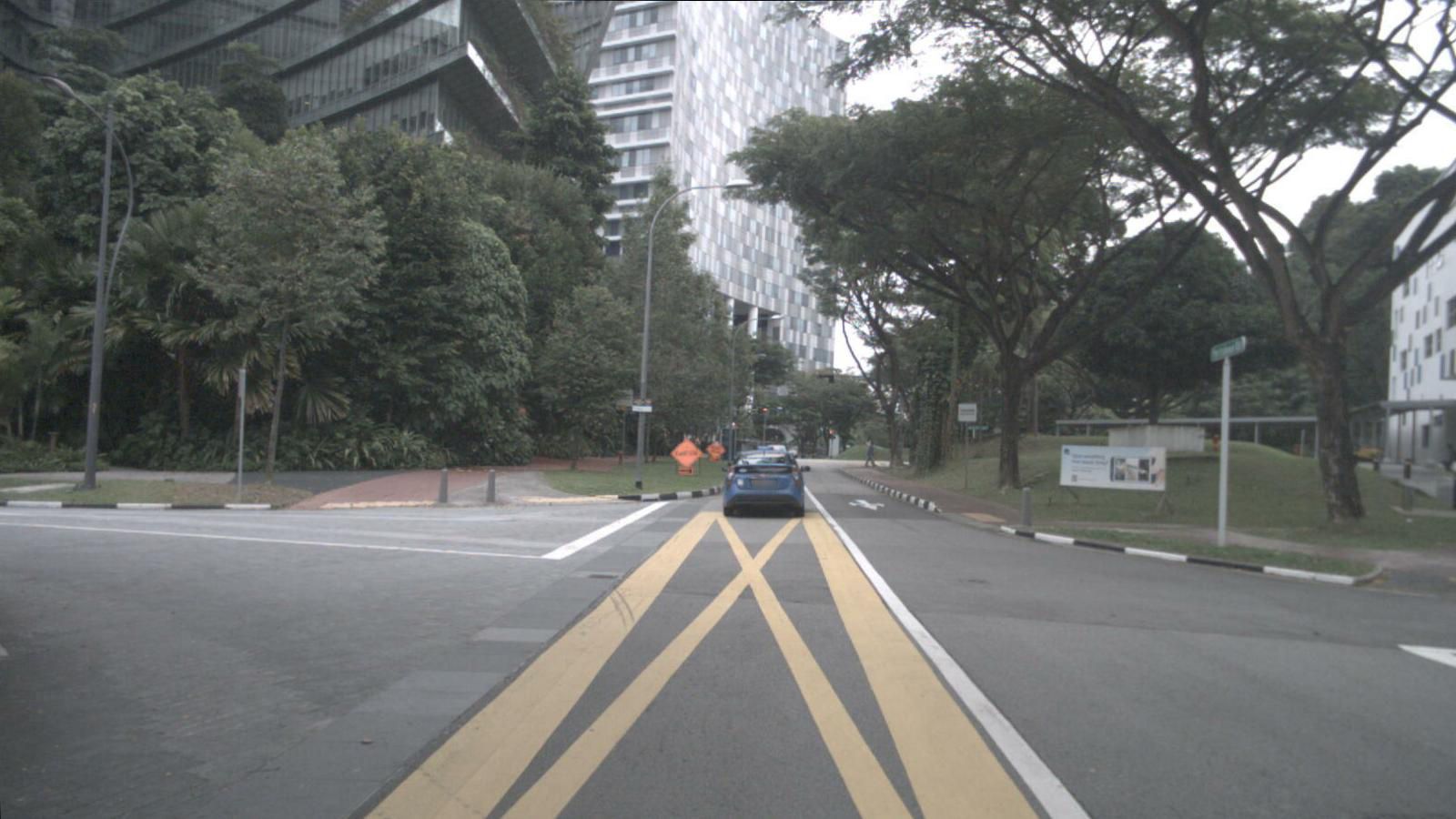}{$t_1$}
        \vspace{0.4em}

        \imgpair{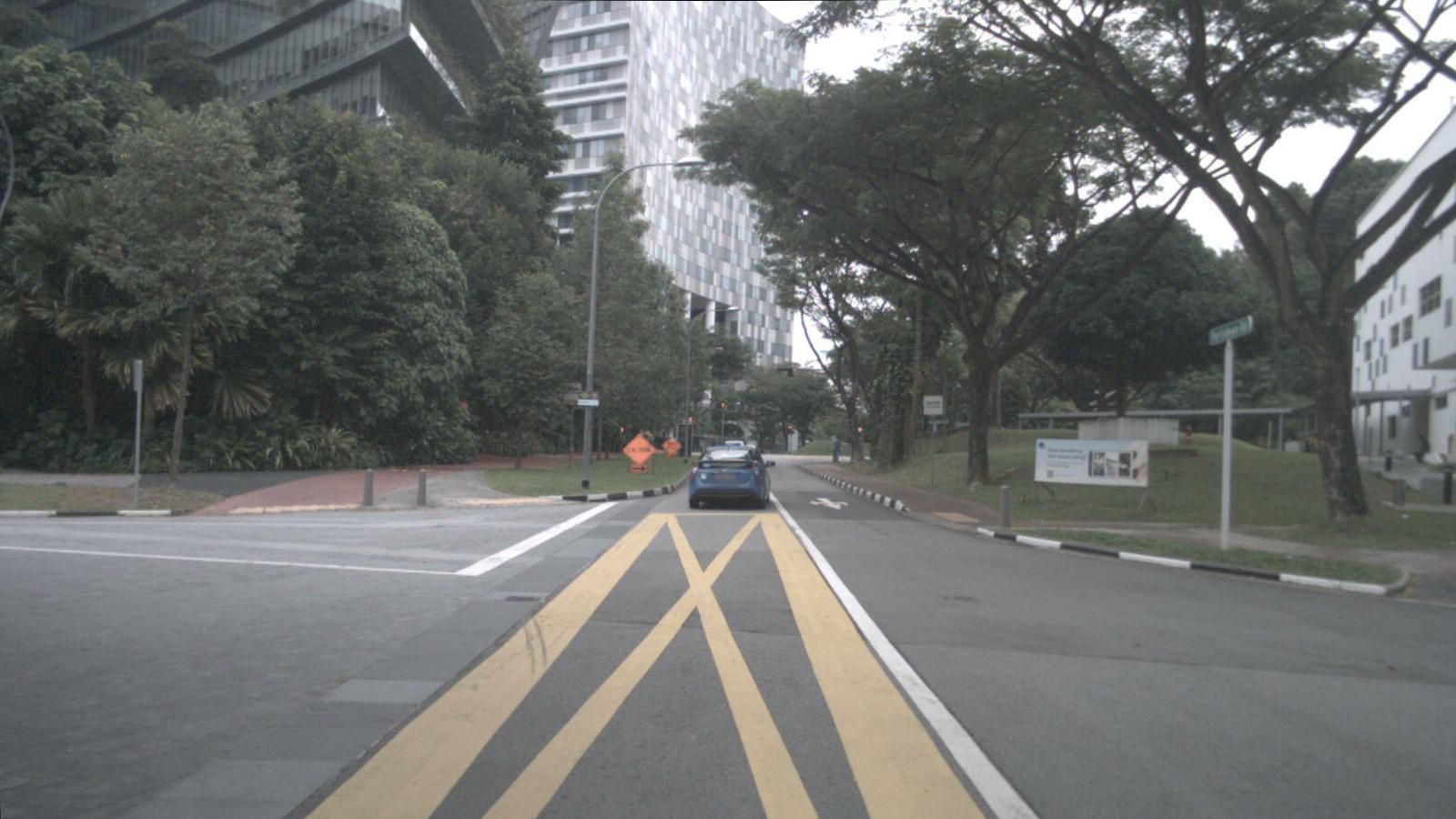}{$t_2$}{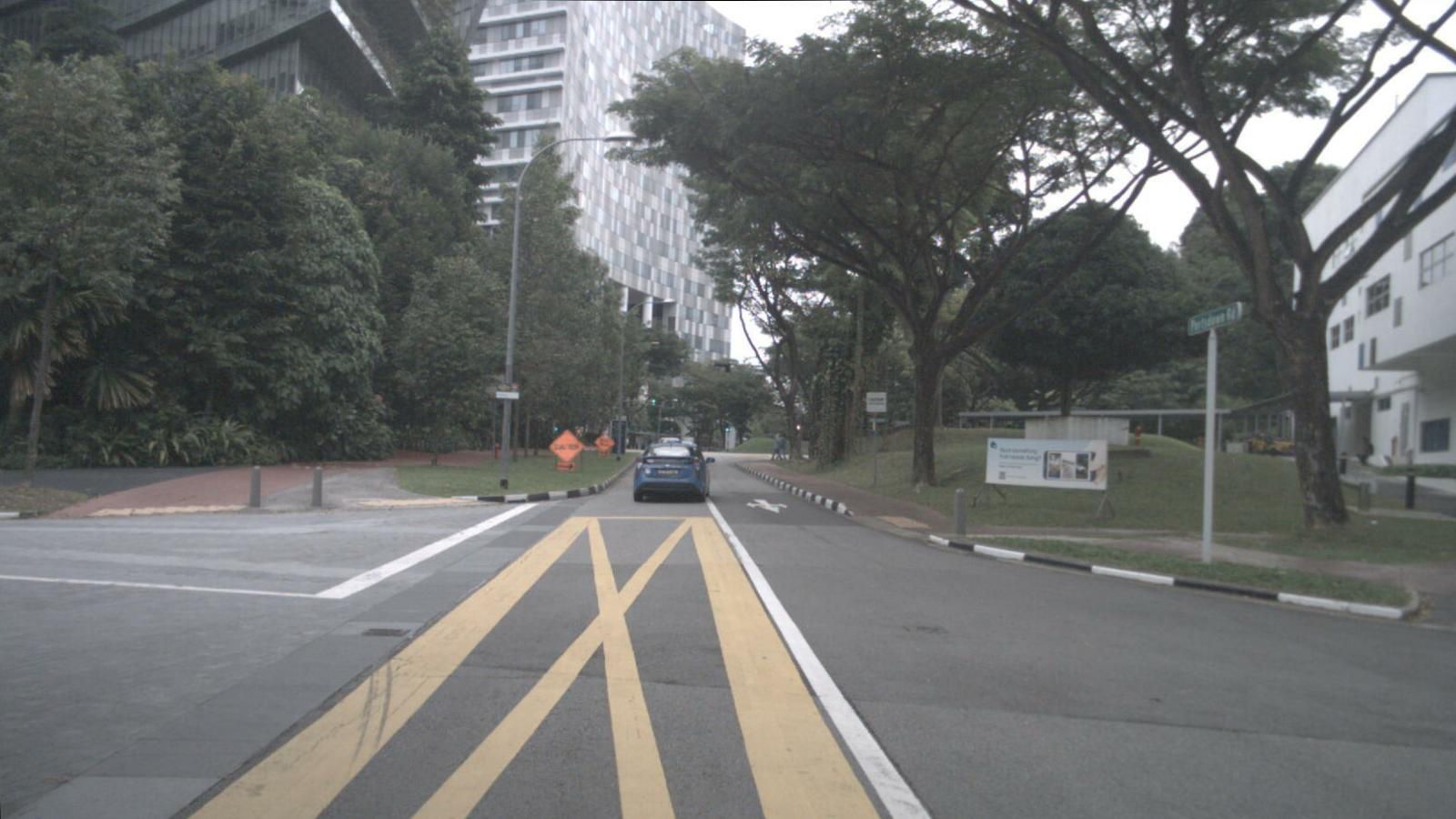}{$t_3$}
        \vspace{0.4em}

        \imgpair{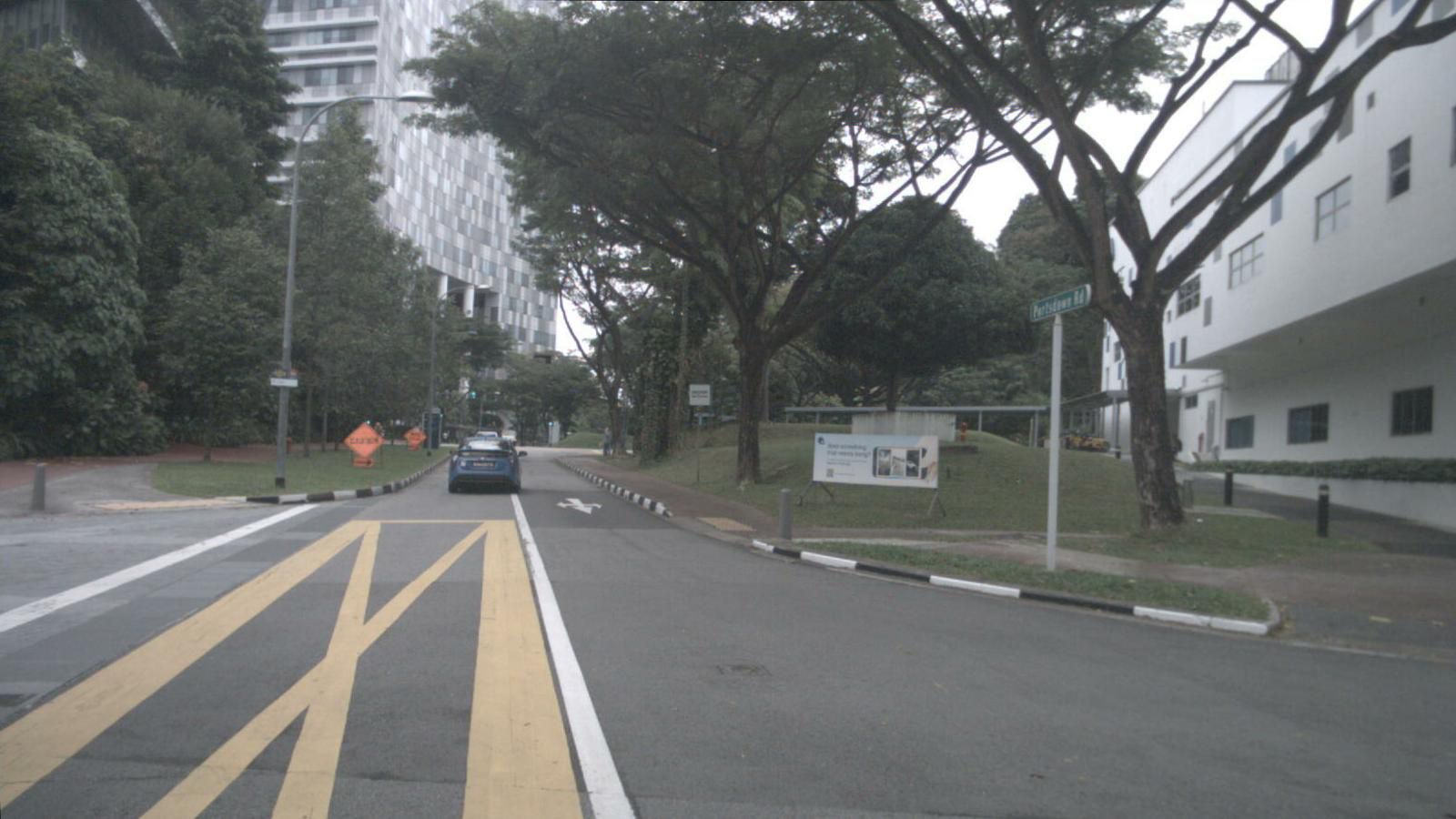}{$t_4$}{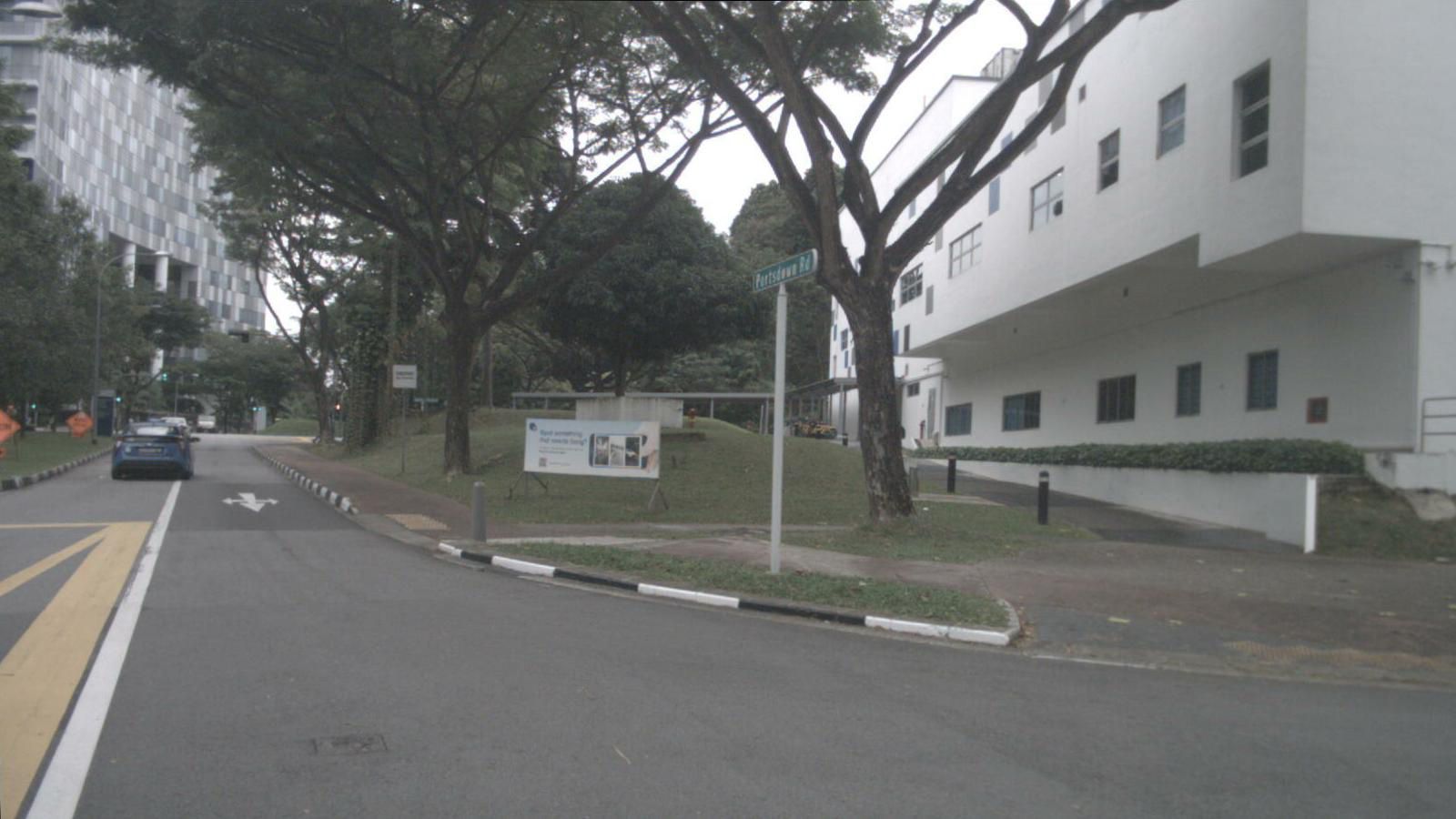}{$t_5$}
        \vspace{0.4em}

        \imgpair{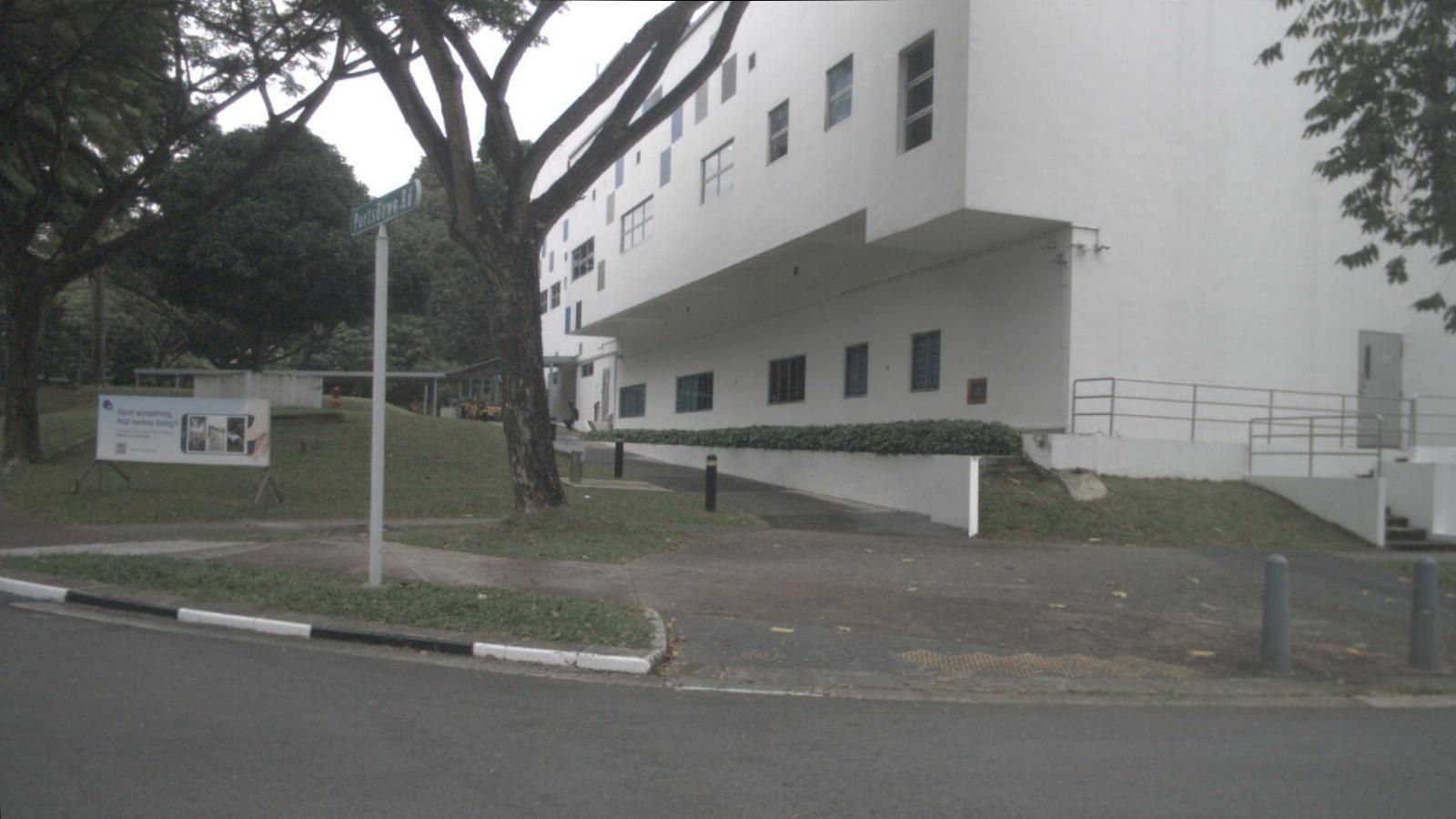}{$t_6$}{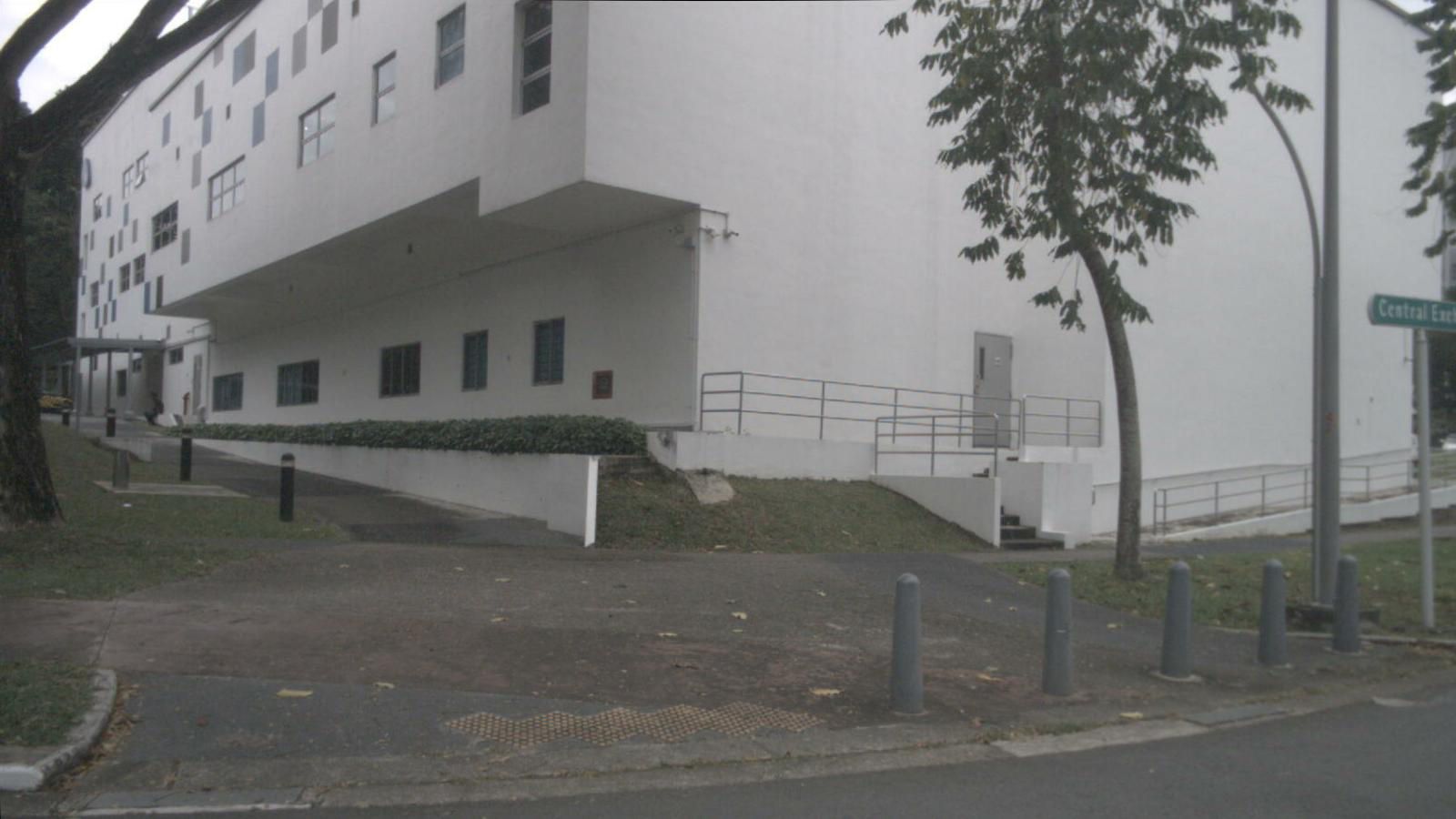}{$t_7$}
        \vspace{0.4em}

        \imgpair{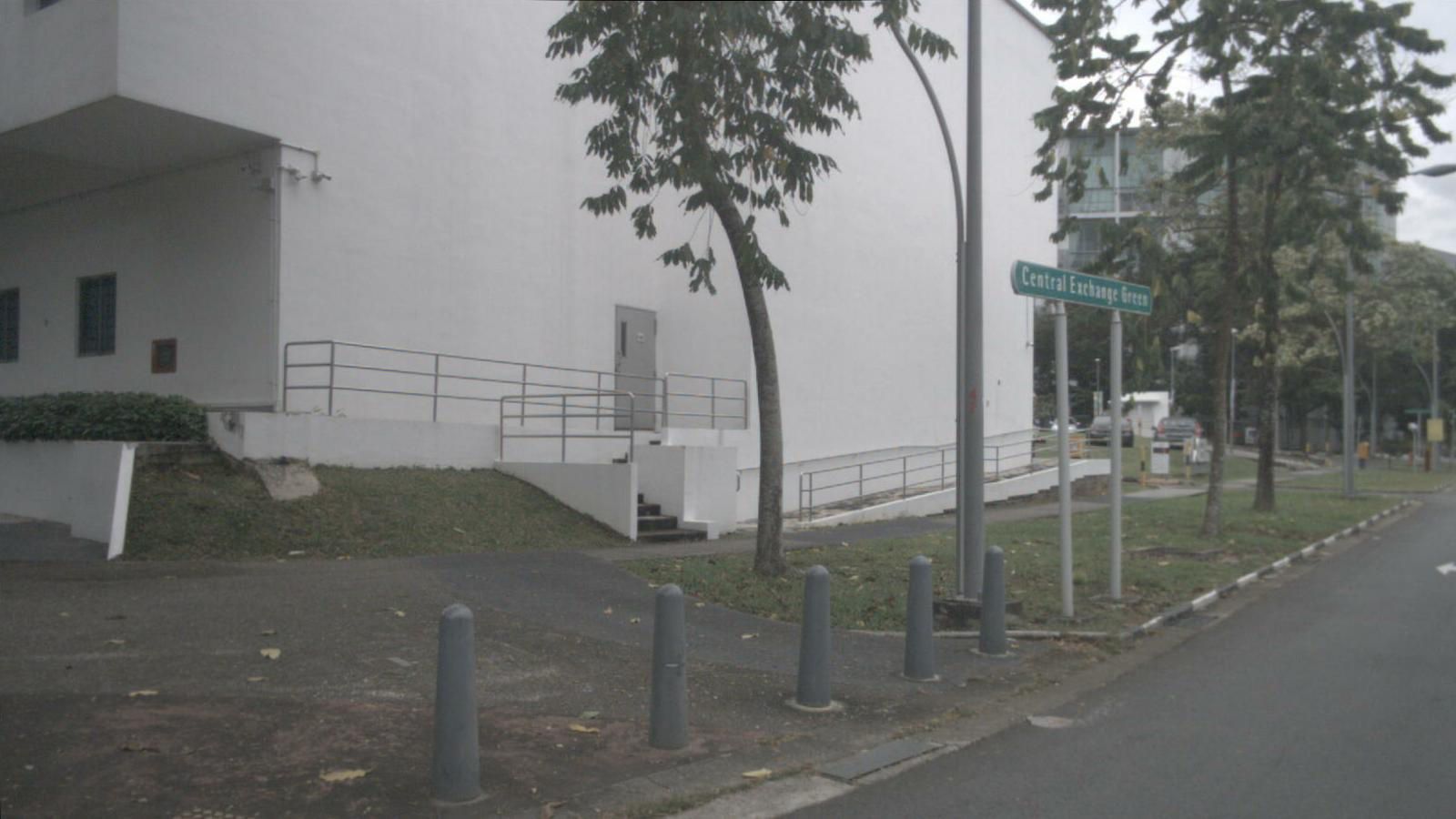}{$t_8$}{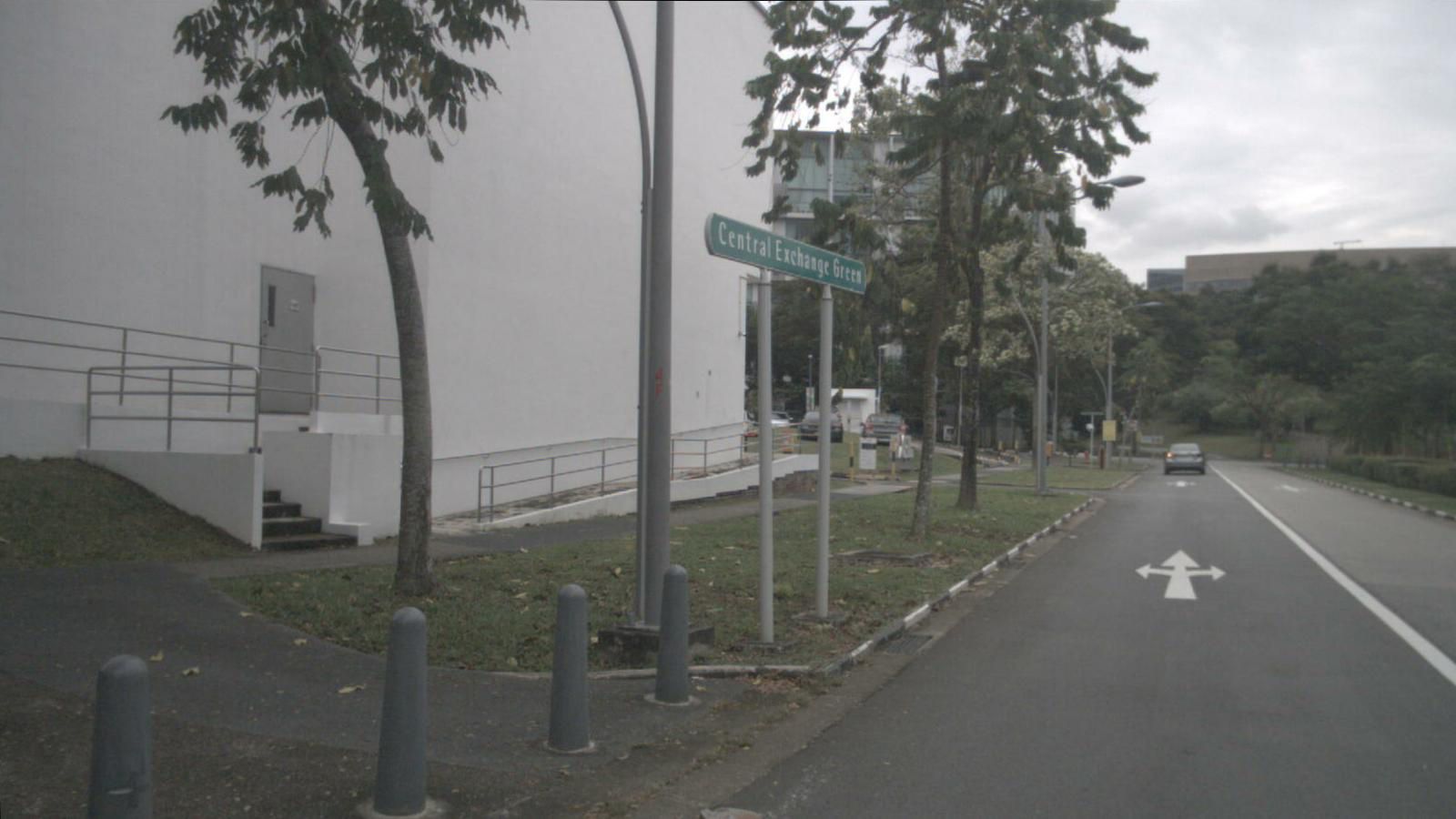}{$t_9$}
    }
    {
        \qatextblock
        {R2-2-A Temporal Counting}
        {You are a mature and cautious driver with many years of driving experience. You now need assistance answering questions related to the driving process. \par Answer with only the numeric result (integer).}
        {Observing the sequence starting from this frame for the next 5.0 seconds (including the current frame), how many unique vehicles appear in CAM\_FRONT?}
        {7.}
        }

    \vspace{0.5em}
    \noindent\rule{0.97\linewidth}{0.4pt}
    \caption{Representative example from Rank 2-2-A Questions.}
    \label{fig:r2_2_a}
\end{figure}

\begin{figure}[!htbp]
    \centering

    \rowexampleblock
    {
        \imgpair{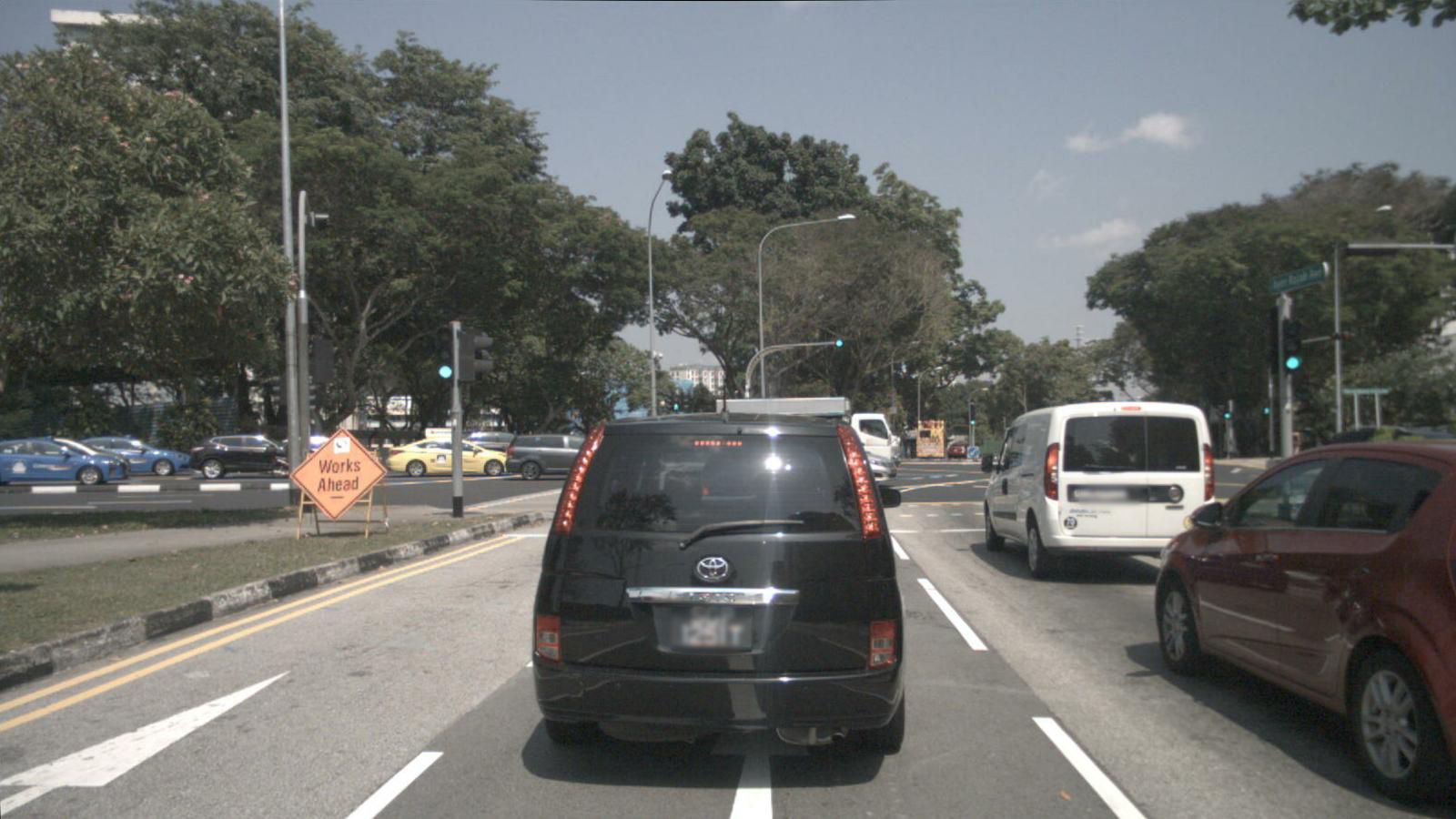}{$t_0$}{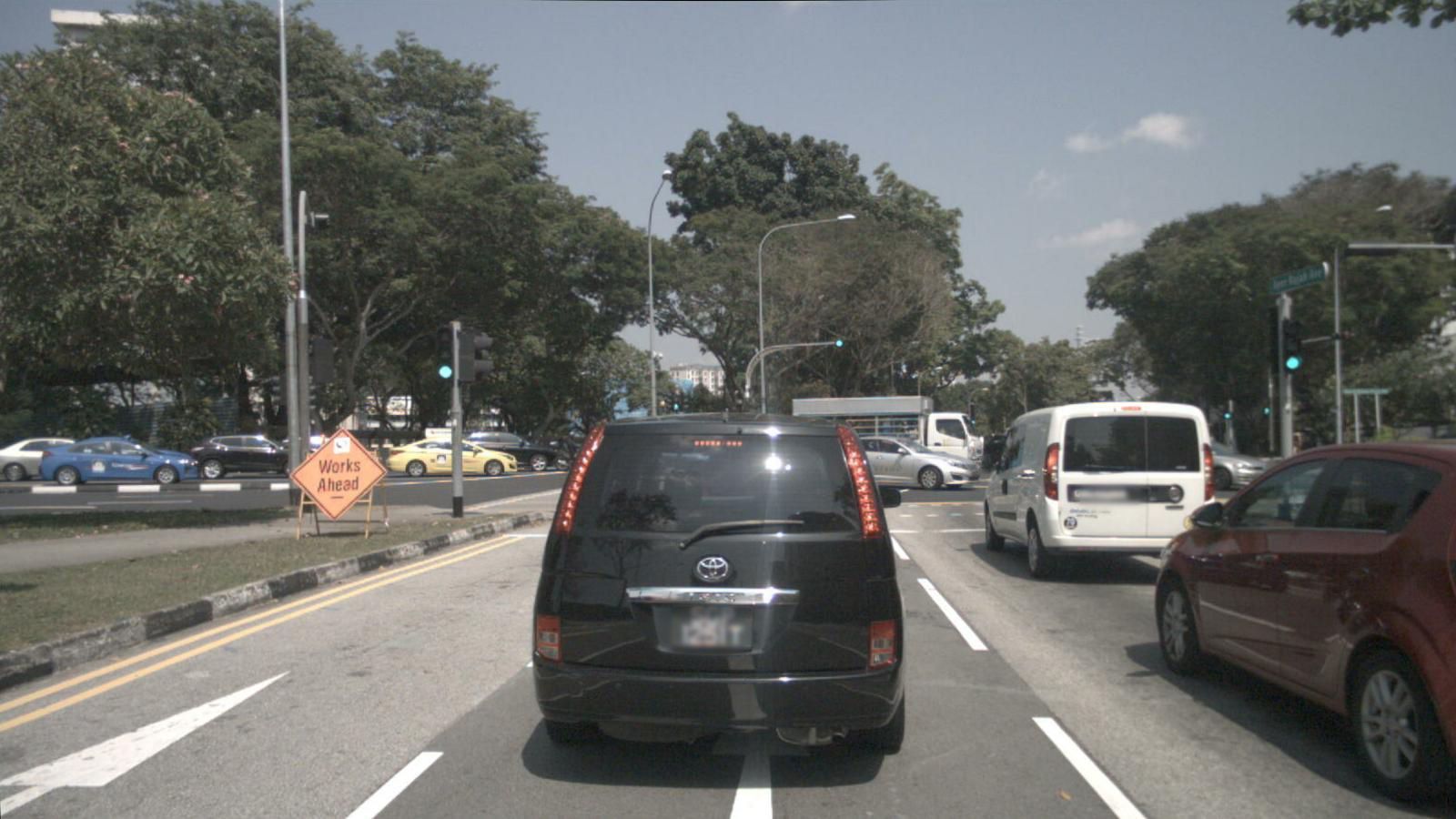}{$t_1$}
        \vspace{0.4em}

        \imgpair{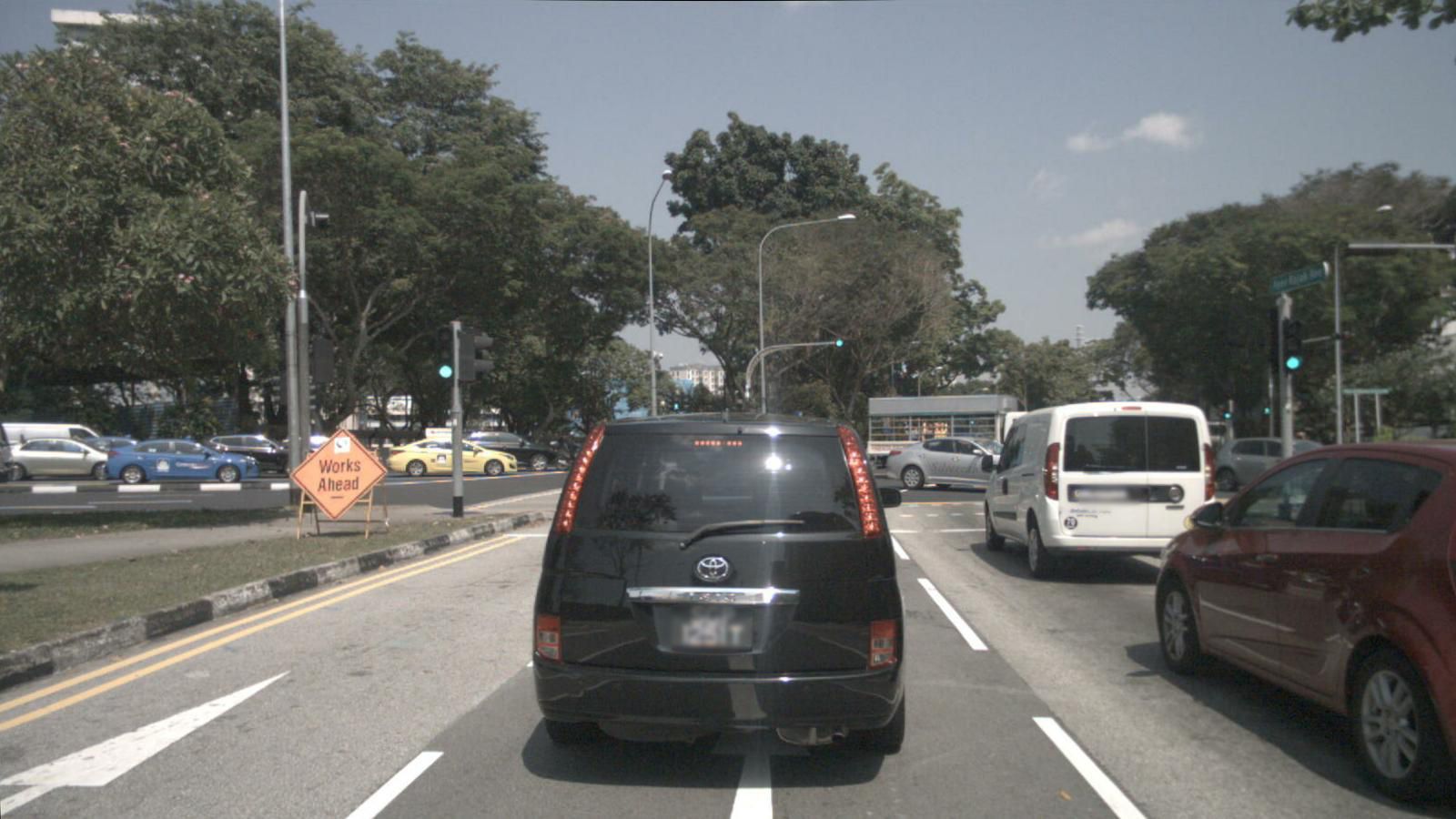}{$t_2$}{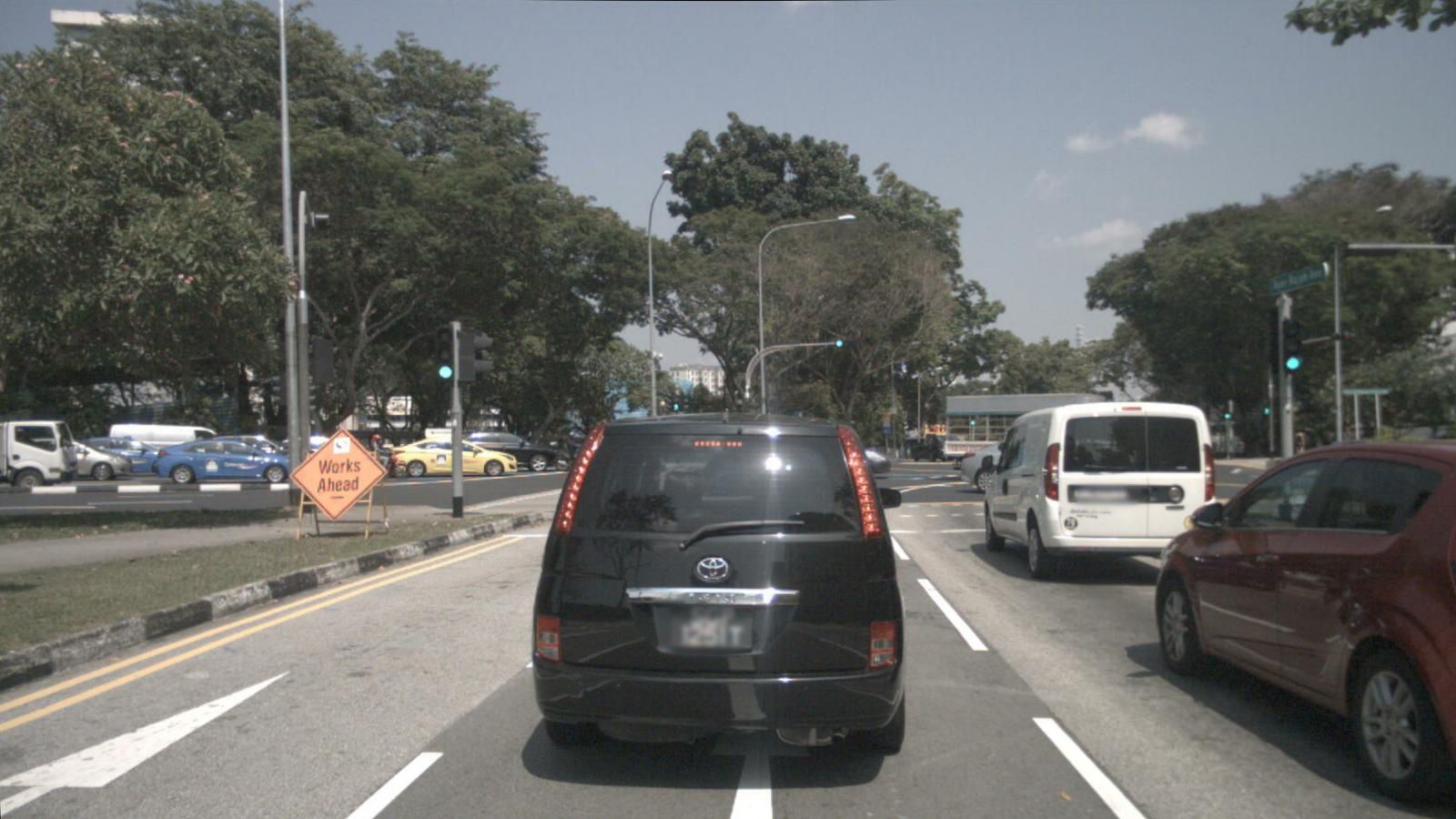}{$t_3$}
        \vspace{0.4em}

        \imgpair{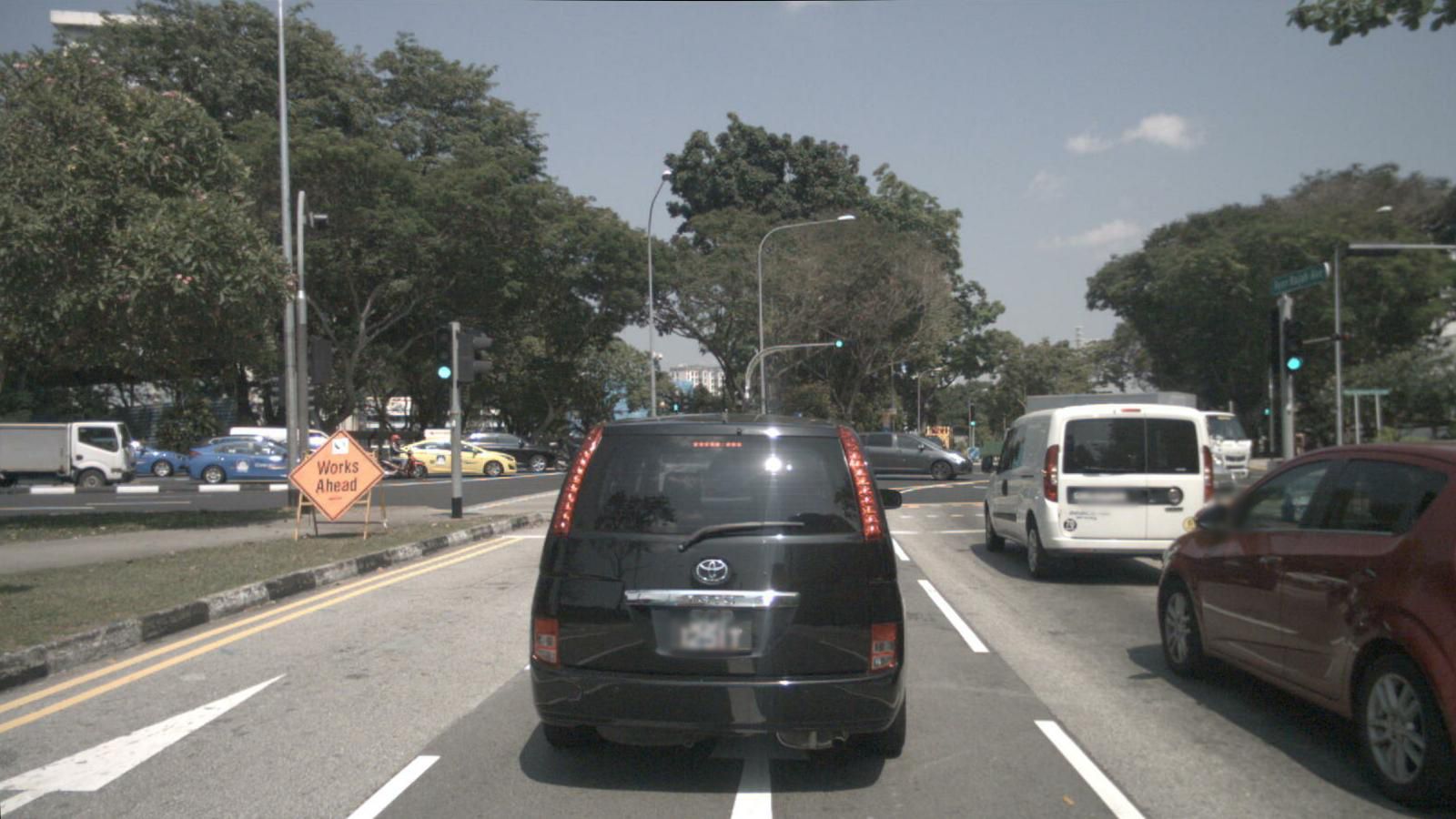}{$t_4$}{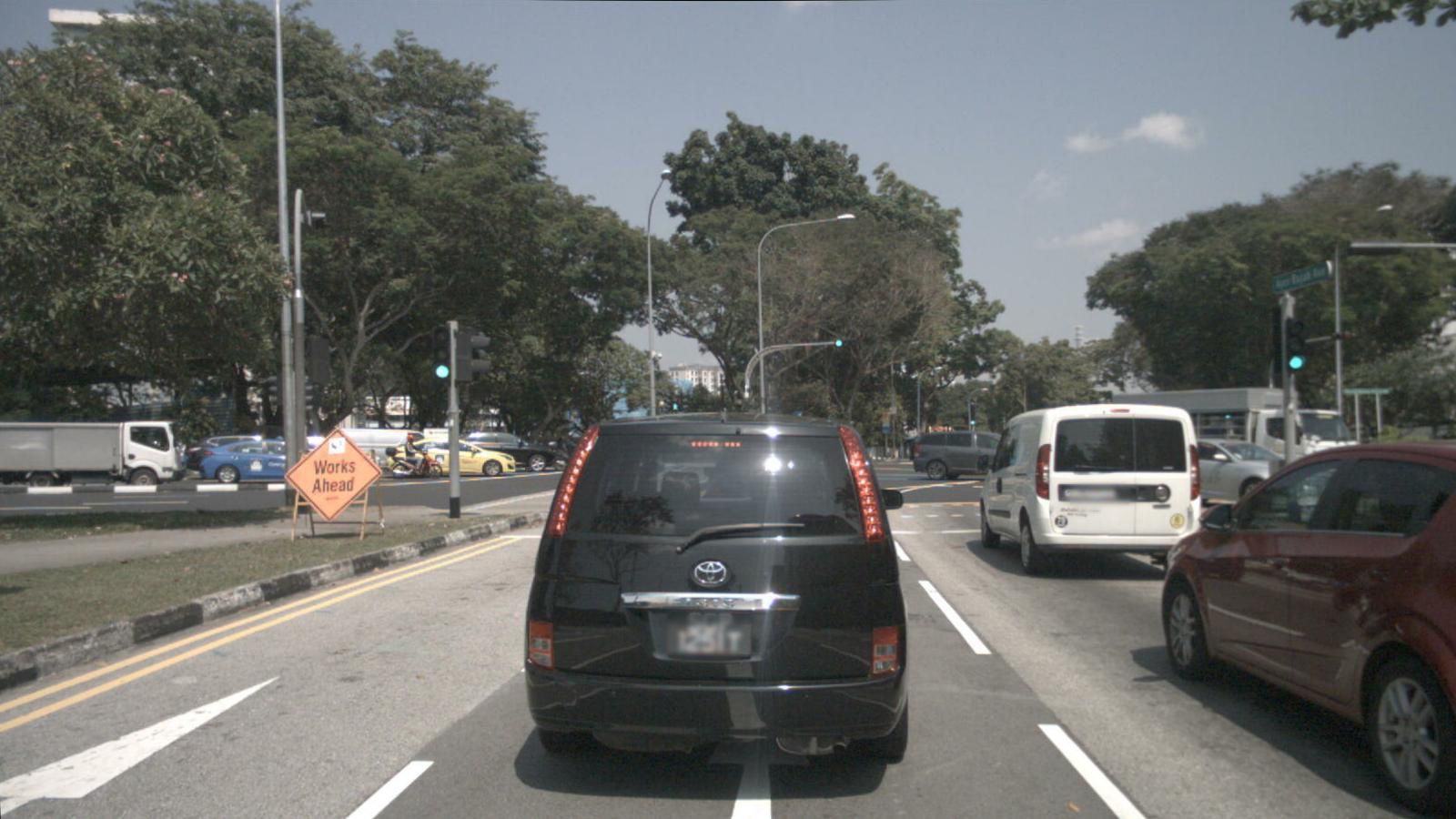}{$t_5$}
        \vspace{0.4em}

        \imgpair{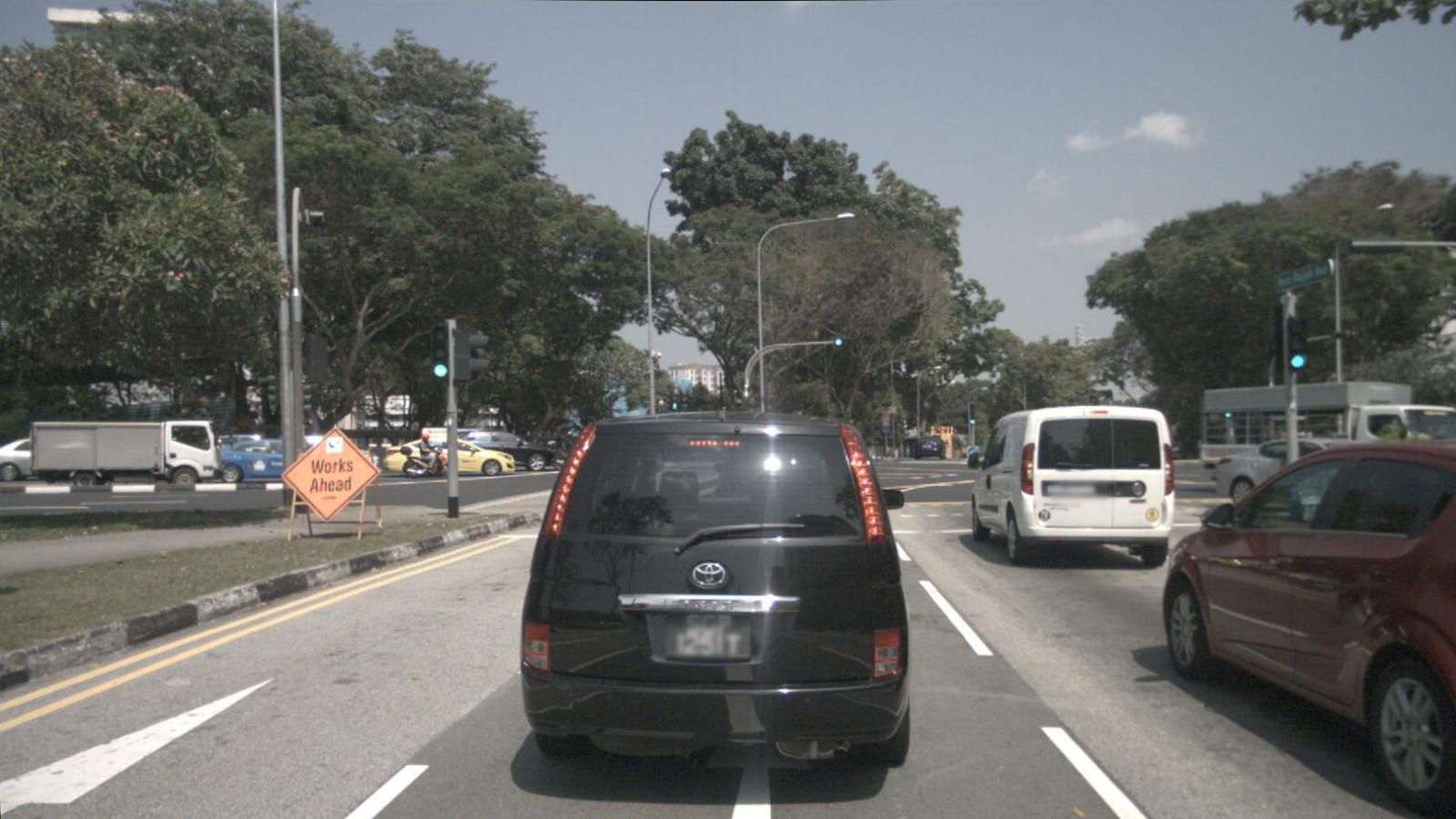}{$t_6$}{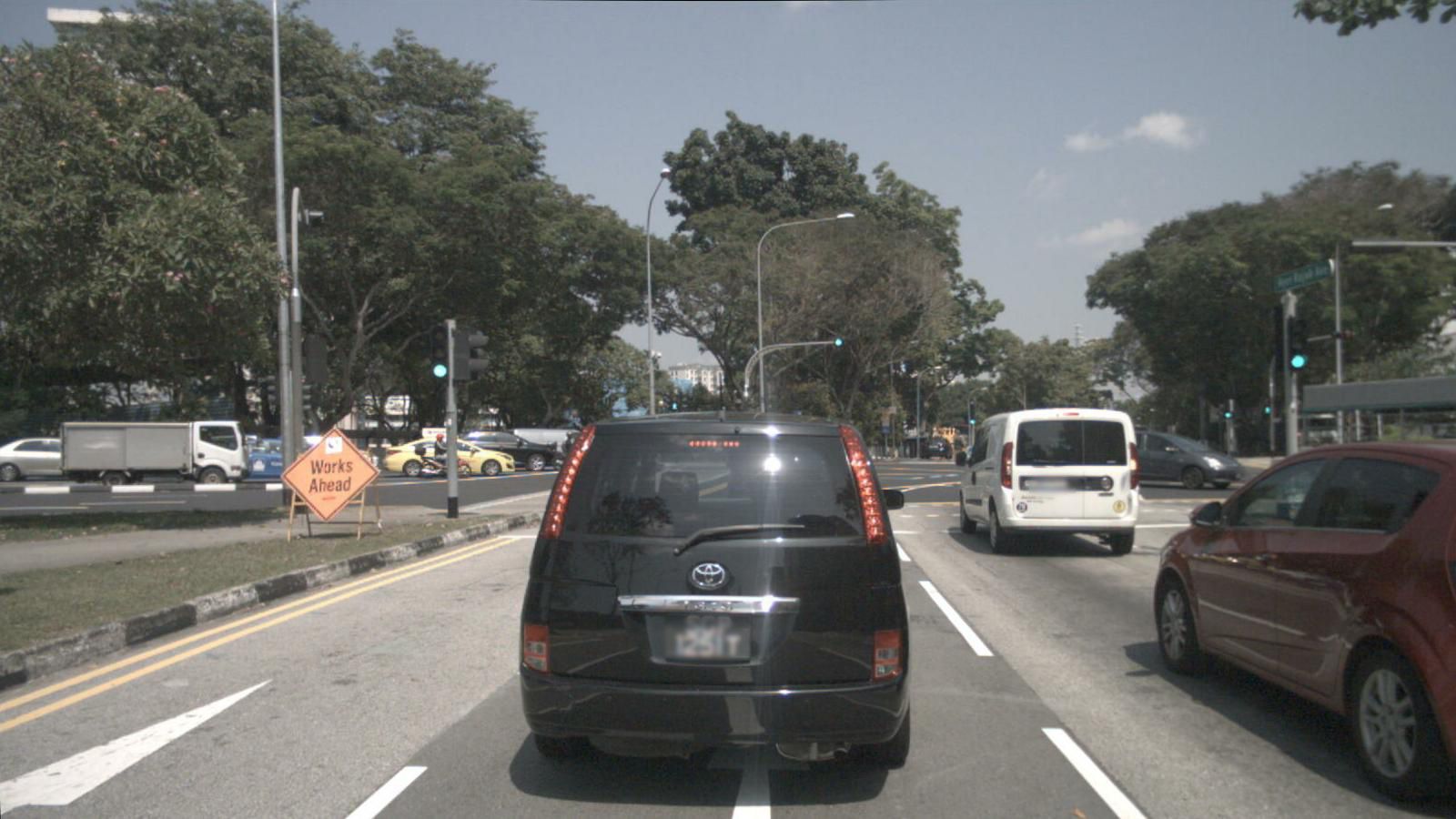}{$t_5$}
    }
    {
        \qatextblock
        {R2-2-B Temporal Status Recognition}
        {You are a mature and cautious driver with many years of driving experience. You now need assistance answering questions related to the driving process. \par Choose the correct option and output only the option letter (e.g., A/B/C/D).}
        {Focus on the Vehicle at pixel coordinates [1599, 714] in the initial frame (Image resolution: 1600x900), located in the lower-right region. Observing the clip from t=0s to t=3.5s, how does its state change? \par ["A. It transitions from parked to moving.", "B. It transitions from stopped to moving.", "C. No Change (It remains stopped).", "D. No Change (It remains parked)."]}
        {B. It transitions from stopped to moving.}
    }

    \vspace{0.5em}
    \noindent\rule{0.97\linewidth}{0.4pt}
    \caption{Representative example from Rank 2-2-B Questions.}
    \label{fig:r2_2_B}
\end{figure}

\begin{figure}[!htbp]
    \centering
    
    \rowexample
    {\includegraphics[width=0.95\linewidth]{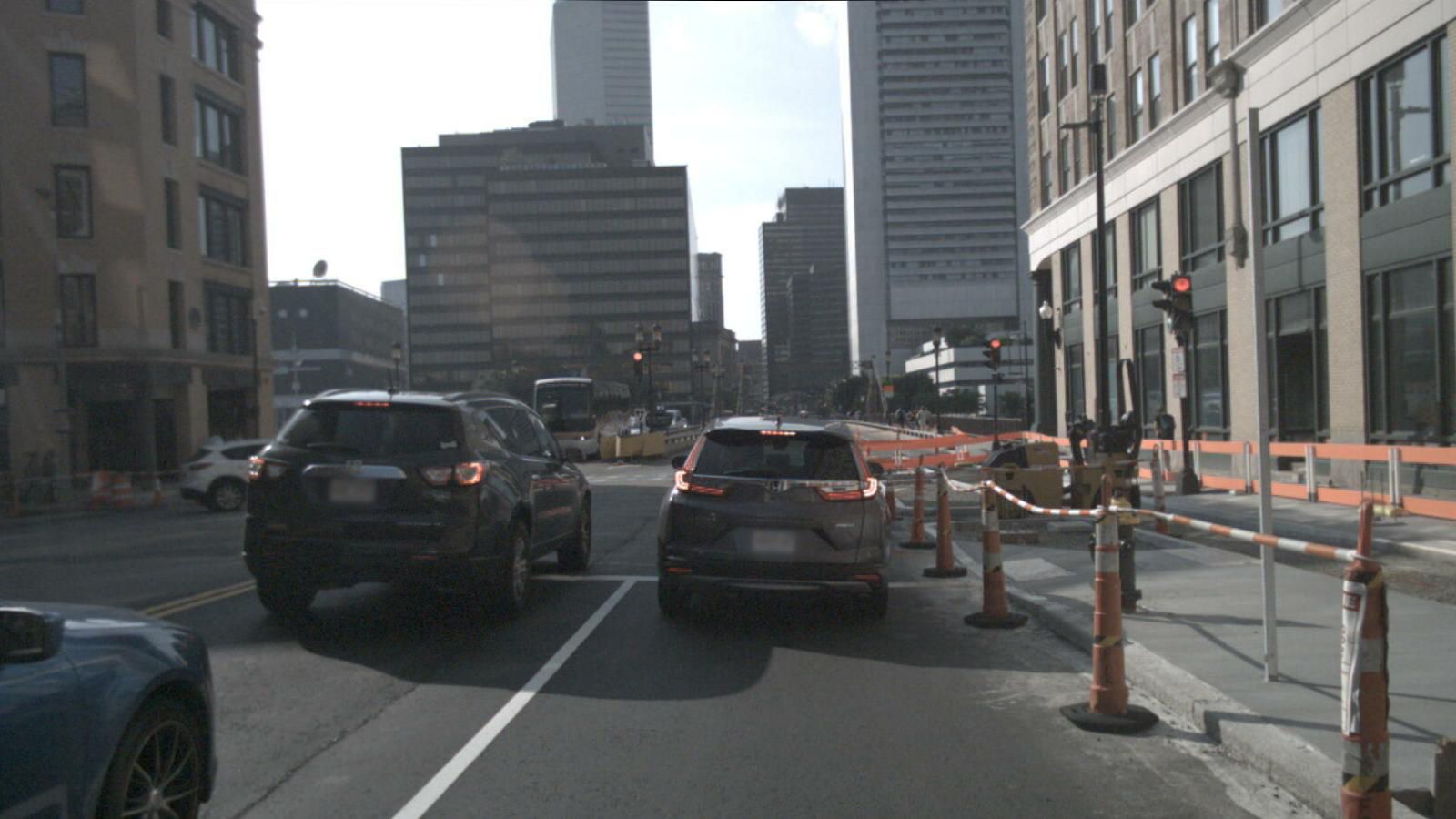}}
    {You are a mature and cautious driver with many years of driving experience. You now need assistance answering questions related to the driving process. \par Choose the correct option and output only the option letter (e.g., A/B/C/D).}
    {In CAM\_FRONT, what is the position of the Vehicle located at pixel coordinates [860, 549] relative to the Vehicle located at pixel coordinates [500, 536] (Image resolution: 1600x900. Using the car as a reference frame)? \par ["A. Left.", "B. Rear-Left.", "C. Rear-Right.", "D. Front-Right."]}
    {D. Front-Right.}
    {R2-3 Spatial Relations}

    \vspace{0.5em}
    \noindent\rule{0.97\linewidth}{0.4pt}
    \caption{Representative example from Rank 2-3 Questions.}
    \label{fig:r2_3}
\end{figure}

\begin{figure}[!htbp]
    \centering
    
    \rowexample
    {\includegraphics[width=0.95\linewidth]{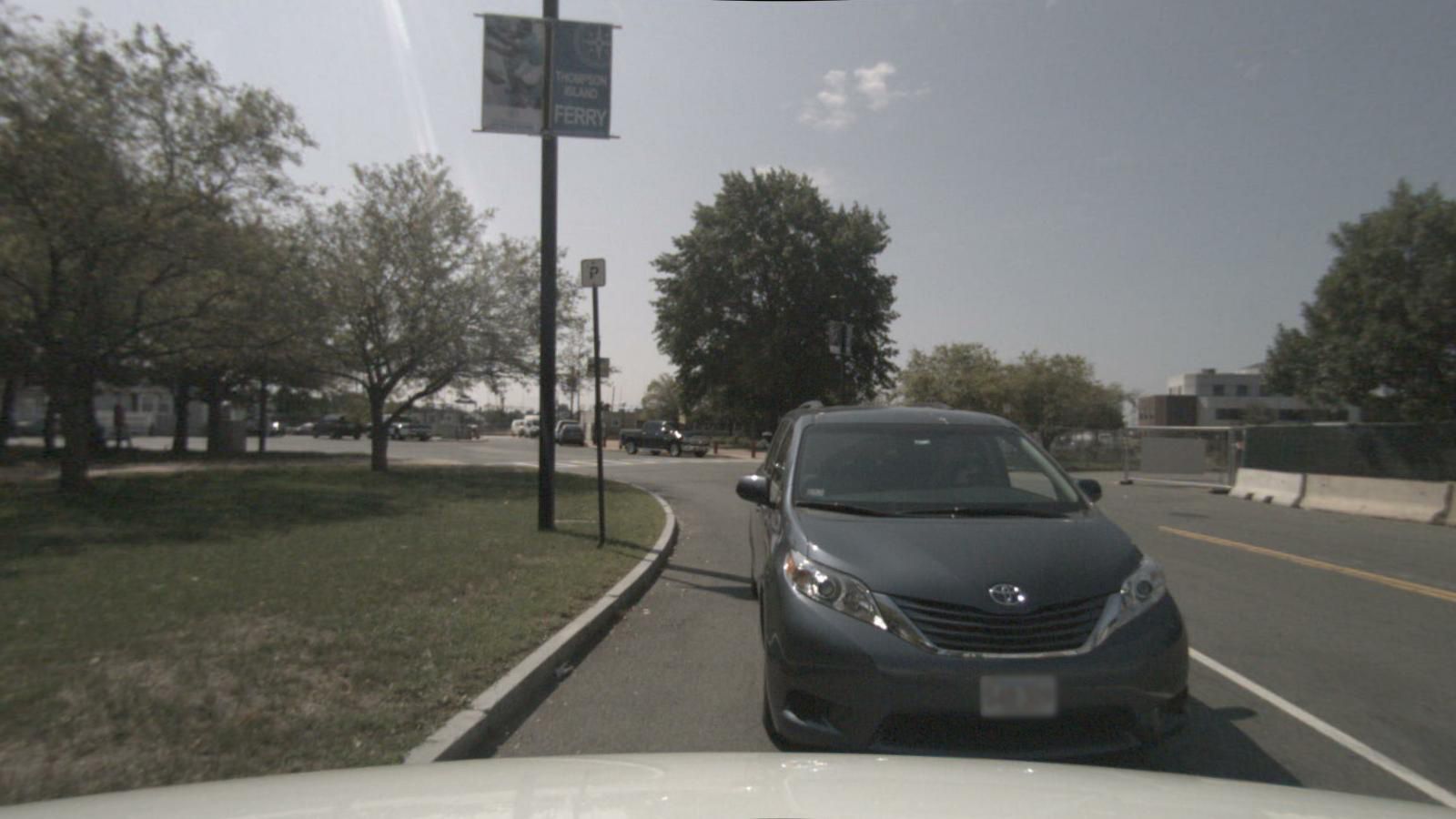}}
    {You are a mature and cautious driver with many years of driving experience. You now need assistance answering questions related to the driving process. \par Choose the correct option and output only the option letter (e.g., A/B/C/D).}
    {In CAM\_BACK, what is the future state of the Black car, located at [755.0, 400.0, 1404.0, 900.0]? \par ["A. Slightly offset to the right while maneuvering.", "B. Decelerate gradually without braking.", "C. Back up.", "D. Accelerate and go ahead."]}
    {B. Decelerate gradually without braking.}
    {R3-1 Outcome Prediction}

    \vspace{0.5em}
    \noindent\rule{0.97\linewidth}{0.4pt}
    \caption{Representative example from Rank 3-1 Questions.}
    \label{fig:r3_1}
\end{figure}

\begin{figure}[!htbp]
    \centering

    \rowexampleblock
    {
        \imgpair{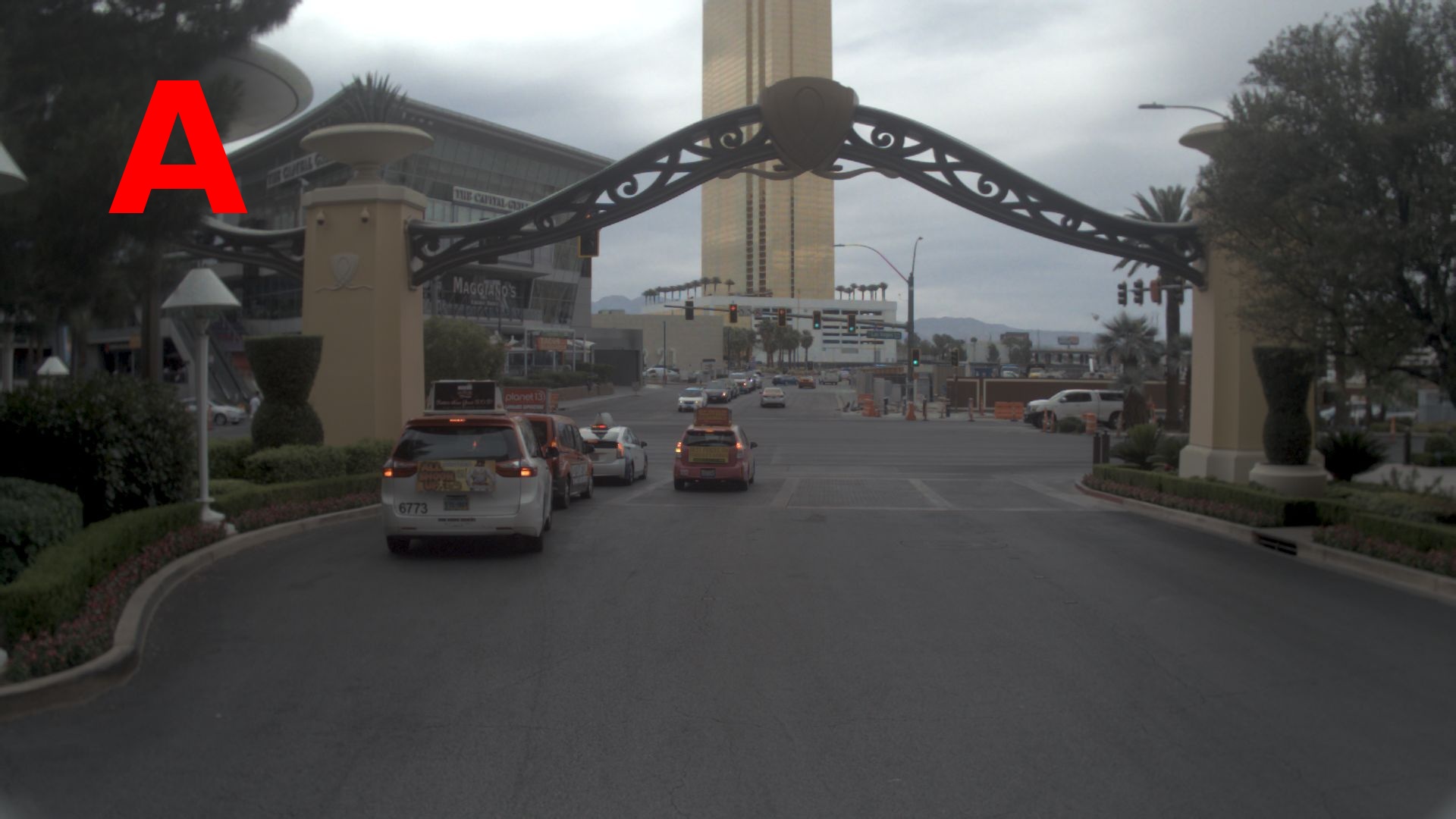}{$F_A$}{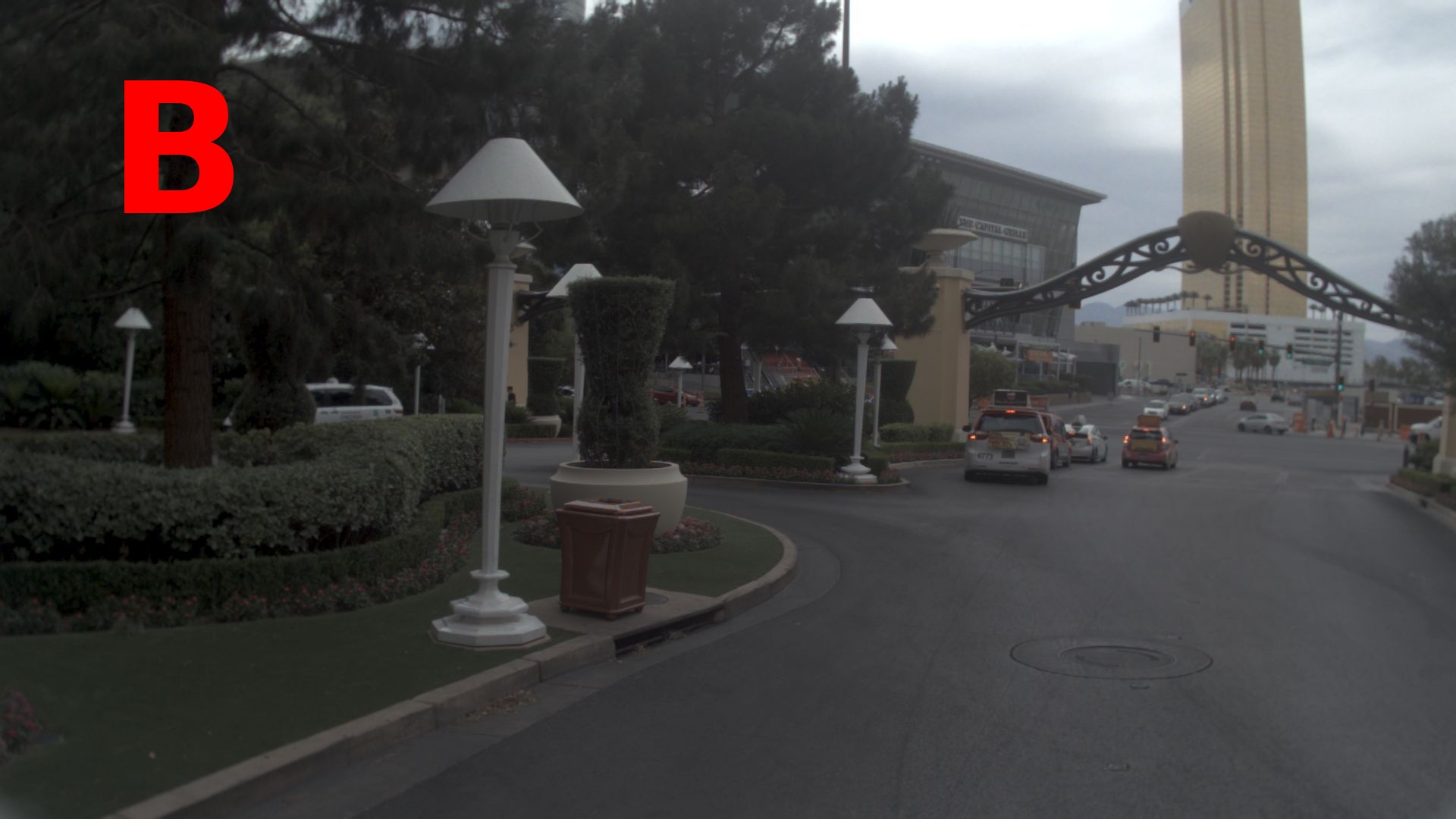}{$F_B$}
        \vspace{0.4em}

        \imgpair{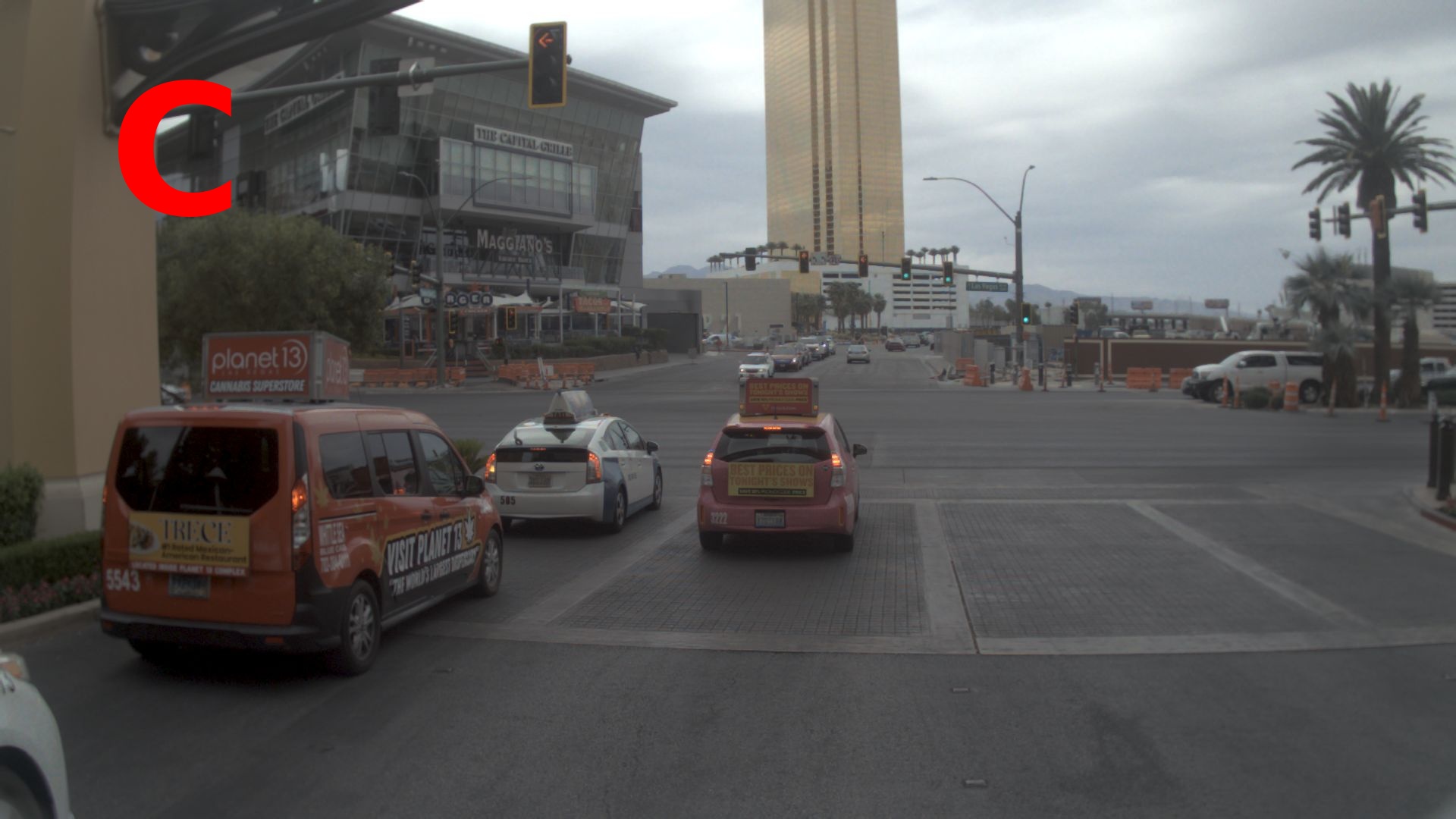}{$F_C$}{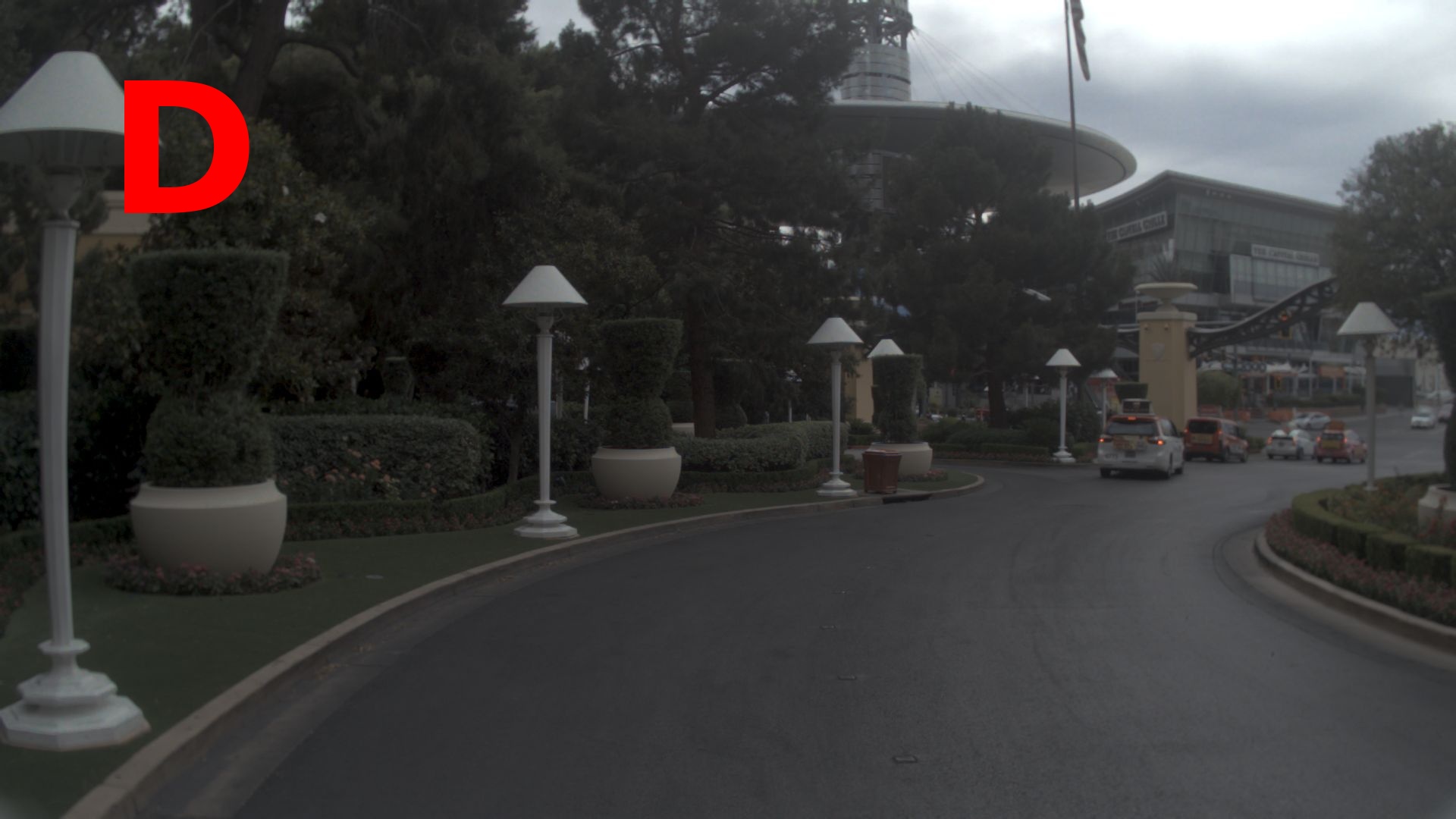}{$F_D$}
        \vspace{0.4em}

        \imgsinglecenter{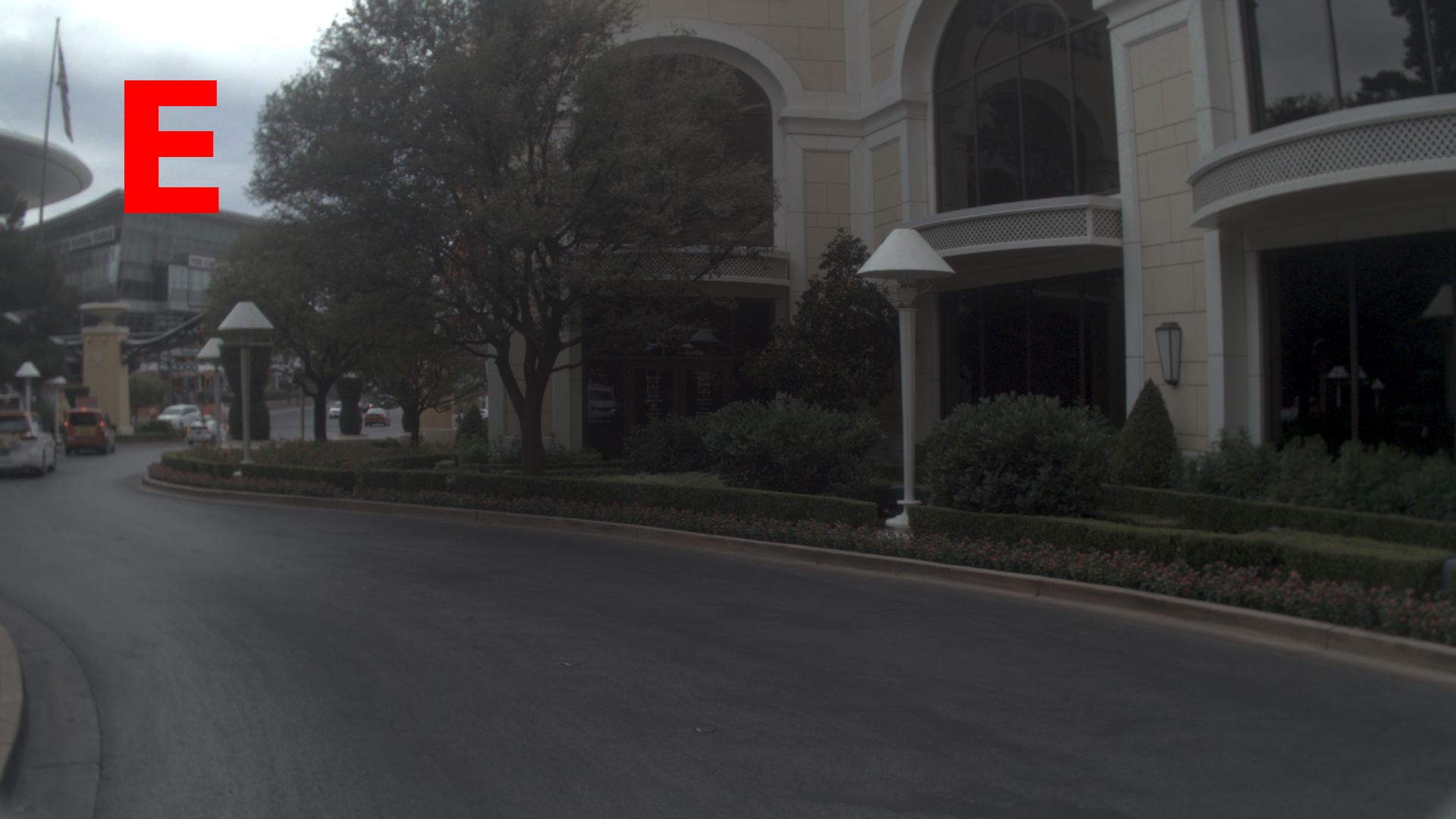}{$F_E$}
    }
    {
        \qatextblock
        {R3-2 Sequential Planning}
        {You are a mature and cautious driver with many years of driving experience. You now need assistance answering questions related to the driving process. \par Output only the ordered list of option labels in the format ["A", "B", ...].}
        {Please determine the correct order of the images. Option labels: A, B, C, D, E.}
        {["E", "D", "B", "A", "C"]}
    }

    \vspace{0.5em}
    \noindent\rule{0.97\linewidth}{0.4pt}
    \caption{Representative example from Rank 3-2 Questions.}
    \label{fig:r3_2}
\end{figure}

\newcommand{\closeloopexample}[3]{
\vspace{0.5em}
\noindent
\setlength{\fboxsep}{6pt}
\setlength{\fboxrule}{0.5pt}
\noindent\rule{0.97\linewidth}{0.4pt}
\begin{minipage}[t]{0.97\linewidth}
\vspace{0pt}
\begin{minipage}[t]{0.60\linewidth}
    \vspace{0pt}
    \centering
    #1
\end{minipage}
\hfill
\begin{minipage}[t]{0.37\linewidth}
    \vspace{0pt}
    {\small \textbf{#3}\par}
    \vspace{0.4em}
    {\footnotesize \textbf{Sce. Description:} #2\par}
\end{minipage}
\end{minipage}
\vspace{0.5em}
}

\section{Description of scenarios in \textit{R4}}
\label{R4_description}
\begin{figure}[!htp]
    \centering
    \includegraphics[width=0.7\linewidth]{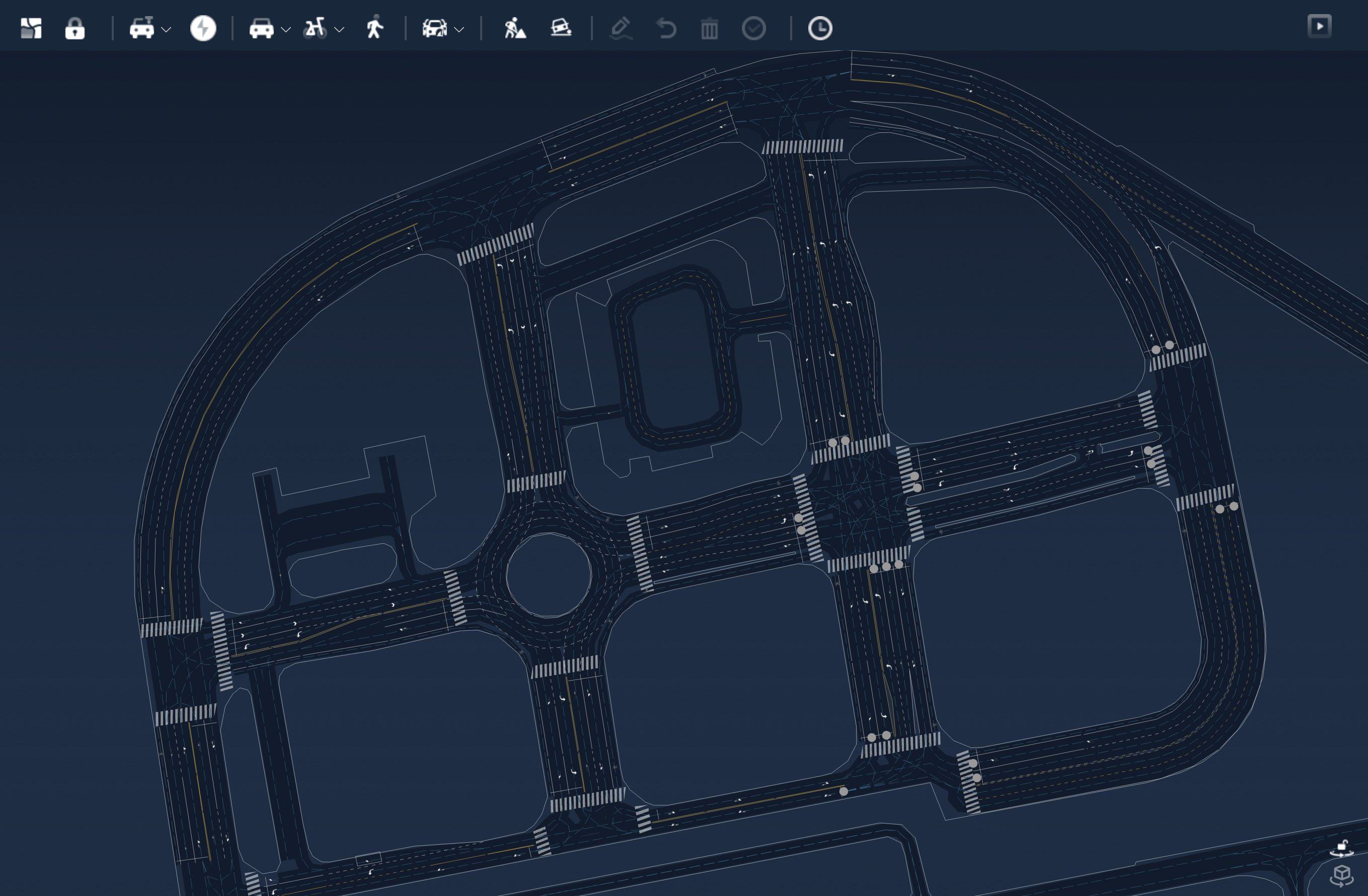}
    \caption{Visualization of part of the Scene Editor.}
    \label{fig:Fig_Editor_TJ}
\end{figure}

The current version of \textsc{DriveHierarchy} features 10 categories of typical scenarios, all of which are constructed using the scenario editor illustrated in Fig~\ref{fig:Fig_Editor_TJ}. The specific details of these 10 scenario categories are outlined below. As noted in Section~\ref{Future_work}, future scenarios will be further expanded upon this foundation.

\begin{figure}[!htbp]
    \centering
    
    \closeloopexample
    {\includegraphics[width=0.95\linewidth]{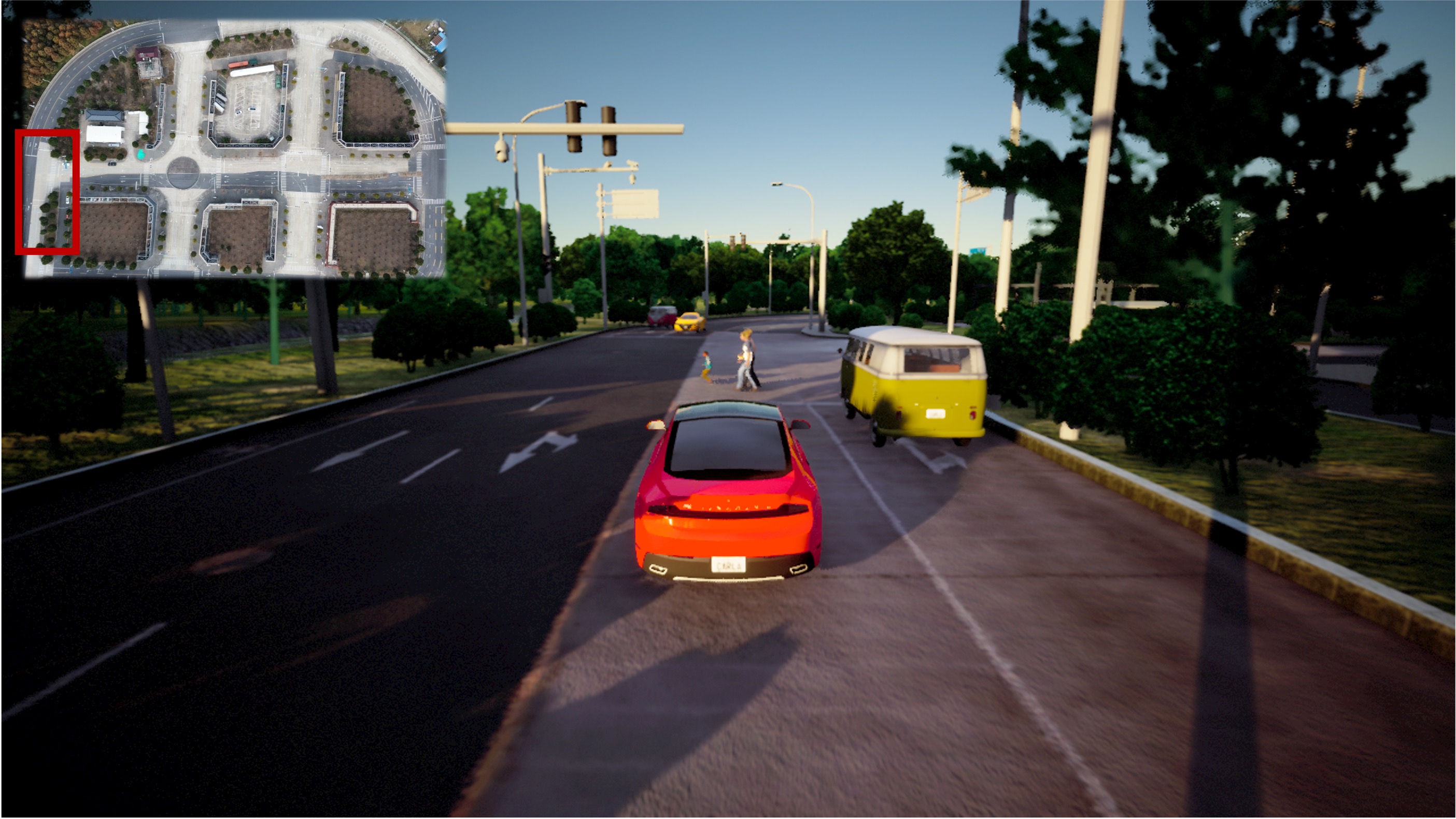}}
    {The ego vehicle encounters pedestrians crossing ahead and must detect the crossing intent, decelerate in time, yield safely when required, and resume driving only after the conflict region is clear.}
    {R4-1 Pedestrian encounters.}

    \closeloopexample
    {\includegraphics[width=0.95\linewidth]{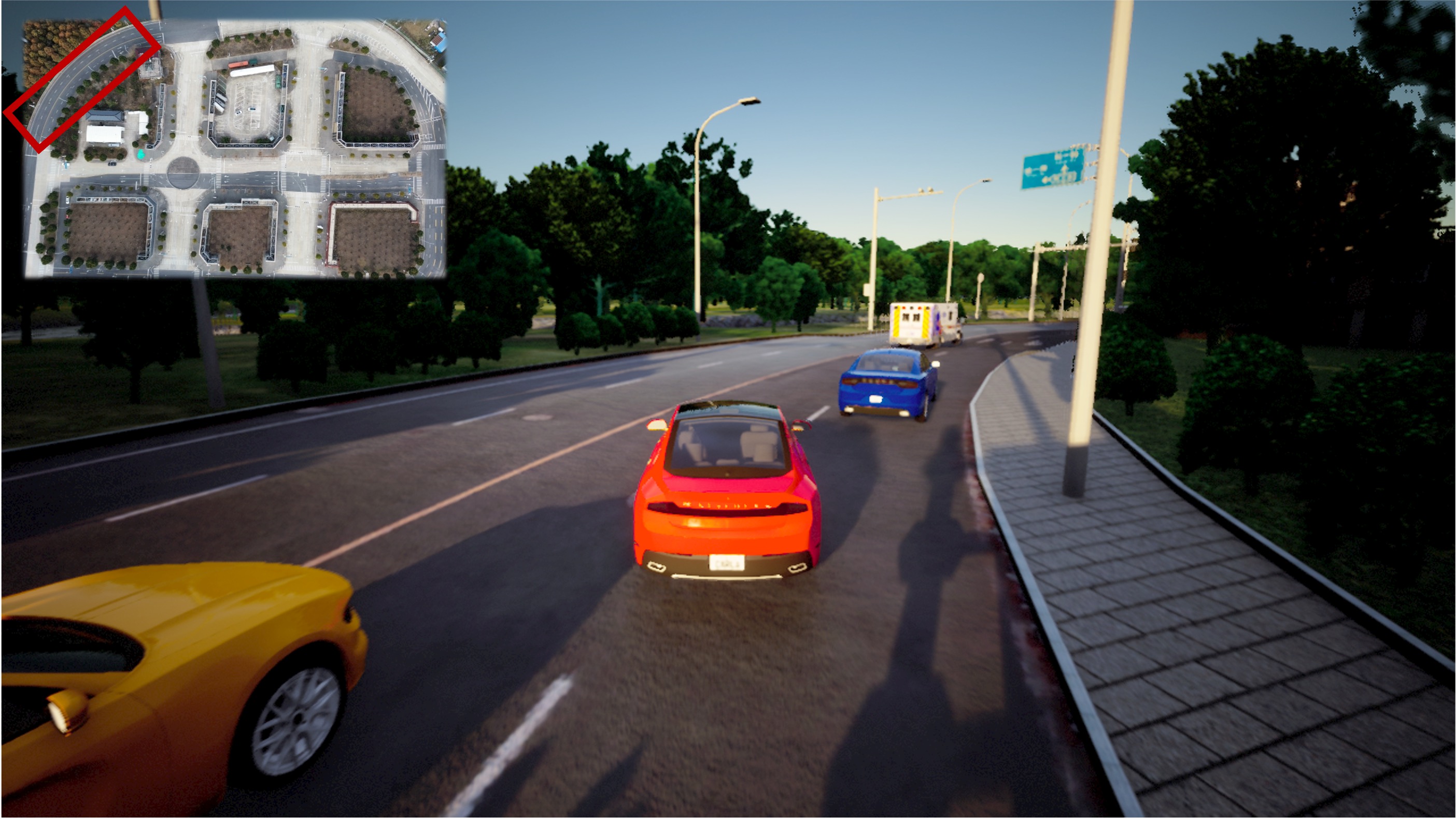}}
    {The ego vehicle faces static obstacles along its lane and must select a safe avoidance maneuver while preserving route continuity and preventing unnecessary collision risk.}
    {R4-2 Obstacle avoidance.}

    \vspace{0.5em}
    \noindent\rule{0.97\linewidth}{0.4pt}
    \caption{Representative Scenarios from Rank 4-1 to Rank 4-2.}
    \label{fig:r4_1}
\end{figure}

\begin{figure}[!htbp]
    \centering
    
    \closeloopexample
    {\includegraphics[width=0.95\linewidth]{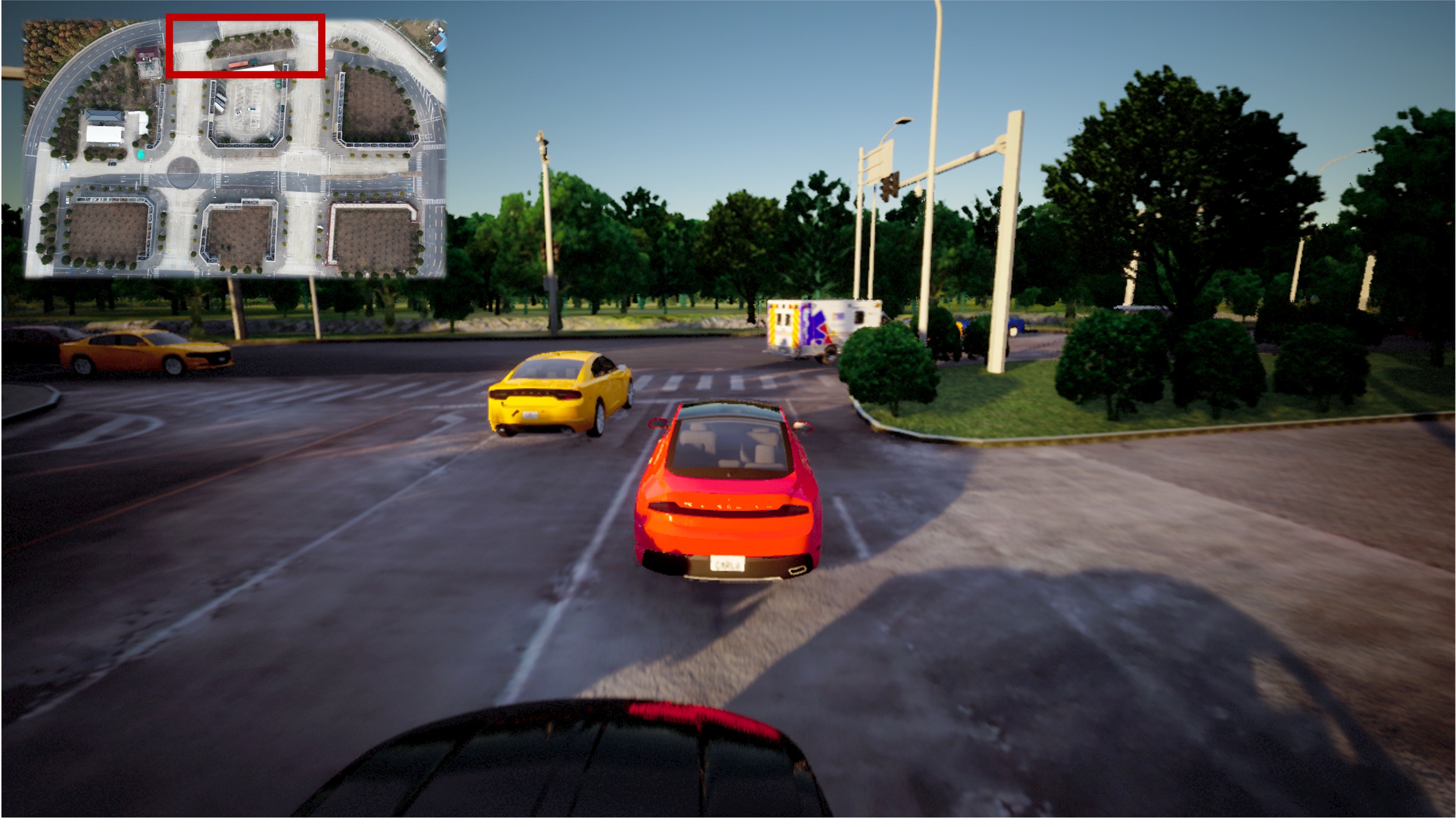}}
    {The ego vehicle performs a turning operation under constrained road geometry and surrounding traffic, requiring coordinated speed control, path tracking, and conflict-aware maneuver execution.}
    {R4-3 Turning maneuver.}

    \closeloopexample
    {\includegraphics[width=0.95\linewidth]{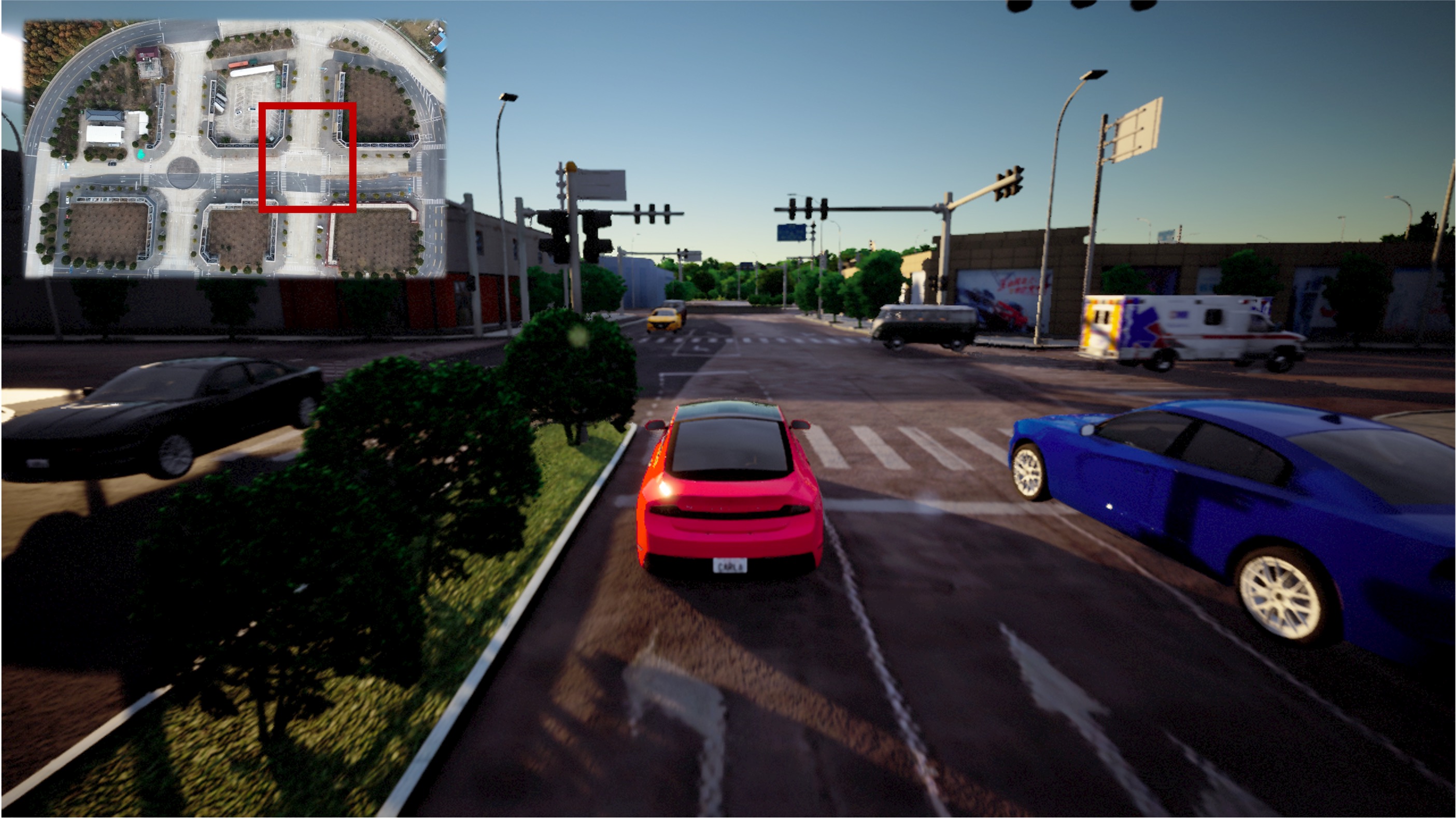}}
    {The ego vehicle traverses an intersection with non-fixed entry and exit directions, and must resolve multi-agent interactions, respect traffic rules, and choose an appropriate passage strategy under potentially competing movements.}
    {R4-4 Intersection.}

    \closeloopexample
    {\includegraphics[width=0.95\linewidth]{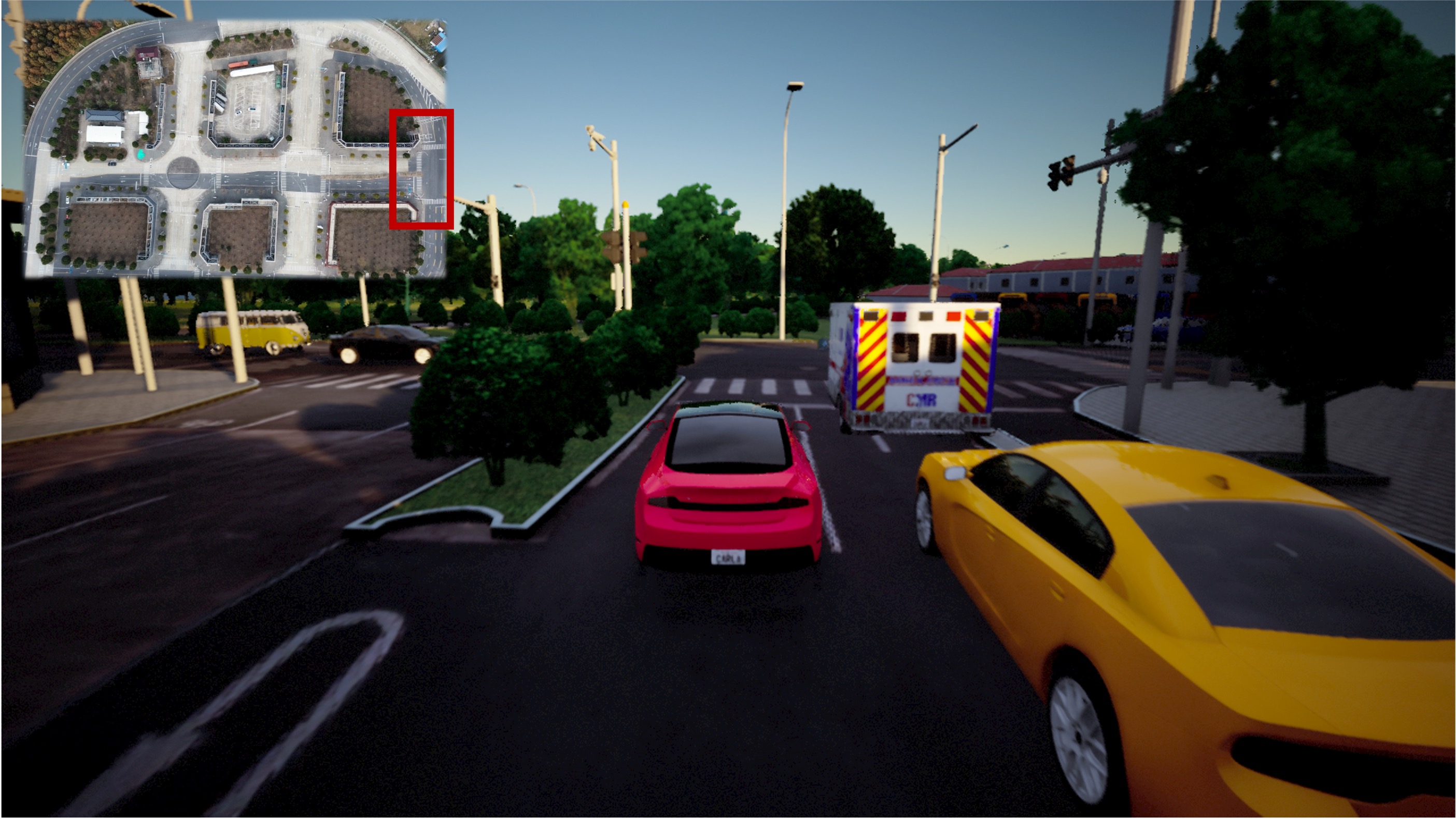}}
    {The ego vehicle traverses a T-intersection and must infer right-of-way, negotiate crossing or joining traffic, and complete the maneuver without blocking or unsafe encroachment.}
    {R4-5 T-intersection.}

    \closeloopexample
    {\includegraphics[width=0.95\linewidth]{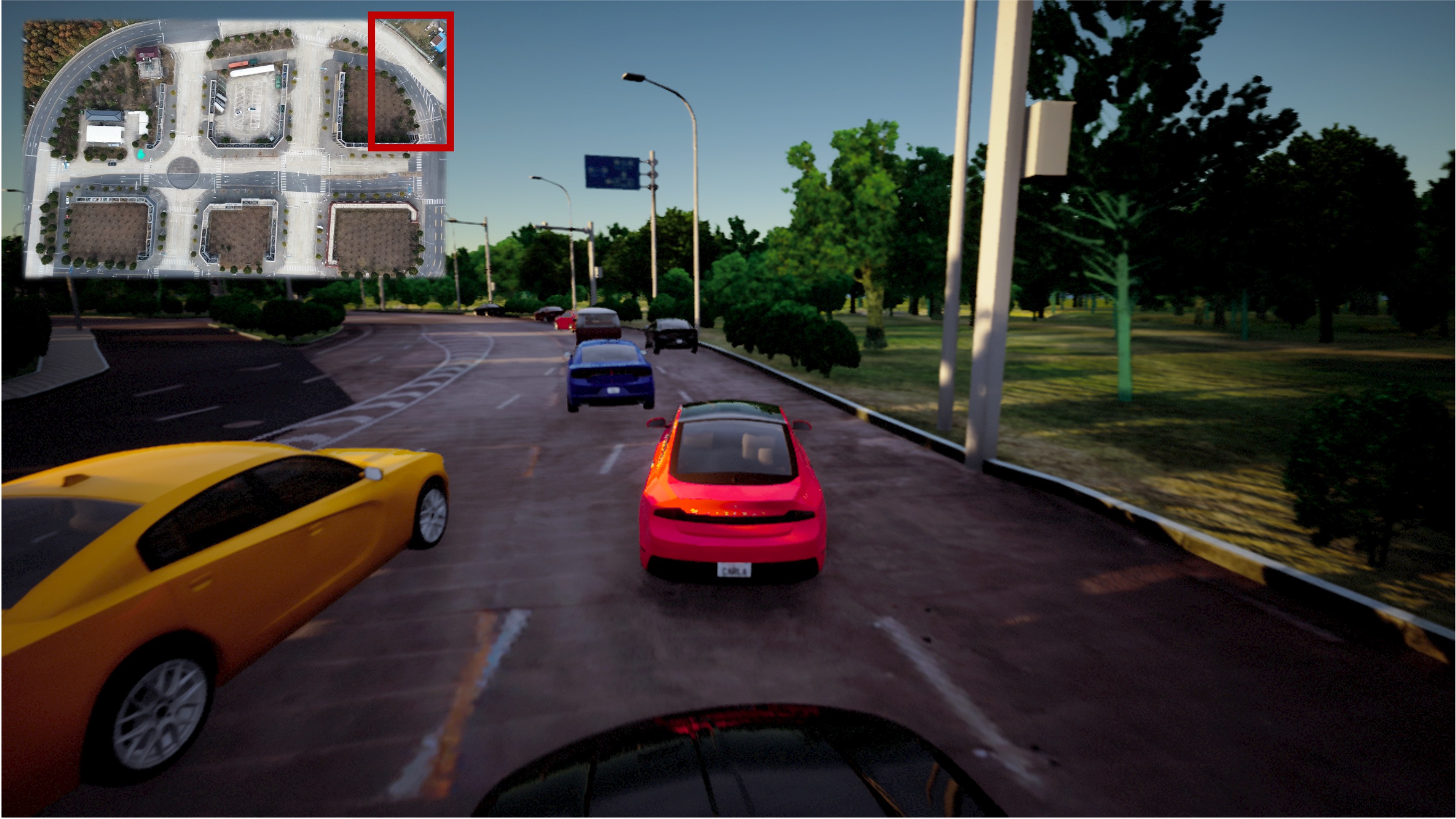}}
    {The ego vehicle operates in disturbed traffic conditions involving irregular surrounding motion and must remain stable under interference, preserving both safety margins and route progress.}
    {R4-6 Traffic disturbances.}

    \vspace{0.5em}
    \noindent\rule{0.97\linewidth}{0.4pt}
    \caption{Representative Scenarios from Rank 4-3 to Rank 4-6.}
    \label{fig:r4_3}
\end{figure}

\begin{figure}[!htbp]
    \centering
    
    \closeloopexample
    {\includegraphics[width=0.95\linewidth]{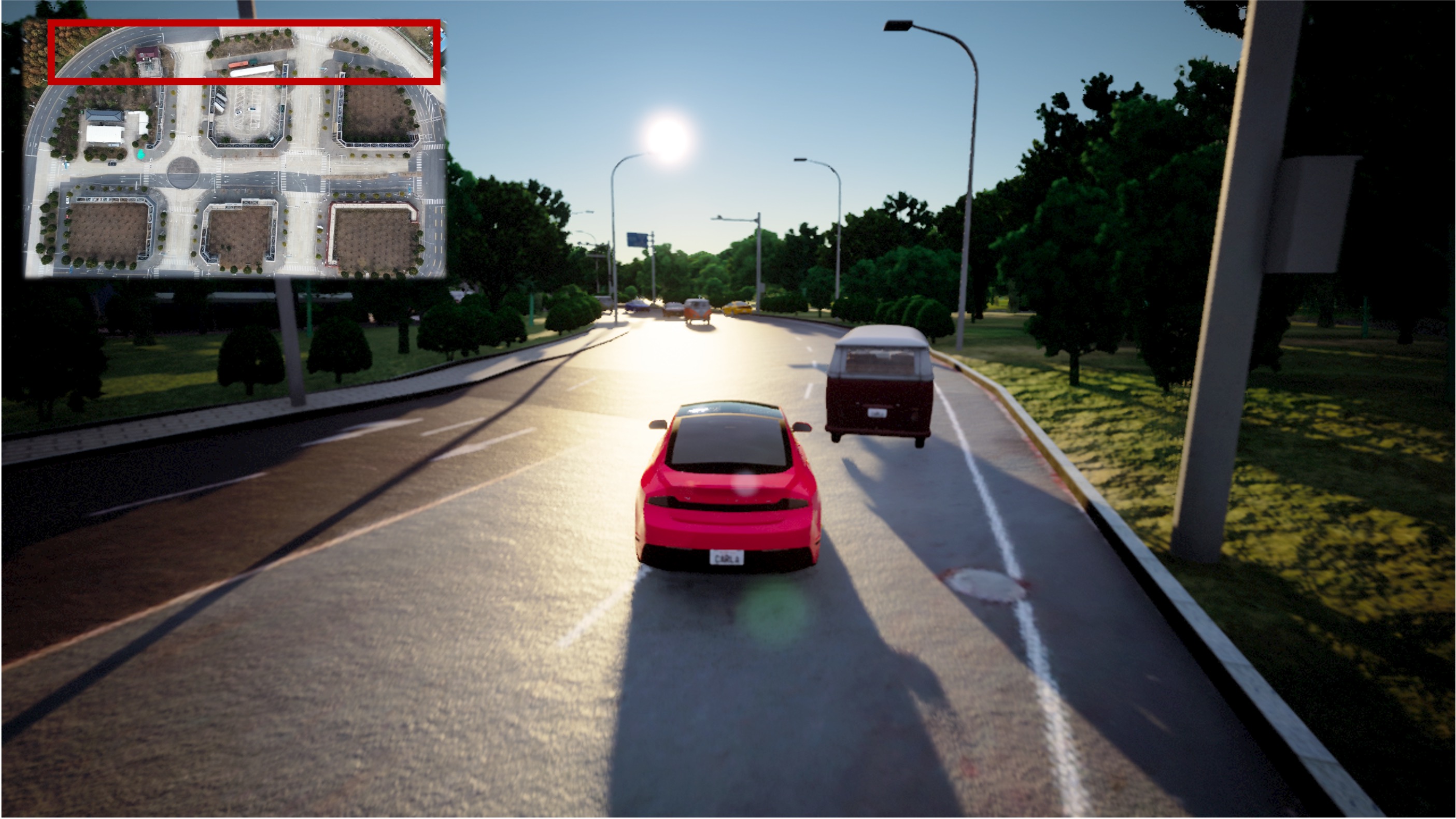}}
    {The ego vehicle encounters abrupt deceleration of the leading vehicle and must react promptly with safe longitudinal control while avoiding rear-end collision and excessive instability.}
    {R4-7 Sudden braking.}

    \closeloopexample
    {\includegraphics[width=0.95\linewidth]{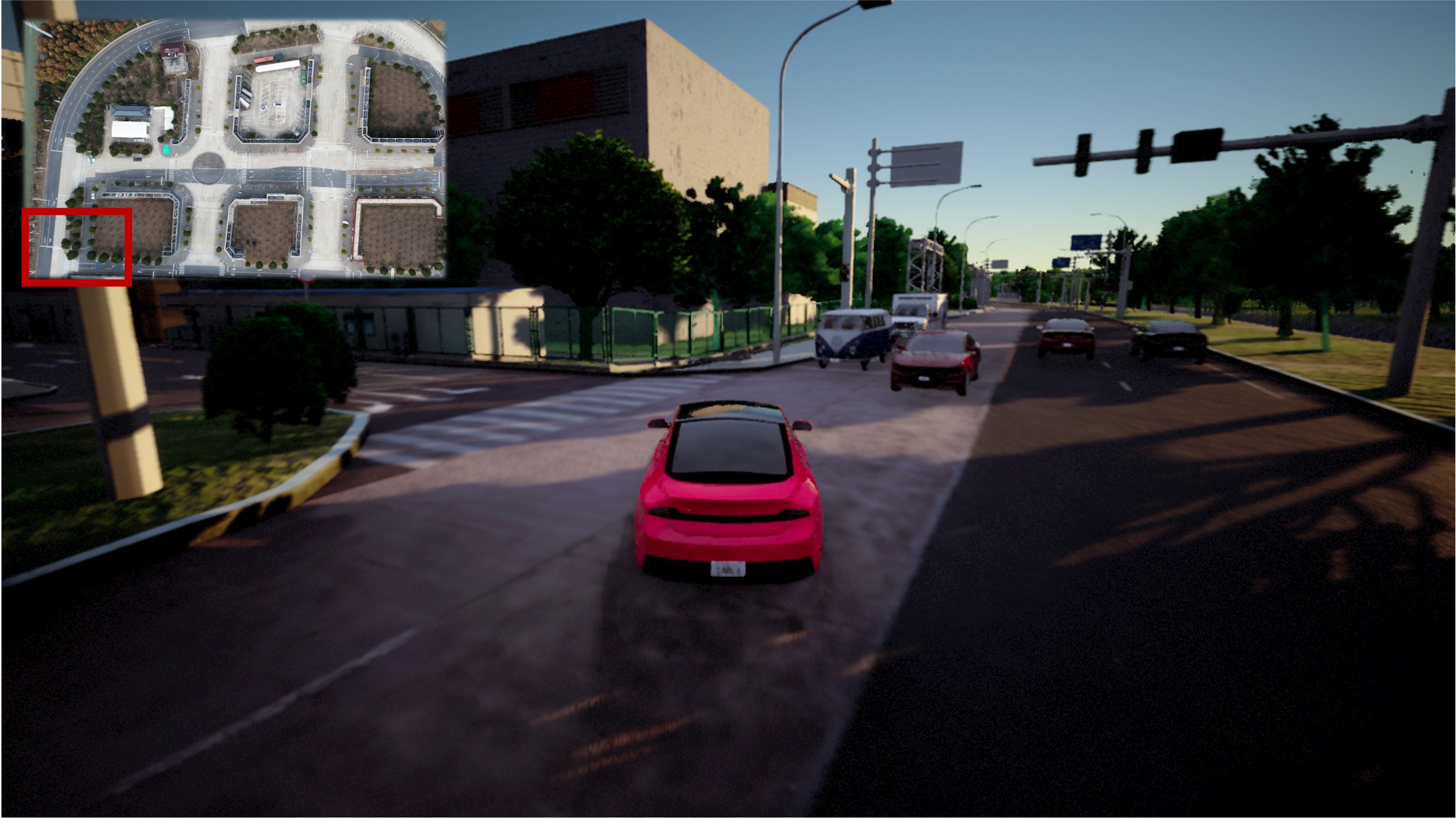}}
    {The ego vehicle performs lane joining and turning in the presence of neighboring traffic, requiring accurate gap assessment, timing control, and smooth trajectory adjustment.}
    {R4-8 Merging.}

    \closeloopexample
    {\includegraphics[width=0.95\linewidth]{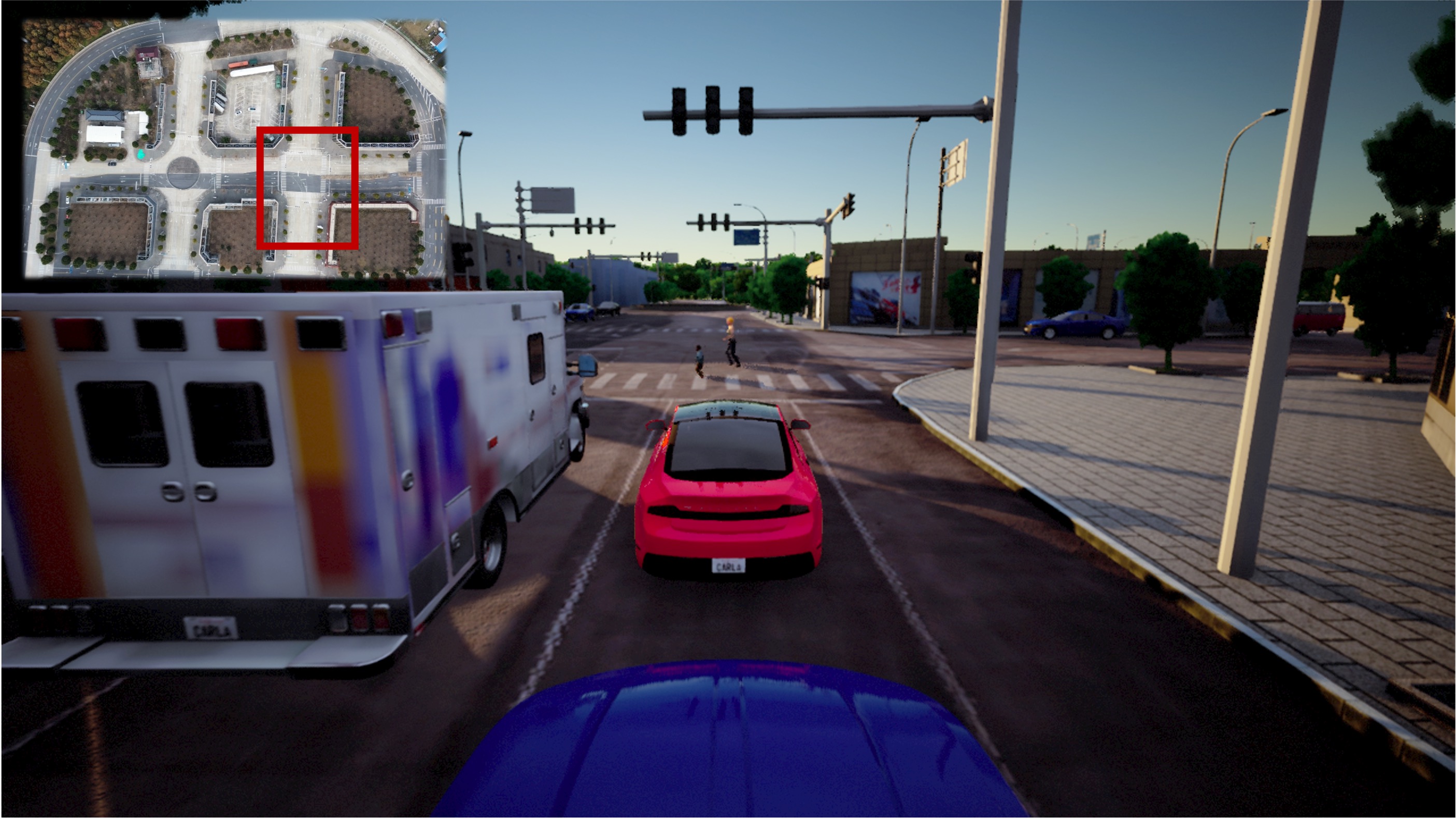}}
    {The ego vehicle approaches a yield situation and must identify priority conflicts, slow or stop when necessary, and proceed only when the available gap satisfies safe interaction constraints.}
    {R4-9 Yielding.}

    \closeloopexample
    {\includegraphics[width=0.95\linewidth]{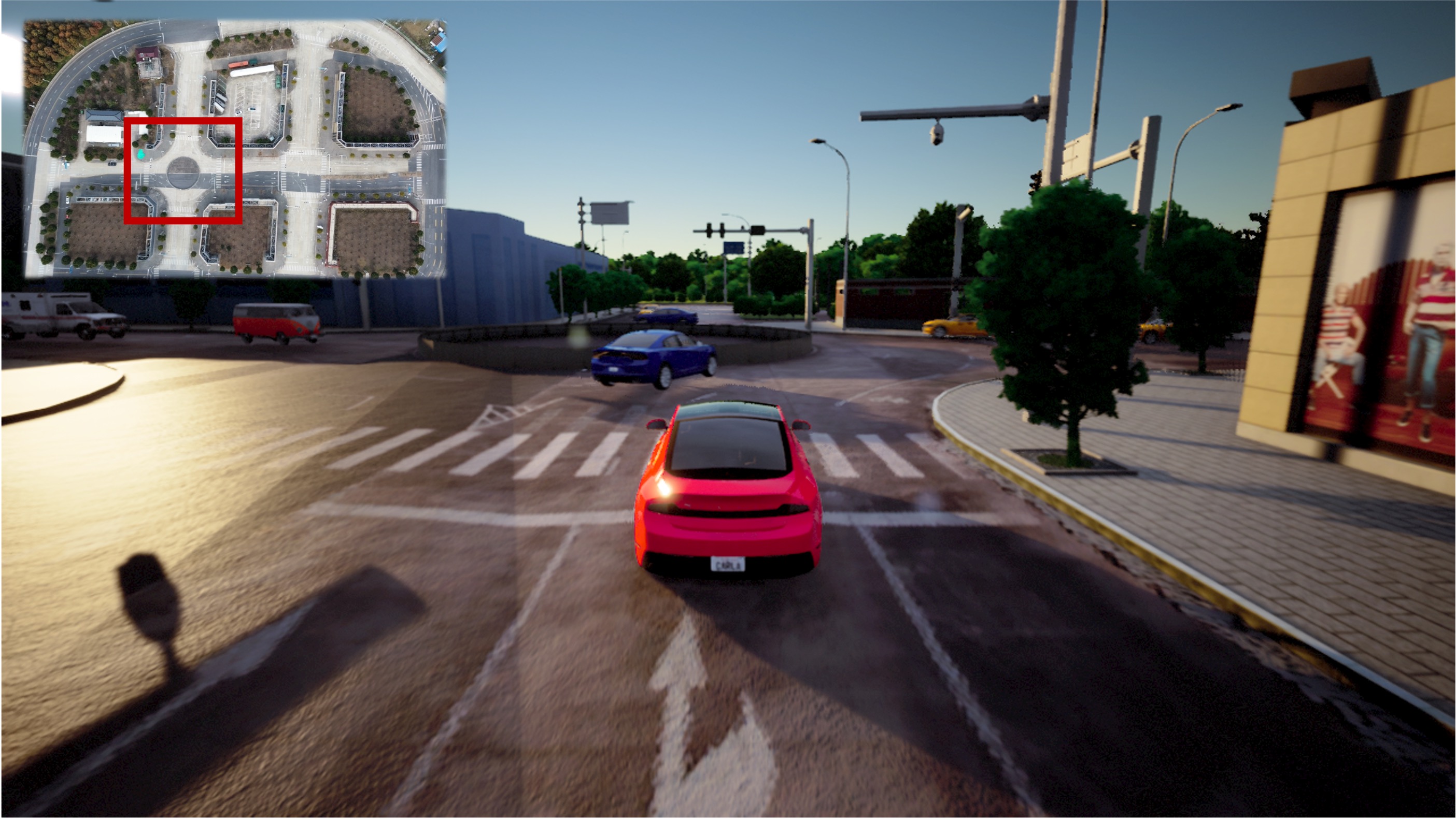}}
    {The ego vehicle navigates a roundabout under continuous multi-agent interaction with non-fixed entry and exit directions, and must maintain lane discipline, merge at an appropriate time, and exit without disrupting surrounding flow.}
    {R4-10 Roundabout navigation.}

    \vspace{0.5em}
    \noindent\rule{0.97\linewidth}{0.4pt}
    \caption{Representative Scenarios from Rank 4-7 to Rank 4-10.}
    \label{fig:r4_7}
\end{figure}

\section{Public Resources Used}
\label{Public_Resources_Used}
We build \textsc{DriveHierarchy} using public datasets, simulators, model checkpoints, and training/evaluation frameworks. 
We do not claim ownership of these resources, and all original data, codebases, and checkpoints remain the property of their respective authors or organizations under their corresponding licenses or terms of use. 
We release only the benchmark assets, annotations, metadata, and code that are permitted for redistribution, and users should obtain the original materials from the official sources when required. 
In addition, part of the closed-loop scoring implementation is adapted from Bench2Drive to fit our benchmark design and evaluation. 

\begin{table}[htp]
    \centering
    \caption{\textbf{Public resources used in this work.} Exact checkpoint names for evaluated models are listed in Table~\ref{tab:model_configurations}.}
    \label{tab:public_resources}
    \begin{tabularx}{\linewidth}{
    >{\raggedright\arraybackslash}X
    >{\raggedright\arraybackslash}X
    >{\raggedright\arraybackslash}X
    Y}
    \toprule
    \textbf{Resource} & \textbf{Type} & \textbf{License / Terms} & \textbf{Official Link} \\
    \midrule
    NuScenes & Dataset & CC BY-NC-SA 4.0 & \href{https://www.nuscenes.org}{Data} \\
    nuPlan & Dataset & Apache License 2.0 & \href{https://www.nuplan.org/}{Data} \\
    NAVSIM & Dataset & Apache License 2.0 & \href{https://github.com/autonomousvision/navsim}{Code} \\
    DriveLM & Dataset & Apache License 2.0 & \href{https://github.com/OpenDriveLab/DriveLM}{Code} \\
    DriveBench & Dataset / Benchmark & Apache License 2.0 & \href{https://drive-bench.github.io/}{Project} \\
    LingoQA & Dataset & Non-commercial use & \href{https://github.com/wayveai/lingoqa}{Project} \\
    DRAMA & Dataset & Non-commercial academic use & \href{https://usa.honda-ri.com/drama}{Data} \\
    CARLA & Simulator & MIT & \href{https://carla.org/}{Project} \\
    SUMO & Simulator & EPL-2.0 & \href{https://sumo.dlr.de}{Project} \\
    Bench2Drive & Code / Benchmark & CC-BY-NC-ND & \href{https://github.com/Thinklab-SJTU/Bench2Drive}{Code} \\
    Bench2Drive-VL & Code / Benchmark & CC-BY-NC-ND & \href{https://github.com/Thinklab-SJTU/Bench2Drive-VL}{Code} \\
    Qwen family & Model checkpoints & Apache License 2.0 & \href{https://huggingface.co/Qwen}{Model} \\
    InternVL family & Model checkpoints & Apache License 2.0 & \href{https://huggingface.co/OpenGVLab}{Model} \\
    MiniCPM-V & Model checkpoint & Apache License 2.0 & \href{https://huggingface.co/openbmb}{Model} \\
    Gemma & Model checkpoints & Google Gemma 27 Terms of Use & \href{https://ai.google.dev/gemma/terms}{Model} \\
    Pixtral & Model checkpoint & Apache License 2.0 & \href{https://huggingface.co/mistralai/Pixtral-12B-2409}{Model} \\
    ZwZ & Model checkpoint & Apache License 2.0 & \href{https://huggingface.co/inclusionAI}{Model} \\
    DA-DriveLM & Model checkpoint & MIT & \href{https://huggingface.co/OpenGVLab/Mini-InternVL2-4B-DA-DriveLM}{Model} \\
    ReasonDrive & Model checkpoint & Apache License 2.0 & \href{https://huggingface.co/ac4462/Qwen2.5-VL-7B-DriveLM}{Model} \\
    swift-sft & Training framework & Apache License 2.0 & \href{https://github.com/modelscope/ms-swift}{Code} \\
    vLLM & Software & Apache License 2.0 & \href{https://github.com/vllm-project/vllm}{Code} \\
    \bottomrule
    \end{tabularx}
\end{table}



\end{document}